%% file: main.tex
\documentclass{article}
\PassOptionsToPackage{table}{xcolor}

\usepackage{microtype}
\usepackage{graphicx}
\usepackage{subcaption}
\usepackage{booktabs} 

\usepackage[hyperfootnotes=false]{hyperref}

\usepackage[preprint]{icml2026}

\usepackage{amsmath}
\usepackage{amssymb}
\usepackage{mathtools}
\usepackage{amsthm}

\input{math_commands.tex}

\usepackage{xurl}
\usepackage{multirow}

\usepackage{url}
\usepackage{threeparttable}
\usepackage{adjustbox}
\usepackage[most]{tcolorbox}

\usepackage{tabularx}
\usepackage{makecell}
\usepackage{enumitem}

\input{tables/table_palette}

\definecolor{citeblue}{RGB}{45,85,145}

\hypersetup{
  colorlinks=true,
  citecolor=citeblue, 
  linkcolor=citeblue,    
  urlcolor=citeblue   
}
\definecolor{qwenblue}{HTML}{5B7DB1}
\definecolor{glmviolet}{HTML}{7867A8}
\definecolor{kimigreen}{HTML}{5F8F7B}

\definecolor{boxgray}{HTML}{F7F8FA}
\definecolor{textgray}{HTML}{777777}
\definecolor{linegray}{HTML}{D9DCE2}

\tcbuselibrary{listings,breakable,skins}

\definecolor{promptframe}{HTML}{3B6EA8}
\definecolor{promptback}{HTML}{F5F8FC}

\definecolor{OPDBlue}{HTML}{2F6BFF}
\definecolor{MetricGray}{HTML}{777777}

\definecolor{BestBlue}{RGB}{185, 192, 218}
\definecolor{SecondBlue}{RGB}{210, 215, 232}

 \newtcblisting{promptbox}[1]{%
    breakable,
    enhanced,
    arc=2pt,
    boxrule=0.5pt,
    colframe=promptframe,
    colback=promptback,
    coltitle=white,
    fonttitle=\bfseries\small,
    title={#1},
    top=4pt,
    bottom=4pt,
    left=6pt,
    right=6pt,
    listing only,
    listing options={
      basicstyle=\ttfamily\small,
      breaklines=true,
      breakatwhitespace=true,
      keepspaces=true,
      columns=fullflexible,
      breakindent=0pt,
      postbreak={}
    }
  }

\tcbset{
    reasoningcase/.style={
        enhanced,
        colback=gray!5,
        colframe=gray!40,
        coltext=black!85,
        coltitle=black,
        fonttitle=\bfseries\small,
        fontupper=\small,
        boxrule=0.5pt,
        arc=1.5mm,
        outer arc=1.5mm,
        left=3mm,
        right=3mm,
        top=2.5mm,
        bottom=2.5mm,
        toptitle=2.2mm,
        bottomtitle=2.2mm,
        boxsep=0pt
    }
}

\usepackage[capitalize,noabbrev]{cleveref}

\theoremstyle{plain}
\newtheorem{theorem}{Theorem}[section]
\newtheorem{proposition}[theorem]{Proposition}
\newtheorem{lemma}[theorem]{Lemma}
\newtheorem{corollary}[theorem]{Corollary}
\theoremstyle{definition}

\theoremstyle{remark}

\icmltitlerunning{Fork Preference Distillation}


\begin{document}


\twocolumn[
\icmltitle{Fork Preference Distillation}

\icmlsetsymbol{equal}{*}
\icmlsetsymbol{corr}{\textdagger}

\begin{icmlauthorlist}
\icmlauthor{Jingwei Song}{sjtu,equal}
\icmlauthor{Haofeng Xu}{hku,equal}
\icmlauthor{Jiacheng Liu}{sjtu,equal}
\icmlauthor{Qian Xu}{ruc}
\icmlauthor{Linfeng Zhang}{sjtu,corr}
\end{icmlauthorlist}

\icmlaffiliation{sjtu}{Shanghai Jiao Tong University}
\icmlaffiliation{hku}{The University of Hong Kong}
\icmlaffiliation{ruc}{Renmin University of China}

\icmlcorrespondingauthor{Jingwei Song}{songjingwei@connect.hku.hk}

\vskip 0.3in
]

\printAffiliationsAndNotice{\textsuperscript{*}Equal contribution \textsuperscript{\textdagger}Corresponding author }

\begin{abstract}
\input{sections/0_abstract}
\end{abstract}

\input{sections/1_introduction}
\input{sections/2_related_work}
\input{sections/preliminary}
\input{sections/3_method}
\input{sections/4_experiments}
\input{sections/5_ablation_analysis}
\input{sections/6_conclusion}


\clearpage
\begingroup
\raggedbottom
\bibliography{icml2026}
\bibliographystyle{icml2026}

\clearpage
\endgroup
\appendix
\onecolumn
\input{sections/X_appendix}

\end{document}

%% file: math_commands.tex
\usepackage{amsmath,amsfonts,bm}

\def\1{\bm{1}}

\DeclareMathAlphabet{\mathsfit}{\encodingdefault}{\sfdefault}{m}{sl}
\SetMathAlphabet{\mathsfit}{bold}{\encodingdefault}{\sfdefault}{bx}{n}

\newcommand{\E}{\mathbb{E}}

\newcommand{\KL}{D_{\mathrm{KL}}}



%% file: tables/table_palette.tex
\definecolor{tblbase}{HTML}{6BAED6}

%% file: sections/0_abstract.tex
In sampled-token on-policy distillation (OPD), the teacher scores only the token the student chose on its own rollout. Discouraging an unfavorable choice leaves open what should replace it, giving little direct feedback when the student rarely tries the teacher's preferred alternative. Fork Preference Distillation (FPD) addresses this limitation at \emph{decision forks}, prefixes where a strong teacher and a frozen foil policy rank different next tokens first. At each fork, FPD augments standard OPD with a soft-label Bradley--Terry loss matching the strong teacher's conditional odds over the two candidates, even when the student samples neither. In three-domain multi-teacher distillation across math, code, and instruction following, FPD reuses a specialist from another domain as the cross-domain foil. In a two-token model, FPD preserves the strong teacher's distribution as the stationary target; for fixed teacher odds and step size, raising the preferred token's probability from $\epsilon$ to at least $1/2$ takes $\Theta\bigl(1/(\epsilon\log(1/\epsilon))\bigr)$ gradient steps under vanilla OPD versus $O\bigl(\log(1/\epsilon)/\lambda\bigr)$ under FPD ($\lambda > 0$, $\epsilon \to 0$). With Qwen3-1.7B students and 4B/8B teachers, FPD improves the seven-benchmark average over vanilla OPD by 3.0 and 8.4 points at matched steps; with Qwen3-4B specialists across three domains, FPD gains 4.6 points and stays stable over 240 updates, whereas vanilla OPD degenerates after about 180.

%% file: sections/1_introduction.tex
\section{Introduction}
\label{sec:introduction}
\input{figures/FPD}
Recent advances in large language models (LLMs) have broadened the range of complex tasks that a single model can perform, including mathematical reasoning, software engineering, and multi-step tool use~\citep{moonshot2026kimik31pro,deepseek2026v41flash,zai2026glm53,anthropic2026fable52,openai2026gpt62}. During post-training, on-policy distillation (OPD) is widely used to integrate expertise across domains and to recover capabilities acquired in earlier training stages. DeepSeek-V4 and Kimi K3 use multi-teacher OPD to consolidate domain specialists into a unified model~\citep{deepseekai2026v4,kimi2026k3}, while GLM-5 uses earlier checkpoints as teachers to recover skills that degraded during subsequent training~\citep{glm2026glm5}. In both settings, the student generates its own responses and receives token-level supervision at each prefix it visits.

Despite eliminating off-policy exposure bias, standard OPD can stall at critical reasoning pivots even on prefixes the student visits repeatedly. Whether the teacher is larger than the student or is a same-size model refined by reinforcement learning, its preferred next token (e.g., a reflective pivot such as \texttt{Wait}) often starts with very low probability under the student. In reverse-KL OPD, as the student probability of a token approaches zero, the gradient with respect to its logit vanishes, even under full-vocabulary supervision, while sampled-token OPD penalizes the student's chosen token without indicating what should replace it. Recent OPD variants inject teacher-preferred tokens heuristic-wise, such as using the teacher's top token as a hard target~\citep{tgpo} or switching to local top-$k$ matching when sampled advantages are nonpositive~\citep{jia2026aopd}. Conversely, standard pairwise preference methods such as DPO~\citep{rafailov2023direct} supply direct contrast between two candidates, but rely on hard binary labels that discard probability calibration and drive logit margins to infinity without a finite minimizer~\citep{azar2024general}.

To overcome the vanishing-signal bottleneck without sacrificing distributional calibration, we introduce \emph{\textbf{F}ork \textbf{P}reference \textbf{D}istillation} (\textbf{FPD}), a post-training paradigm that unifies trajectory-level on-policy distillation with calibrated preference learning at critical decision points. Instead of treating all token positions uniformly or relying solely on sampled-token scores, FPD centers targeted pairwise supervision around \emph{decision forks}: student-generated prefixes where a strong teacher $p^+$ and a frozen foil policy $p^-$ disagree on the top-ranked next token ($a \neq b$). Here $p^-$ acts purely as a foil, exposing prefixes where habitual behavior diverges from expert reasoning and proposing the competing continuation $b$, while the learning target comes exclusively from $p^+$. In multi-teacher distillation, existing domain specialists naturally serve both roles: for each domain, its own specialist acts as $p^+$ while the lowest-scoring specialist from another domain acts as $p^-$, turning cross-domain specialists into mutual foils at no extra model footprint.

FPD maintains on-policy distribution alignment across all response tokens while directly aligning the student with the strong teacher's conditional odds over $\{a, b\}$ at each fork via a soft-label Bradley--Terry objective. We show why this resolves a fundamental structural weakness of reverse KL: at any candidate pair, reverse KL weights a mismatch in conditional odds by the total probability mass $m = p(a) + p(b)$ that the student currently assigns to the pair. When $m$ is small, even a complete reversal of the teacher's ranking contributes negligibly to the reverse-KL loss. FPD decouples conditional preference learning from $m$, conveying whether the teacher favors $a$ over $b$ narrowly or decisively even when the student's rollout samples a third token. Crucially, once the student matches the teacher's conditional odds, the fork preference gradient vanishes, leaving the strong teacher $p^+$ as the unique interior stationary distribution of the ideal fixed-prefix objective (Section~\ref{sec:method-bound}).

Empirically, with a Qwen3-1.7B student, FPD improves the seven-benchmark average over vanilla OPD by 3.0 points with a 4B teacher and 8.4 points with an 8B teacher at matched training steps. In three-domain multi-teacher distillation with Qwen3-4B specialists, FPD outperforms standard multi-teacher OPD by 4.6 points on average and remains stable through 240 updates, whereas vanilla OPD degenerates after roughly 180 updates due to entropy and length collapse. Gains hold consistently across fork preference weights and foil checkpoints, requiring only one additional prefill per response.

Our contributions are threefold:
\begin{itemize}
    \item We propose \emph{\textbf{F}ork \textbf{P}reference \textbf{D}istillation} (\textbf{FPD}), a post-training paradigm that uses top-1 disagreements between a strong teacher and a frozen foil policy to determine both \emph{where} to intervene on student trajectories and \emph{which} continuations to compare. In early training rollouts, the student favors the foil's token over the strong teacher's choice at 81\% of forks, confirming that forks isolate systematic student bottlenecks.

    \item We prove that reverse KL attenuates conditional preference errors within a token pair by the pair's student probability mass, resulting in $\Theta\!\big(1/(\epsilon\log(1/\epsilon))\big)$ escape steps when the teacher's preferred token has initial probability $\epsilon \to 0$. By directly matching calibrated conditional odds at forks, FPD accelerates escape to $O\!\big(\log(1/\epsilon)/\lambda\big)$ steps while preserving the strong teacher as the unique stationary distribution.

    \item We demonstrate the effectiveness of FPD across single-teacher and multi-teacher distillation. Distilling Qwen3-1.7B from 4B and 8B teachers yields +2.0 and +8.8 average benchmark gains over vanilla OPD at best checkpoints. In three-domain distillation with Qwen3-4B specialists reused as cross-domain foils, FPD improves average performance by +4.6 points, also improves over the D3 baseline (28.78 $\rightarrow$ 31.98), and prevents late-stage training collapse.
\end{itemize}

%% file: figures/FPD.tex
\begin{figure*}[!t]
  \centering
  \includegraphics[width=\linewidth]{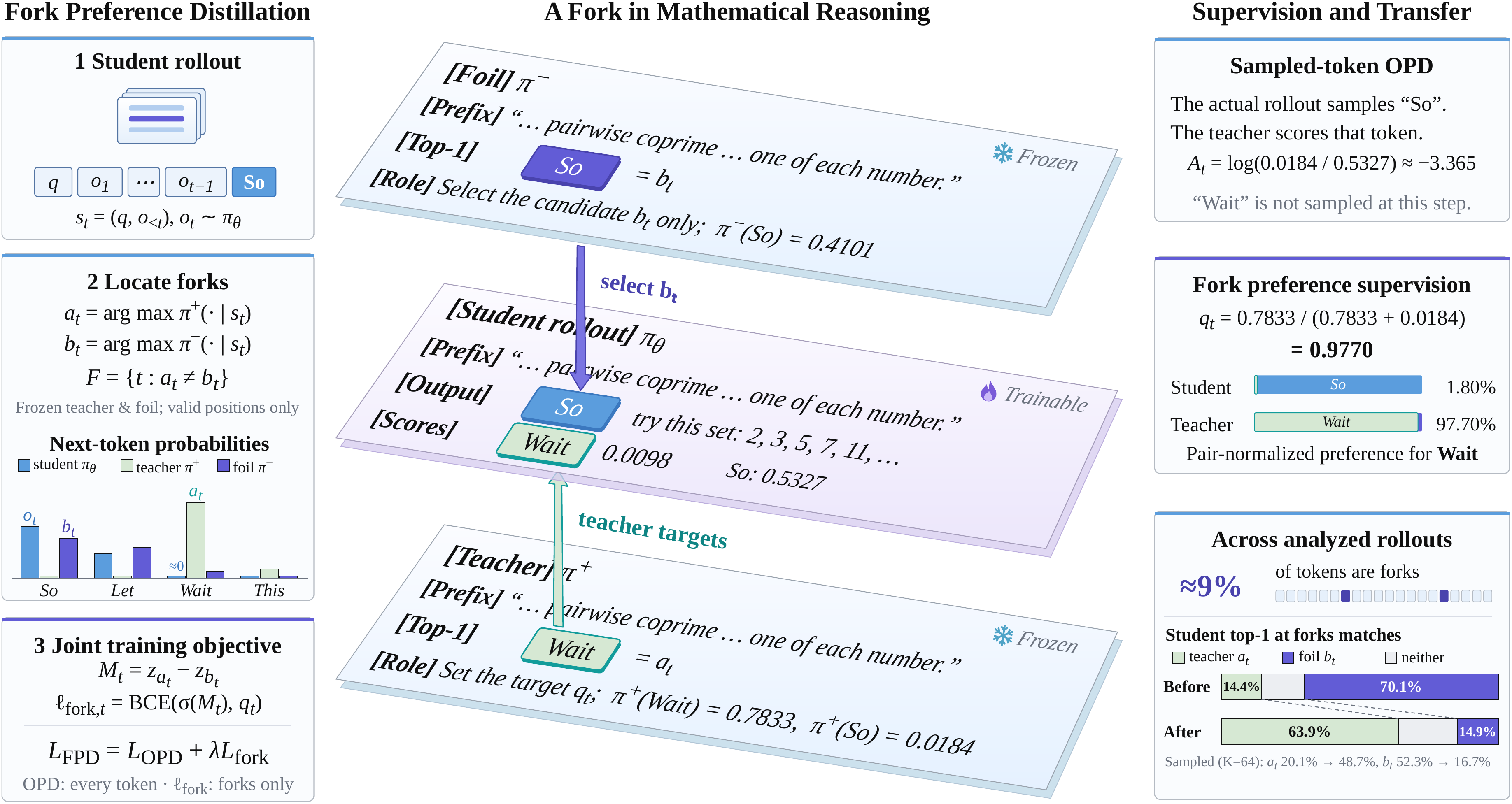}
  \caption{Overview of \textbf{Fork Preference Distillation (FPD)}. \emph{Left:} On student rollouts, FPD locates \emph{decision forks} $\mathcal{F} = \{t : a_t \neq b_t\}$ where a frozen strong teacher $\pi^+$ and foil policy $\pi^-$ disagree on the top next token, combining on-policy distillation with fork preference learning. \emph{Center:} At a mathematical reasoning fork, the student and foil favor a habitual continuation (\texttt{So}) whereas the strong teacher prefers a reflective pivot (\texttt{Wait}). \emph{Right:} Sampled-token OPD penalizes the chosen token without surfacing the teacher's alternative, while FPD matches the teacher's conditional odds over $\{a_t, b_t\}$ directly.}
  \label{fig:fpopd}
\end{figure*}

%% file: sections/2_related_work.tex
\section{Related Work}
\label{sec:related}

\paragraph{Token-level supervision on student-generated trajectories.}
On-policy distillation reduces the mismatch between training and inference contexts by supervising the student on its own trajectories. GKD studies a range of distillation objectives on student-generated sequences~\citep{agarwal2024gkd}. MiniLLM uses reverse KL for generative distillation~\citep{gu2024minillm}, while DistiLLM combines skew divergences with adaptive reuse of previously generated student outputs~\citep{ko2024distillm}. Recent work examines which teacher predictions warrant stronger supervision. \emph{Rethinking OPD} finds that successful distillation is accompanied by progressive alignment on high-probability tokens~\citep{li2026rethinking}. TGPO explicitly supervises the student on the teacher's top-ranked token~\citep{tgpo}, while AOPD matches the teacher's distribution over a local candidate set when the student-sampled token has nonpositive advantage~\citep{jia2026aopd}. Other approaches reweight existing supervision. R\textsuperscript{2}-OPD weights teacher corrections according to their alignment with task outcomes and the magnitude of teacher--student disagreement~\citep{xu2026r2opd}. Dr.~OPD learns token weights through bilevel optimization to maximize the student's expected task reward~\citep{wang2026dropd}.

\paragraph{Distillation targets from differences across models and contexts.}
Differences in the scores that two policies assign to the same token can provide a distillation signal. G-OPD adjusts this signal through the choice of reference policy and reward scale~\citep{yang2026exopd}. Direct-OPD uses the change in a weak model's token log-probabilities before and after reinforcement learning as a token-level reward~\citep{feng2026directopd}. W2S-OPD uses the corresponding logit differences to construct a proxy teacher by adding them to the logits of the student's base model~\citep{yu2026w2sopd}. Related methods obtain a contrast from a single teacher evaluated under different contexts. CCE-OPD constructs an auxiliary target by comparing the teacher's predictions on the original question and on a version with selected conditions masked~\citep{luo2026cceopd}. Contrastive self-distillation encourages the student to approach the teacher distribution conditioned on a correct solution and move away from the distribution conditioned on an incorrect one~\citep{baumann2026repulsive}. In these approaches, differences between predictions modify the reward or distort the training target.

\paragraph{Identifying and learning from token-level disagreements.}
The work most closely related to ours uses predictive disagreement to isolate critical token decisions. \emph{Demystifying OPD} defines a \emph{fork} as a prefix at which the student and teacher prefer different next tokens, then intervenes at these positions during generation to measure how changing a token affects task performance~\citep{li2026demystifying}. DUET compares two teachers with identical weights, only one of which sees a prohibition in its instructions; their disagreement determines which positions receive supervision and pushes the student toward the constrained teacher and away from the unconstrained one~\citep{li2026duet}. Rather than constructing repulsive targets, FPD uses a frozen foil policy solely to locate decision forks and propose a competing candidate on student-generated prefixes, while training the student to match the strong teacher's exact conditional odds over the pair regardless of which token is sampled.

%% file: sections/preliminary.tex
\providecommand{\E}{\mathbb{E}}
\providecommand{\KL}{\mathrm{KL}}
\providecommand{\sg}{\operatorname{sg}}

\section{Preliminaries}
\label{sec:preliminary}

\paragraph{Setup.}
Let $\mathcal V$ be a finite vocabulary.
Given a prompt $x\sim\mathcal D$, a language model $p_\theta$ generates a response $y=(y_1,\ldots,y_{|y|})$ autoregressively, $p_\theta(y\mid x)=\prod_{t=1}^{|y|}p_\theta(y_t\mid s_t)$, where $s_t=(x,y_{<t})$ is the prefix at step $t$ and $p_\theta(\cdot\mid s)=\mathrm{softmax}(z_\theta(s))$ for logits $z_\theta(s)\in\mathbb R^{|\mathcal V|}$.
In distillation, $p_\theta$ is the student, and a frozen teacher language model $p^+$ with the same tokenizer scores the same prefixes.

\paragraph{On-policy distillation.}
On-policy distillation (OPD) trains the student on prefixes from its own rollouts, so that training matches the prefixes the student meets at inference and avoids the exposure bias of distilling on fixed sequences~\citep{agarwal2024gkd,gu2024minillm}.
Let $p_{\bar\theta}$ be the student that generated the current rollouts, and write $\E_{\bar\theta}$ for the expectation over $x\sim\mathcal D$ and $y\sim p_{\bar\theta}(\cdot\mid x)$.
On each visited prefix, OPD minimizes a divergence between the student and the teacher, such as the forward KL, the reverse KL or the generalized Jensen--Shannon divergence~\citep{agarwal2024gkd}.
We use the reverse KL $K_t(\theta)=\KL\big(p_\theta(\cdot\mid s_t)\,\|\,p^+(\cdot\mid s_t)\big)$, as in~\citep{gu2024minillm,lu2025opd}, which gives the objective
\begin{equation}
\mathcal L_{\mathrm{OPD}}(\theta;\bar\theta)
=\E_{\bar\theta}\bigg[\sum_{t=1}^{|y|}K_t(\theta)\bigg],
\label{eq:opd-baseline}
\end{equation}
where the rollouts, and hence the prefixes, are held fixed when differentiating with respect to $\theta$.

\paragraph{Sampled-token estimator.}
Computing $K_t$ exactly requires the teacher's full distribution over $\mathcal V$ at every visited prefix.
Because $K_t$ is an expectation under the student, large-scale implementations instead query the teacher only at the sampled tokens~\citep{lu2025opd}, through the \emph{distillation advantage}
\begin{equation}
A_t=\log p^+(y_t\mid s_t)-\log p_{\bar\theta}(y_t\mid s_t).
\label{eq:opd-adv}
\end{equation}
Under full-softmax sampling from $p_{\bar\theta}(\cdot\mid s_t)$, $-A_t$ is an unbiased single-sample estimate of the divergence, $\E_{\bar\theta}[-A_t\mid s_t]=K_t(\bar\theta)$, and the student minimizes the surrogate
\begin{equation}
\mathcal L^{\mathrm{sample}}_{\mathrm{OPD}}(\theta;\bar\theta)
=-\E_{\bar\theta}\bigg[\sum_{t=1}^{|y|}\sg[A_t]\,\log p_\theta(y_t\mid s_t)\bigg],
\label{eq:opd-sample}
\end{equation}
where $\sg[\cdot]$ stops gradients.
Under the same full-softmax sampling assumption, the score-function identity gives identical gradients for \eqref{eq:opd-sample} and \eqref{eq:opd-baseline} at $\theta=\bar\theta$, with the prefixes held fixed.
Each term raises the probability of the sampled token when $A_t>0$, that is, when the teacher assigns the token a higher probability than the student does, and lowers it when $A_t<0$.
Practical training uses nucleus-sampled rollouts and clipped surrogates (Appendix~\ref{app:optimization_protocol}).

\paragraph{Pairwise preference models.}
The Bradley--Terry (BT) model~\citep{bradley1952rank} assigns each item $u$ a score $r_u\in\mathbb R$ and models the probability that $u$ is preferred over $v$ as
\begin{equation}
P(u\succ v)=\sigma(r_u-r_v)=\frac{e^{r_u}}{e^{r_u}+e^{r_v}},
\label{eq:bt-model}
\end{equation}
where $\sigma$ is the logistic function.
When the scores are the logits of a softmax distribution, $P(u\succ v)$ is that distribution restricted to $\{u,v\}$ and renormalized, so the softmax normalizer cancels.
Given a target probability $q$, the score difference $\Delta=r_u-r_v$ is fit with the logistic loss
\begin{equation}
\ell_{\mathrm{BT}}(\Delta;q)=-q\log\sigma(\Delta)-(1-q)\log\sigma(-\Delta),
\label{eq:bt}
\end{equation}
the cross-entropy between the target $q$ and the model probability $\sigma(\Delta)$.
Its derivative $\sigma(\Delta)-q$ vanishes exactly when the model probability matches the target.
A soft label $q\in(0,1)$ therefore gives the unique minimizer $\Delta=\log\big(q/(1-q)\big)$.
The hard label $q=1$ used in direct preference optimization (DPO)~\citep{rafailov2023direct} has no finite minimizer: the loss keeps decreasing as $\Delta\to\infty$, which can cause overfitting~\citep{azar2024general}.

%% file: sections/3_method.tex
\section{Fork Preference Distillation}
\label{sec:method}

Fork Preference Distillation (FPD) augments OPD with direct comparisons at \emph{decision forks}, where a strong teacher and a frozen foil policy prefer different next tokens. The foil identifies the competing choice; the strong teacher determines the target odds. We first define this comparison, then present the objective and its multi-teacher extension.

\subsection{Fork Selection and Calibrated Preferences}
\label{sec:method-slow}

\paragraph{Why compare alternatives?}
Fix a student-generated prefix $s$ and a full-support teacher $p^+$. Write $p=p_\theta(\cdot\mid s)$, $K=\KL(p\|p^+)$, and $A(v)=\log[p^+(v)/p(v)]$. The exact reverse-KL descent direction is
\begin{equation}
 -\partial_{z_v}K=p(v)\big(A(v)+K\big).
 \label{eq:opd-push}
\end{equation}
For a fixed positive teacher on a finite vocabulary, this signal is $\epsilon\log(1/\epsilon)+O(\epsilon)$ when $p(v)=\epsilon\to0$, even if $v$ is the teacher's first choice. Reverse KL also weights the conditional discrepancy within a token pair by its total student probability mass (Appendix~\ref{app:stationary}). These observations motivate supervising the relative preference directly, independently of whether either candidate is sampled.

\paragraph{Locating decision forks.}
Let $p^-$ be a frozen foil with the same tokenizer as the student and strong teacher. At each valid response prefix, select
\begin{equation}
 a=\arg\max_v p^+(v\mid s),\qquad
 b=\arg\max_v p^-(v\mid s).
 \label{eq:top-choices}
\end{equation}
A fork occurs when $a\ne b$, using fixed tie-breaking. Both candidates are selected from the full vocabulary, and neither depends on the student's ranking or its sampled token. An earlier checkpoint can supply the foil in single-teacher transfer; existing specialists supply it in the multi-teacher setting.

\paragraph{Matching the teacher's conditional odds.}
On a fork, define the student margin and the two conditional probabilities
\begin{equation}
 M=z_a-z_b,\quad u=\sigma(M),\qquad
 q=\frac{p^+(a)}{p^+(a)+p^+(b)}.
 \label{eq:pair-quantities}
\end{equation}
FPD matches the strong teacher's odds with a soft-label Bradley--Terry loss,
\begin{equation}
 \ell^{\mathrm{fork}}
 =-q\log u-(1-q)\log(1-u).
 \label{eq:fp-fork-loss}
\end{equation}
The frozen teacher alone supplies $q$; the foil supplies only $b$. Since $u=p(a)/(p(a)+p(b))$, the vocabulary normalizer cancels. The descent direction is
\[
 -\nabla_z\ell^{\mathrm{fork}}=(q-u)(e_a-e_b),
\]
where $e_v$ is the coordinate vector for token $v$. This gives direct supervision on both candidates even when the rollout samples a third token. The correction contains no factor of their total mass $p(a)+p(b)$ and vanishes once the conditional odds match. The base OPD term continues to supervise probability mass allocation. Appendix~\ref{app:case-study} walks through this contrast on a concrete mathematical reasoning fork.

\subsection{Single-Teacher FPD}
\label{sec:method-fp}
\label{sec:method-bound}

A fixed strong teacher--foil pair scores the prefixes of freshly sampled student responses. Let $\mathcal R$ contain the valid response positions $(i,t)$ in an optimizer step and $\mathcal F\subseteq\mathcal R$ its forks. With $N=|\mathcal R|>0$ and $F=|\mathcal F|$, we optimize
\begin{equation}
 \mathcal L_{\mathrm{FPD}}
 =\frac1N\sum_{\mathcal R}\ell^{\mathrm{base}}_{it}
 +\frac{\lambda}{\max\{1,F\}}
       \sum_{\mathcal F}\ell^{\mathrm{fork}}_{it}.
 \label{eq:fp-objective}
\end{equation}
The base loss uses the teacher's feedback on each sampled token, with detached advantages from \eqref{eq:opd-adv}. It covers every valid response position, while $\lambda$ weights the mean comparison loss over forks. If $F=0$, the objective reduces to the base loss. Teacher targets and candidate IDs remain fixed during the update. Training uses nucleus-sampled rollouts and clipped base surrogates, detailed in Appendix~\ref{app:optimization_protocol}. The same reduction applies to multi-teacher FPD.

\paragraph{Why the fork correction helps.}
We isolate the effect of the pair term through the exact fixed-prefix objective $J_\lambda(p)=K+\lambda\ell^{\mathrm{fork}}$, corresponding to $N=F=1$. Below we assume free logits and a full-support teacher.

\begin{proposition}[Teacher target preservation, informal]
\label{prop:fp-main-target}
For any fixed distinct pair and $\lambda\ge0$, $p^+$ is the unique interior minimizer and stationary distribution of $J_\lambda$. Its excess objective is
\begin{equation}
 J_\lambda(p)-J_\lambda(p^+)
 =K+\lambda\,\mathrm{kl}(q\|u)\ge0,
 \label{eq:fp-main-target}
\end{equation}
where $\mathrm{kl}$ is Bernoulli KL.
\end{proposition}

\begin{proposition}[Recovery of a rare teacher preference, informal]
\label{prop:fp-main-recovery}
At a binary fork, fix $q\in(1/2,1)$, step size $\eta>0$, and $\lambda>0$. Initialize $p_0(a)=\epsilon\in(0,1/2)$. Let $T_\lambda$ count exact-gradient steps on both logits until $p(a)\ge1/2$. As $\epsilon\to0$,
\begin{equation}
 \begin{aligned}
 T_0&=\Theta\!\left(\frac{1}{\epsilon\log(1/\epsilon)}\right),
 T_\lambda&=O\!\left(\frac{\log(1/\epsilon)}{\lambda}\right).
 \end{aligned}
 \label{eq:fp-main-recovery}
\end{equation}
\end{proposition}

\paragraph{Interpretation.}
In the binary model, direct odds matching accelerates recovery of a rare teacher preference while retaining the ideal target. Propositions~\ref{prop:fp-target} and~\ref{prop:fp-escape} in Appendix~\ref{app:theory} give the formal statements and proofs, with stochastic extensions and practical scope discussed there. Under additional reward-alignment and step-size conditions, this recovery yields simultaneous gains in return and entropy at a matched budget (Corollary~\ref{cor:fp-entropy-return}, Appendix~\ref{app:entropy}).

\subsection{Multi-Teacher FPD}
\label{sec:method-mopd}

Multi-teacher OPD (MOPD) trains one student on several domains with a specialist teacher for each. This roster naturally supplies foils: a teacher that provides strong supervision in its own domain can propose competing continuations in another. FPD therefore extends to MOPD by reusing the existing specialists for both roles.

Let $\{p_d^+\}_{d=1}^D$ be the specialists. Before training, assign each domain $d$ the lowest-scoring remaining specialist on that domain,
\begin{equation}
 \begin{aligned}
 w(d)&=\arg\min_{j\ne d}\operatorname{Acc}_j(d),\\
 (p_i^+,p_i^-)&=(p_{d_i}^+,p_{w(d_i)}^+),
 \end{aligned}
 \label{eq:mopd-weak}
\end{equation}
where $d_i$ is the domain of prompt $i$ and $\operatorname{Acc}_j(d)$ is specialist $j$'s benchmark score on domain $d$. This map remains fixed throughout training. The domain specialist selects $a_{it}$ and scores both candidates to set $q_{it}$; the routed foil selects $b_{it}$. Its probabilities do not enter the target odds. The same specialist can thus teach on its own prompts and supply alternatives on others.

The multi-teacher base evaluates the clipped surrogate on the pre-update student's Top-16 support, using its renormalized reference probabilities as weights in the detached advantages, as specified in \eqref{eq:impl_topk}. Fork selection still uses the full vocabulary, so the pair term can supervise the domain teacher's preferred token even when the base support omits it. We retain \eqref{eq:fp-objective}, pooling valid-token and fork counts over all domains in the optimizer step. A domain with $F_d>0$ forks contributes its mean fork loss with weight $\lambda F_d/F$.

Algorithm~\ref{alg:forkpair-step} in Appendix~\ref{app:training-algorithm} summarizes both settings. Scoring, reduction, and computational costs are detailed in Appendices~\ref{app:implementation} and~\ref{app:complexity}.

%% file: sections/4_experiments.tex
\section{Experiments}
\label{sec:experiments}

\input{tables/3_main_4b_24k}

\subsection{Experimental Setup}
\label{sec:exp-setup}
We initialize the student from Qwen3-1.7B~\citep{yang2025qwen3}. In single-teacher math distillation, the default strong teacher is the non-thinking Qwen3-4B-Math-RL checkpoint (trained with DAPO on DAPO-Math-17k), paired with a frozen foil $p^-$ chosen from Qwen3-4B-Base (B), Qwen3-4B instruct (I), or the initial student (self-foil); the 8B study uses Logics-STEM-8B-RL as $p^+$ and Qwen3-8B (instruct) as $p^-$.
Training uses the no-thinking DAPO-Math-17k mixture~\citep{yu2025dapo} (17{,}183 prompts capped at 1{,}024 tokens, one rollout per prompt) under the 24K protocol (23{,}552 response tokens plus a 2{,}048-token prompt), with batch size $16$, $80$ optimizer steps, checkpoints every $20$ steps, and the clipped importance ratio of~\citet{yu2025dapo} for the base OPD loss.
For single-teacher math evaluation, we report $\mathrm{avg}@32$ on AIME24/25~\citep{maaaime} and AMC23/24~\citep{maaamc}, and $\mathrm{avg}@1$ on MATH500~\citep{hendrycks2021math,lightman2023lets}, MinervaMath~\citep{lewkowycz2022minerva}, and OlympiadBench~\citep{he2024olympiadbench} (24K context, temperature $1.0$, top-$p$ $0.95$, thinking disabled).
For three-domain multi-teacher distillation, we evaluate AIME24 ($\mathrm{avg}@32$), LiveCodeBench~\citep{jain2025livecodebench} (LCB, $\mathrm{avg}@4$), and IFBench~\citep{pyatkin2025generalizing} ($\mathrm{avg}@2$) with a 23{,}552-token response budget.
In both settings, Avg.\ is the unweighted mean across benchmarks; each configuration is trained once, and all reference checkpoints are evaluated under the same protocol. Full training hyperparameters and evaluation details are provided in Appendix~\ref{app:implementation} (Table~\ref{tab:exp_hyperparameters}).

\subsection{Single-Teacher Distillation}
\label{sec:exp-single}

\paragraph{Main results with the 4B teacher.}
Table~\ref{tab:main_4b_24k} compares vanilla OPD, AOPD~\citep{jia2026aopd}, ExOPD~\citep{yang2026exopd}, and FPD across three foil choices (B, I, and self-foil) together with a hard-label ablation at each method's best checkpoint.
All soft-label FPD variants outperform every baseline: FPD reaches 61.7 ($\lambda{=}0.8$, B), 61.4 ($\lambda{=}1.0$, I), and 61.0 ($\lambda{=}1.0$, self-foil), compared with 59.7 for vanilla OPD, 59.5 for AOPD ($K{=}32$, $\tau{=}0$), and 60.5 for ExOPD ($\lambda{=}1.25$).
By contrast, replacing the calibrated soft target $q$ with a hard binary label ($q\equiv 1$, i.e., $\ell^{\mathrm{fork}}_{\mathrm{hard}}=-\log\sigma(M)$; $\lambda{=}1.0$, I, hard label) drops the average to 60.1---below ExOPD (60.5) and 1.3 points below its matched soft-label counterpart ($\lambda{=}1.0$, I, 61.4)---confirming the necessity of soft-label calibration.
Soft-label FPD yields a +2.0-point gain over vanilla OPD at best checkpoints; per-checkpoint curves in Figure~\ref{fig:4b-curves} further show that FPD ($\lambda{=}0.8$, B) leads vanilla OPD at every step, by +7.6 points at step 20 and +2.3--3.0 points at matched step 80, while sustaining higher student entropy (0.40--0.55 versus 0.22--0.32), lower 8-gram repeat coverage (30--35\% versus 38--44\%), and modest rollout truncation (predominantly below 20\%, panels~i--l).
Improvements are broad across competition benchmarks, including AIME25 (34.1 $\rightarrow$ 37.5--37.7), AMC23 (83.4 $\rightarrow$ 85.4), OlympiadBench (62.0 $\rightarrow$ 63.9), MinervaMath (36.0 $\rightarrow$ 39.0), and AIME24 (47.6 $\rightarrow$ 50.1), while MATH500, already near 90, moves slightly.
Decision forks occur at roughly 9\% of response positions, and early in training the student prefers the foil over the strong teacher at 81\% of them, focusing supervision where sampled-token OPD is slowest.

\input{figures/4b_curves}
\input{tables/4_main_8b_24k}

\paragraph{Main results with the 8B teacher.}
Table~\ref{tab:main_8b_24k} repeats the comparison with Logics-STEM-8B-RL ($p^+$) and Qwen3-8B ($p^-$).
Because the 8B teacher's long chain-of-thought generates 19--22K-token AIME responses with 65--80\% truncation under the 24K budget, vanilla OPD learns slowly and peaks at 46.2 (step 80).
ExOPD ($\lambda{=}1.25$) and AOPD ($K{=}32$, $\tau{=}0$) improve over vanilla OPD to 48.9 and 51.1, respectively, whereas FPD ($\lambda{=}0.3$) reaches 55.0 at step 60 (+8.8 over vanilla OPD and +3.9 over AOPD) and achieves the highest score on all seven benchmarks.
Across the full 8B $\lambda$ sweep in Appendix~\ref{app:lambda-all-full} (Table~\ref{tab:lambda_all}), FPD ($\lambda{=}0.5$) reaches 54.6 at step 80 (+8.4 at matched step 80), while larger weights ($\lambda\in\{0.7,1.0,1.5,2.0\}$) remain above vanilla OPD (49.1--52.8) but trail $\lambda\in\{0.3,0.5\}$.
Per-checkpoint curves in Appendix~\ref{app:8b-curves} (Figure~\ref{fig:8b-curves}) confirm that FPD ($\lambda{=}0.3$) leads vanilla OPD, ExOPD, and AOPD across all evaluated checkpoints: AOPD peaks early at step 40 before receding, whereas FPD continues improving through step 60 while maintaining higher student entropy and the lowest 8-gram repeat coverage.
An independent seed-repeat run checkpointed every 5 updates confirms this ordering across all 16 checkpoints (Appendix~\ref{app:8b-seed-repeat}, Figure~\ref{fig:save5-8b}), peaking at 54.9 at step 65 versus 47.4 for vanilla OPD, with vanilla entropy collapsing (0.26 $\rightarrow$ 0.11) and repetition rising (51\% $\rightarrow$ 58\%), whereas FPD sustains entropy in 0.13--0.68 (ending at 0.46) and cuts repetition from 48\% to 30\%.

\subsection{Cross-Domain Multi-Teacher Distillation}
\label{sec:exp-multi}

\input{tables/5_mopd_240}
We next evaluate multi-teacher FPD (Section~\ref{sec:method-mopd}) using three Qwen3-4B specialists---Qwen3-4B-Math-RL, Qwen3-4B-Code-RL, and Qwen3-4B-IF-RL---on 82{,}043 prompts (17{,}183 math, 19{,}963 code, and 44{,}897 IF; Appendix~\ref{app:mopd-configuration}).
Each prompt uses its domain specialist as $p^+$ and the lowest-scoring out-of-domain specialist as $p^-$ (math/code $\rightarrow$ IF, IF $\rightarrow$ code).
All arms train for 240 updates with a 1:1:1 batch of 48 prompts (16 per domain), checkpoints every 20 steps, and the same student Top-16 base surrogate, isolating the fork preference term.
In Table~\ref{tab:mopd_240}, FPD ($\lambda{=}0.3$) achieves the highest three-benchmark average of 33.19 (step 39) versus 28.56 for vanilla MOPD (step 139), leading at all 12 checkpoints (+4.7 at step 19 and +4.6 averaged across the run), with AIME24 peaking at 53.4 and IFBench rising from 19.8 to 28.0. Appendix~\ref{app:mopd-4x4} (Figure~\ref{fig:mopd-4x4}) reports all 12 checkpoints and per-domain rollout diagnostics.
Crucially, vanilla MOPD suffers late-stage code collapse (LCB 15.4 at step 39 $\rightarrow$ 0.6 at step 239), whereas FPD stays stable through step 239 (AIME24 49--53, LCB 14--20, IFBench 25--28).

\subsection{Composability with Adaptive Domain Sampling}
\label{sec:exp-d3}

D3-MOPD~\citep{sun2026d3mopd} adapts only the per-domain sampling mixture using each domain's reverse-KL gap and descent velocity, leaving the token-level objective unchanged (Appendix~\ref{app:d3}).
Pairing FPD ($\lambda{=}0.3$) with D3 under the same cross-domain foils and 240-update budget (bottom two rows of Table~\ref{tab:mopd_240}) lifts the best-checkpoint average from 28.78 (step 99) to 31.98 (step 79), improving all three domains: AIME24 (46.5 $\rightarrow$ 50.6), LCB (13.9 $\rightarrow$ 17.3), and IFBench (26.0 $\rightarrow$ 28.0); across all 12 checkpoints, average gains are +5.5 on AIME24, +7.0 on LCB, and +1.6 on IFBench.
Relative to the fixed-mixture FPD (33.19), FPD + D3 is lower on average and on AIME24/LCB; its IFBench score (28.0) is comparable to the fixed arm's peak (28.0), although the fixed arm's best-average checkpoint has 25.7. Thus FPD also improves over the D3 baseline, but the fixed 1:1:1 mixture remains stronger on this run.
More importantly, while the D3 baseline suffers severe late-stage code collapse (LCB dropping to 0.0 at step 179 and 1.5 at step 239), FPD + D3 sustains 14--20 on LCB through step 239. Appendix~\ref{app:d3-composition} (Figure~\ref{fig:d3-composition}) shows the corresponding target mixtures: the baseline remains spread over code and IF, whereas FPD+D3 shifts most sampling budget to IF.

%% file: tables/3_main_4b_24k.tex
\begin{table*}[!t]
\centering
\small
\setlength{\tabcolsep}{3pt}
\renewcommand{\arraystretch}{1.03}
\caption{Single-teacher distillation from Qwen3-4B-Math-RL to Qwen3-1.7B across seven mathematical reasoning benchmarks under the 24K protocol (best checkpoint per method). In this and subsequent results tables, the best and second-best distilled scores per column are \textbf{bold} and \protect\underline{underlined}, and the {\setlength{\fboxsep}{1.5pt}\protect\colorbox{tblbase!25}{highlighted}} row marks the highest Avg.}
\label{tab:main_4b_24k}
\begin{adjustbox}{max width=\linewidth}
\begin{tabular}{l|c|ccccccc|c}
\toprule
Method & Best step & AIME24 & AIME25 & AMC23 & AMC24 & MATH500 & Minerva & Olympiad & Avg. \\
\midrule
\multicolumn{10}{c}{\textit{Reference models}} \\
\cmidrule(lr){1-10}
Qwen3-4B-Math-RL (teacher) & -- & 62.4 & 54.4 & 93.8 & 72.6 & 93.4 & 45.2 & 71.1 & 70.4 \\
Qwen3-4B (instruct, foil I) & -- & 20.0 & 18.1 & 65.5 & 43.5 & 84.2 & 37.1 & 49.1 & 45.4 \\
Qwen3-4B-Base (foil B) & -- & 2.3 & 0.8 & 11.0 & 7.6 & 18.0 & 6.6 & 10.1 & 8.1 \\
Qwen3-1.7B (student) & -- & 13.5 & 9.0 & 46.8 & 32.0 & 71.4 & 27.9 & 39.2 & 34.3 \\
\midrule
\multicolumn{10}{c}{\textit{Distilled models}} \\
\cmidrule(lr){1-10}
Vanilla OPD & 60 & 47.6 & 34.1 & 83.4 & 65.3 & 89.6 & 36.0 & 62.0 & 59.7 \\
AOPD ($K{=}32$, $\tau{=}0$) & 80 & 45.0 & 34.8 & 82.0 & 65.2 & 91.2 & 36.4 & 61.7 & 59.5 \\
ExOPD ($\lambda{=}1.25$) & 80 & \textbf{50.7} & 35.6 & 83.2 & 66.3 & 91.0 & 34.6 & 62.3 & 60.5 \\
\rowcolor{gray!15}
FPD ($\lambda{=}1.0$, I, hard label) & 59 & 47.0 & \textbf{38.0} & 81.6 & 66.7 & 90.4 & 35.7 & 61.3 & 60.1 \\
\rowcolor{gray!15}
FPD ($\lambda{=}1.0$, self-foil) & 80 & 49.4 & 37.5 & \underline{83.7} & \textbf{67.2} & 89.6 & \underline{37.9} & 61.9 & 61.0 \\
\rowcolor{gray!15}
FPD ($\lambda{=}1.0$, I) & 80 & 48.5 & 37.5 & 83.4 & 66.6 & \textbf{91.6} & \textbf{39.0} & \underline{62.9} & \underline{61.4} \\
\rowcolor{tblbase!25}FPD ($\lambda{=}0.8$, B) & 80 & \underline{50.1} & \underline{37.7} & \textbf{85.4} & \underline{67.1} & \underline{91.4} & 36.4 & \textbf{63.9} & \textbf{61.7} \\

\bottomrule
\end{tabular}
\end{adjustbox}
\end{table*}

%% file: figures/4b_curves.tex
\begin{figure*}[!t]
  \centering
  \includegraphics[width=\linewidth]{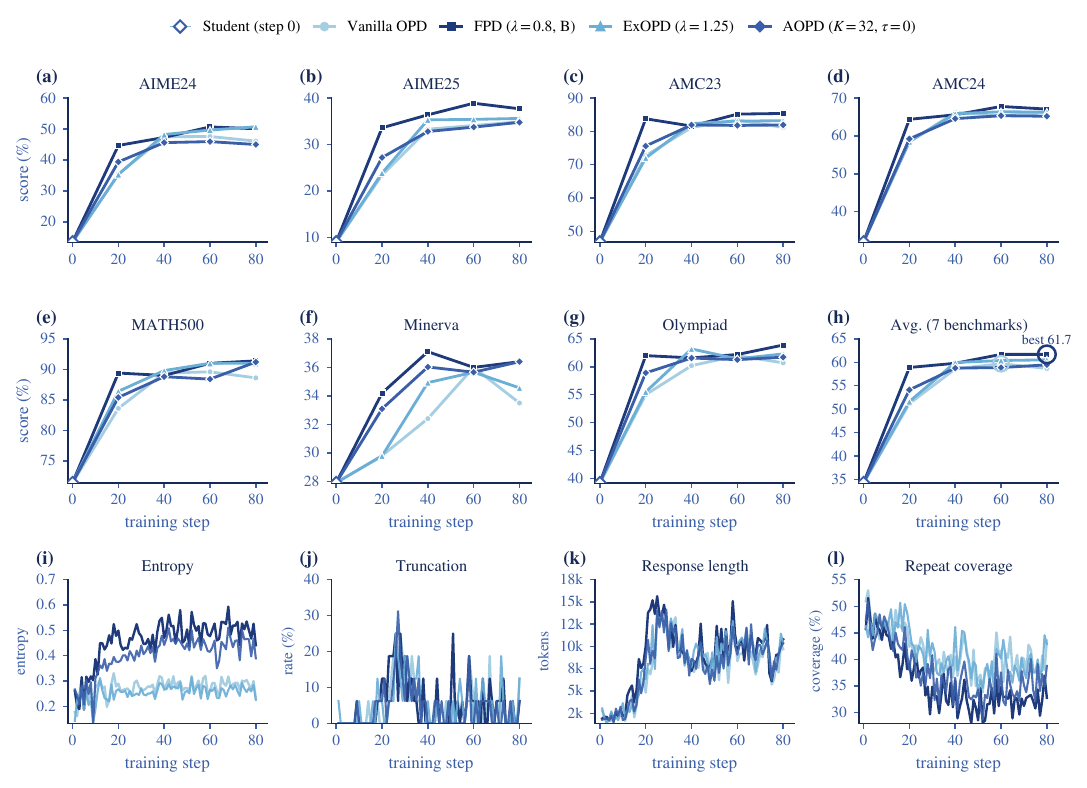}
  \caption{Per-checkpoint accuracy and training-rollout diagnostics for the 4B-teacher runs (checkpoints every 20 updates). Panels~(a)--(g) show individual benchmarks and (h) their unweighted average, with the diamond marking the initial student (step~0). Panels~(i)--(l) track per-update student entropy, rollout truncation rate, mean response length, and 8-gram repeat coverage.}
  \label{fig:4b-curves}
\end{figure*}

%% file: tables/4_main_8b_24k.tex
\begin{table*}[!t]
\centering
\small
\setlength{\tabcolsep}{3pt}
\renewcommand{\arraystretch}{1.03}
\caption{Single-teacher distillation from Logics-STEM-8B-RL to Qwen3-1.7B with Qwen3-8B as the foil policy across seven mathematical reasoning benchmarks under the 24K protocol.}
\label{tab:main_8b_24k}
\begin{adjustbox}{max width=\linewidth}
\begin{tabular}{l|c|ccccccc|c}
\toprule
Method & Best step & AIME24 & AIME25 & AMC23 & AMC24 & MATH500 & Minerva & Olympiad & Avg. \\
\midrule
\multicolumn{10}{c}{\textit{Reference models}} \\
\cmidrule(lr){1-10}
Logics-STEM-8B-RL (teacher) & -- & 60.9 & 48.5 & 88.8 & 70.5 & 89.0 & 44.1 & 67.1 & 67.0 \\
Qwen3-8B (foil) & -- & 25.5 & 19.0 & 68.4 & 46.4 & 83.8 & 36.8 & 50.6 & 47.2 \\
Qwen3-1.7B (student) & -- & 13.5 & 9.0 & 46.8 & 32.0 & 71.4 & 27.9 & 39.2 & 34.3 \\
\midrule
\multicolumn{10}{c}{\textit{Distilled models}} \\
\cmidrule(lr){1-10}
Vanilla OPD (8B) & 80 & 28.0 & 22.2 & 62.1 & 52.8 & 82.6 & 25.4 & 50.1 & 46.2 \\
AOPD ($K{=}32$, $\tau{=}0$) & 40 & \underline{34.9} & \underline{24.5} & \underline{71.5} & \underline{59.0} & 81.6 & \underline{32.7} & \underline{53.4} & \underline{51.1} \\
ExOPD ($\lambda{=}1.25$) & 80 & 30.0 & 23.6 & 67.0 & 55.8 & \underline{83.6} & 30.9 & 51.2 & 48.9 \\
\rowcolor{tblbase!25}FPD ($\lambda{=}0.3$) & 60 & \textbf{38.5} & \textbf{28.2} & \textbf{75.0} & \textbf{62.3} & \textbf{88.2} & \textbf{35.7} & \textbf{56.8} & \textbf{55.0} \\
\bottomrule
\end{tabular}
\end{adjustbox}
\end{table*}

%% file: tables/5_mopd_240.tex
\begin{table*}[!t]
\centering
\footnotesize
\setlength{\tabcolsep}{7pt}
\renewcommand{\arraystretch}{1.1}
\caption{Three-domain multi-teacher distillation from Qwen3-4B math, code, and instruction-following specialists to Qwen3-1.7B over 240 updates under fixed 1:1:1 and adaptive (D3) domain sampling.}
\label{tab:mopd_240}
\begin{tabular}{l|c|ccc|c}
\toprule
Method & Best step & AIME24 & LCB & IFBench & Avg. \\
\midrule
\multicolumn{6}{c}{\textit{Reference models}} \\
\cmidrule(lr){1-6}
Qwen3-4B-Math-RL & -- & 62.1 & 26.2 & 29.0 & 39.1 \\
Qwen3-4B-Code-RL & -- & 30.8 & 29.1 & 27.0 & 29.0 \\
Qwen3-4B-IF-RL & -- & 16.2 & 20.7 & 43.3 & 26.8 \\
Qwen3-1.7B (student) & -- & 12.1 & 13.5 & 18.7 & 14.7 \\
\midrule
\multicolumn{6}{c}{\textit{Fixed 1:1:1 mixture}} \\
\cmidrule(lr){1-6}
Vanilla MOPD & 139 & 45.8 & 13.2 & \underline{26.7} & 28.56 \\
\rowcolor{tblbase!25}FPD ($\lambda{=}0.3$) & 39 & \textbf{53.4} & \textbf{20.5} & 25.7 & \textbf{33.19} \\
\midrule
\multicolumn{6}{c}{\textit{Adaptive domain sampling (D3)}} \\
\cmidrule(lr){1-6}
D3-MOPD baseline & 99 & 46.5 & 13.9 & 26.0 & 28.78 \\
FPD + D3 ($\lambda{=}0.3$) & 79 & \underline{50.6} & \underline{17.3} & \textbf{28.0} & \underline{31.98} \\
\bottomrule
\end{tabular}
\end{table*}

%% file: sections/5_ablation_analysis.tex
\section{Ablations and Analysis}
\label{sec:ablations}

\subsection{Fork Weight and Choice of Foil Policy}
\label{sec:ablation-foil}

Figure~\ref{fig:lambda-all} varies the fork weight $\lambda$ and compares three teacher/foil settings: 4B teacher with Qwen3-4B-Base (B), 4B teacher with Qwen3-4B instruct (I), and 8B teacher with Qwen3-8B; Appendix~\ref{app:lambda-all-full} (Table~\ref{tab:lambda_all}) reports the full seven-benchmark breakdown.
FPD is insensitive to the fork weight in both 4B settings: all 17 tested configurations score between 60.8 and 62.0---within 1.2 points of the 62.0 peak ($\lambda{=}1.5$, B) and consistently above vanilla OPD (59.7). The B-foil curve stays within 0.5 points of the peak across $\lambda\in[0.6,2.0]$ (61.5--62.0), and the I-foil curve remains flat through $\lambda{=}3.0$ (60.9--61.5).
Swapping B for I at matched $\lambda$ shifts the best average by at most 1.0 point, and using the initial student as self-foil ($\lambda{=}1.0$, Table~\ref{tab:main_4b_24k}) reaches 61.0 without an external foil model. By contrast, replacing soft conditional odds with a hard binary label ($q\equiv 1$; $\lambda{=}1.0$, I, Table~\ref{tab:main_4b_24k}) drops the average from 61.4 to 60.1, underscoring the role of calibrated soft targets.
The 8B setting appears more sensitive under the 24K budget: it peaks at $\lambda{=}0.3$ (55.0) and $\lambda{=}0.5$ (54.6), then declines beyond $\lambda{=}0.5$ (52.8 at $\lambda{=}0.7$, down to 49.1 at $\lambda{=}2.0$) while still outperforming 8B vanilla OPD (46.2) across the entire sweep.
This decline stems from an interaction between stronger teacher alignment and the fixed 24K response ceiling rather than instability of the fork objective. Whereas 4B rollouts stay well inside the budget even at large weights (9.2--11.4K mean tokens and 6.0--14.7\% truncation for $\lambda\in[1.0,3.0]$; Table~\ref{tab:timing}), the 8B teacher already generates 19--22K-token AIME responses (65--80\% truncation at 24K). Raising $\lambda$ from 0.3 to 2.0 drives the student closer to this long-chain style, monotonically increasing mean rollout length from 13.2K to 16.5K tokens and training truncation from 31.9\% to 52.4\% (Table~\ref{tab:timing}), so more completions are cut off before producing a final answer; a longer generation budget that accommodates the 8B teacher's full trajectories would mitigate this truncation-induced drop.
In practice, FPD is largely insensitive to $\lambda$ when rollouts fit within the context limit, making $\lambda{=}1.0$ a reliable default. When the length budget is tight relative to the teacher's reasoning chains, lowering $\lambda$ (e.g., to $0.3$--$0.5$) avoids excessive truncation while retaining strong gains.
Per-checkpoint curves in Figure~\ref{fig:4b-curves} and Appendix~\ref{app:8b-curves} (Figure~\ref{fig:8b-curves}) further confirm that FPD leads vanilla OPD at every evaluated checkpoint in both settings.
\input{figures/lambda_all}

The 8B study provides a stronger foil-replacement test: although Qwen3-8B lags far behind Logics-STEM-8B-RL (47.2 vs.\ 67.0), FPD still gains +8.8 points at best checkpoints (+8.4 at matched step 80).
This reflects the asymmetry of FPD: the foil proposes only fork positions and rival tokens, while the strong teacher sets the target odds.

\subsection{Long-Horizon Training Stability}
\label{sec:ablation-stability}

The 240-step multi-teacher run in Table~\ref{tab:mopd_240} and Appendix~\ref{app:mopd-4x4} (Figure~\ref{fig:mopd-4x4}) exposes a late-stage collapse of vanilla OPD absent from the 80-step single-teacher runs: vanilla MOPD's code score falls from 15.4 at step 39 to 0.6 at step 239, whereas FPD stays between 13.9 and 20.5.

The fork preference derivative $\partial_M\ell^{\mathrm{fork}}=\sigma(M)-q$ is bounded in $[-1,1]$ and vanishes once the student matches the teacher's conditional odds, avoiding unbounded sharpening at forks; full-policy entropy and stopping behaviour also reflect interactions with the base OPD objective through shared parameters (see Appendix~\ref{app:entropy} for a theoretical analysis of entropy dynamics).

\subsection{Fork Density and Normalization}
\label{sec:ablation-normalization}

Normalizing the fork loss by the fork count $F=|\mathcal F|$ keeps the effective weight $\lambda$ invariant to batch-level fork density.
Forks cover roughly 9\% of response tokens in single-teacher math distillation and 5\%--11\% in three-domain distillation (highest in math, lowest in IF), so the step-level factor $N/F$ varies naturally between 9 and 22 over the 240-step run without extra hyperparameters.
For $F>0$, scaling per-fork losses by $N/F$ before token averaging recovers \eqref{eq:fp-objective}; batches with $F=0$ apply only the base loss.
Wall-clock training overhead and token throughput measurements for the 4B and 8B runs are reported in Appendix~\ref{app:timing} (Table~\ref{tab:timing}).

%% file: figures/lambda_all.tex
\begin{figure}[!t]
  \centering
  \includegraphics[width=\columnwidth]{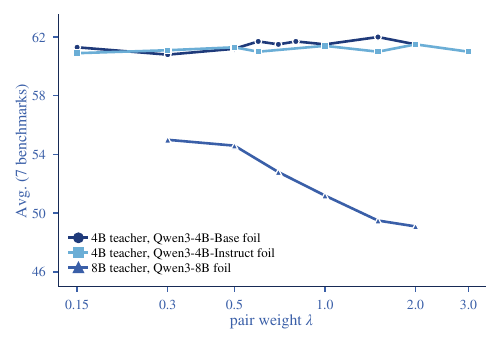}
  \caption{Sensitivity to the fork weight $\lambda$ across three teacher--foil pairs (best-checkpoint seven-benchmark Avg.; full per-benchmark results in Appendix Table~\ref{tab:lambda_all}).}
  \label{fig:lambda-all}
\end{figure}

%% file: sections/6_conclusion.tex
\section{Conclusion}
\label{sec:conclusion}

We introduced Fork Preference Distillation (FPD), a post-training paradigm that unifies trajectory-level on-policy distillation with calibrated pairwise preference learning at teacher-disagreement forks.
In FPD, a frozen foil policy determines only where to intervene and which competing token to contrast against, while the target of every comparison is the strong teacher's exact conditional preference---preserving the strong teacher as the unique stationary distribution for any fork weight and any choice of foil.
Distilling Qwen3-1.7B from 4B and 8B teachers, FPD improves the seven-benchmark average by 3.0 and 8.4 points over vanilla OPD at matched steps, outperforms hard-label supervision, and remains robust across foil checkpoints and moderate fork weights.
In three-domain multi-teacher distillation, reusing cross-domain specialists as mutual foils yields a 4.6-point average gain and sustains stable training over 240 updates, preventing the late-stage entropy and truncation collapse of standard OPD.

%% file: sections/X_appendix.tex
\input{sections/appendix/implements_details}
\input{sections/appendix/algorithm}
\input{sections/appendix/theory}

\input{sections/appendix/entropy}
\input{sections/appendix/additional_experiments}

\input{sections/appendix/complexity}

\input{sections/appendix/case_study}

%% file: sections/appendix/implements_details.tex
\section{Implementation Details}
\label{app:exp_reproducibility}
\label{app:implementation}

\begin{table}[!t]
\centering
\begin{minipage}{0.75\linewidth}
\centering
\caption{Training settings and mathematical evaluation.}
\label{tab:exp_hyperparameters}
\footnotesize
\renewcommand{\arraystretch}{1.02}
\newcommand{\hprow}[2]{\makebox[\linewidth][s]{#1\hfill #2}}
\newcommand{\hpgroup}[1]{\makebox[\linewidth]{\emph{#1}}}
\begin{tabular}{@{}l@{}}
\toprule
\hprow{\textbf{Parameter}}{\textbf{Value}} \\
\midrule
\hpgroup{Optimization} \\
\cmidrule(lr){1-1}
\hprow{Optimizer}{Adam} \\
\hprow{Learning rate}{$1.5\times10^{-6}$} \\
\hprow{Schedule / warmup}{Constant / none} \\
\hprow{Adam $(\beta_1,\beta_2)$}{$(0.9,\,0.999)$} \\
\hprow{Adam $\epsilon$}{$10^{-8}$} \\
\hprow{Weight decay}{$0.01$} \\
\hprow{Global $\ell_2$-norm clip}{$1.0$} \\
\hprow{Master weights / optimizer states}{FP32} \\
\hprow{Actor parallel layout}{TP1 / PP1 / CP4} \\
\midrule
\hpgroup{Distillation loss} \\
\cmidrule(lr){1-1}
\hprow{Method}{Fork Preference Distillation (FPD)} \\
\hprow{Fork weight $\lambda$}{$\{0.15,0.3,0.5,0.6,0.7,0.8,1.0,1.5,2.0,3.0\}$} \\
\hprow{Teacher log-ratio clamp $d_{\max}$}{$10.0$} \\
\hprow{Logistic inverse temperature $\beta$}{$1.0$} \\
\hprow{Logistic coefficient $\eta$ (advantage-shaping baseline)}{$\{0.1,0.15,0.2\}$} \\
\hprow{Linear coefficient $\gamma$ (advantage-shaping baseline)}{$\{0.1,0.15,0.2\}$} \\
\hprow{Importance-ratio clip $(\epsilon_{\mathrm{low}},\epsilon_{\mathrm{high}})$}{$(0.2,\,0.28)$} \\
\hprow{Ratio log-clamp}{$\pm20$} \\
\hprow{Negative-advantage cap}{$3\times$} \\
\hprow{Full-vocabulary KL}{Disabled} \\
\hprow{Soft labels (fork conditional odds)}{Enabled at forks} \\
\hprow{Entropy / reference-KL coefficients}{$0\,/\,0$} \\
\hprow{Reward-derived advantage}{None} \\
\midrule
\hpgroup{Training and rollout} \\
\cmidrule(lr){1-1}
\hprow{Training data}{DAPO-Math-17k} \\
\hprow{Training prompts}{17{,}183} \\
\hprow{Prompts per rollout batch}{16} \\
\hprow{Responses per prompt}{1} \\
\hprow{Optimizer batch (responses)}{16} \\
\hprow{Micro-batch size}{1} \\
\hprow{Prompt length budget (filtered $\le$ 1{,}024)}{2{,}048 tokens} \\
\hprow{Maximum response length}{23{,}552 tokens} \\
\hprow{Total context length (24K protocol)}{25{,}600 tokens} \\
\hprow{Rollout temperature}{$1.0$} \\
\hprow{Rollout top-$p$ / top-$k$}{$0.95$ / disabled} \\
\hprow{Thinking}{Disabled} \\
\hprow{Optimizer steps}{80} \\
\hprow{Checkpoint interval}{20 steps} \\
\bottomrule
\end{tabular}
\end{minipage}
\end{table}

\subsection{Training Configuration}
\label{app:exp_configuration}

All reported single-teacher experiments focus on mathematical reasoning. The student is initialized from Qwen3-1.7B~\citep{yang2025qwen3}; the default strong teacher is the public non-thinking Qwen3-4B-Math-RL checkpoint trained with DAPO on DAPO-Math-17k, and the default foil policy is the frozen Qwen3-4B-Base checkpoint (checkpoint URLs for all single-teacher and multi-teacher models are listed in Table~\ref{tab:model_checkpoints}). Both frozen policies are shared across runs and are never updated. We use the no-thinking DAPO-Math-17k mixture~\citep{yu2025dapo}; after filtering, it contains 17{,}183 prompts with at most 1{,}024 tokens, and each prompt contributes one rollout per optimizer step.

\paragraph{Hardware and software.} Each job uses one node with eight NVIDIA H100-80GB GPUs. Four GPUs host the Megatron actor with tensor parallelism 1, pipeline parallelism 1, and context parallelism 4; the other four GPUs host one SGLang rollout engine each. The teacher caches are constructed on the same node before training. The software environment uses the \texttt{slime:0724} container image with Python~3.12.3, PyTorch~2.11.0 (CUDA~12.9), NCCL~2.26.2, Ray~2.55.1, SGLang~0.5.12.post1, Transformers~5.6.0, and FlashAttention~2.7.4.post1.

\subsection{Teacher and Foil Checkpoints}
\label{app:model_checkpoints}

Table~\ref{tab:model_checkpoints} lists the public repositories for all teacher and foil checkpoints used across the single-teacher (4B and 8B) and three-domain multi-teacher (MOPD) experiments.

\begin{table}[!htbp]
\centering
\footnotesize
\setlength{\tabcolsep}{5pt}
\renewcommand{\arraystretch}{1.05}
\caption{Public repositories for all teacher and foil checkpoints used in our experiments.}
\label{tab:model_checkpoints}
\begin{tabularx}{\linewidth}{@{}llX@{}}
\toprule
Model & Role & Repository URL(s) \\
\midrule
\multicolumn{3}{@{}c}{\textit{Single-teacher distillation (4B)}} \\
\cmidrule(lr){1-3}
Qwen3-4B-Math-RL & Strong teacher & \url{https://modelscope.cn/models/acehxu/Qwen3-4B_Math_teacher_model_dapo17k_nothinking} \\
Qwen3-4B & Foil (I) & \url{https://huggingface.co/Qwen/Qwen3-4B} \\
Qwen3-4B-Base & Foil (B) & \url{https://huggingface.co/Qwen/Qwen3-4B-Base} \\
\midrule
\multicolumn{3}{@{}c}{\textit{Single-teacher distillation (8B)}} \\
\cmidrule(lr){1-3}
Logics-STEM-8B-RL & Strong teacher & \url{https://huggingface.co/Logics-MLLM/Logics-STEM-8B-RL} \\
Qwen3-8B & Foil & \url{https://huggingface.co/Qwen/Qwen3-8B} \\
\midrule
\multicolumn{3}{@{}c}{\textit{Multi-teacher distillation (MOPD)}} \\
\cmidrule(lr){1-3}
Qwen3-4B-Math-RL & Math specialist & \url{https://huggingface.co/pb09204048/Qwen3-4B-Math-RL-Specialist} \\
Qwen3-4B-Code-RL & Code specialist & \url{https://huggingface.co/pb09204048/Qwen3-4B-Code-RL-Specialist} \\
Qwen3-4B-IF-RL & IF specialist & \url{https://huggingface.co/pb09204048/Qwen3-4B-IF-RL-Specialist} \\
\bottomrule
\end{tabularx}
\end{table}

\subsection{Policy Loss and Optimization Protocol}
\label{app:optimization_protocol}

\paragraph{Policy loss and reduction.} For a batch $\mathcal B$ of $N$ responses, let $m_{i,t}$ mark valid response tokens and let $o_{i,t}$ be the sampled token at position $t$. We write $\ell^{+}_{i,t}=\log\pi^{+}(o_{i,t}\mid s_{i,t})$, $\ell^{-}_{i,t}=\log\pi^{-}(o_{i,t}\mid s_{i,t})$, $\ell^{\theta}_{i,t}=\log\pi_{\theta}(o_{i,t}\mid s_{i,t})$, and $\ell^{\bar\theta}_{i,t}=\log\pi_{\bar\theta}(o_{i,t}\mid s_{i,t})$ for the frozen strong teacher, the frozen foil policy, the student, and the rollout behaviour policy, respectively. Define
\begin{equation}
d_{i,t}
=\operatorname{clip}\!\left(\ell^{+}_{i,t}-\ell^{-}_{i,t},
-d_{\max},d_{\max}\right).
\end{equation}
For the base OPD update (and sampled-token advantage-shaping baselines), the per-token advantage is
\begin{equation}
A_{i,t}=
\begin{cases}
\ell^{+}_{i,t}-\ell^{\bar\theta}_{i,t}, & \text{vanilla OPD / FPD base term},\\[2pt]
\ell^{+}_{i,t}-\ell^{\bar\theta}_{i,t}
+\eta\log\sigma\!\left(\beta d_{i,t}\right), & \text{logistic advantage shaping},\\[2pt]
\ell^{+}_{i,t}-\ell^{\bar\theta}_{i,t}
-\gamma\left(\ell^{-}_{i,t}-\ell^{\bar\theta}_{i,t}\right),
& \text{linear advantage shaping}.
\end{cases}
\end{equation}
All teacher and behaviour-policy terms are detached. With $\rho_{i,t}(\theta)=\exp(\operatorname{clip}(\ell^{\theta}_{i,t}-\ell^{\bar\theta}_{i,t},-20,20))$, the implemented per-token loss is
\begin{equation}
\begin{aligned}
\ell_{\mathrm{clip}}(\rho,A)
&=-\min\!\left\{\rho A,
\operatorname{clip}(\rho,1-\epsilon_{\mathrm{low}},
1+\epsilon_{\mathrm{high}})A\right\},\\
\ell_{i,t}
&=
\begin{cases}
\min\!\left(-3A_{i,t},\ell_{\mathrm{clip}}(\rho_{i,t},A_{i,t})\right),
& A_{i,t}<0,\\
\ell_{\mathrm{clip}}(\rho_{i,t},A_{i,t}), & A_{i,t}\ge 0.
\end{cases}
\end{aligned}
\label{eq:impl_actual_policy_loss}
\end{equation}
The base objective is the token-level mean
\begin{equation}
\widehat{\mathcal L}_{\mathrm{impl}}(\theta;\mathcal B)
=\frac{\sum_{i,t}m_{i,t}\,\ell_{i,t}}
{\sum_{i,t}m_{i,t}},
\label{eq:impl_token_mean}
\end{equation}
where gradients are accumulated across token-packed micro-batches before one optimizer step. This is a token-level reduction, not a per-response mean. The single mathematical domain has weight one. Masks include generated response tokens and a terminal EOS when present, exclude prompts and padding, and retain truncated responses; empty masks contribute zero. No advantage whitening, centering, coefficient clipping, or weight cap is applied. Truncated responses retain their standard coefficients, and pure distillation does not use a reward-derived outcome advantage.

\paragraph{Rollout and update schedule.} Each fresh rollout batch forms one optimizer batch, with gradients accumulated across token-packed micro-batches before a single optimizer step, without replay or additional optimization epochs. The student weights are synchronized before sampling and remain unchanged until the update. The trainer uses the rollout engine's log-probabilities as the detached behaviour-policy snapshot for both $A_{i,t}$ and the ratio denominator. Teacher scores and reweighting coefficients remain fixed throughout accumulation. When the trainer and engine log-probabilities agree, $\rho_{i,t}(\bar\theta)=1$, clipping is inactive, and the policy-loss gradient reduces to the detached log-probability surrogate with the same token-level reduction.

\subsection{Evaluation and Reporting}
\label{app:exp_evaluation}

Using the benchmark-specific sample count $k_{\mathcal E}$ and the 24K sampling settings in Table~\ref{tab:exp_hyperparameters}, we report
\begin{equation}
\mathrm{Acc}^{(s)}(\mathcal E)
=\frac{100}{|\mathcal E|k_{\mathcal E}}
\sum_{q\in\mathcal E}\sum_{j=1}^{k_{\mathcal E}}
R\!\left(q,o_{q,j}^{(s)}\right).
\label{eq:exp_avgk}
\end{equation}
Here $k_{\mathcal E}=32$ ($\mathrm{avg}@32$, with $\mathrm{pass}@32$ also logged) for AIME24/25 and AMC23/24, and $k_{\mathcal E}=1$ ($\mathrm{avg}@1$, equivalent to $\mathrm{pass}@1$) for MATH500, MinervaMath, and OlympiadBench. All generations use the 24K protocol ($23{,}552$ maximum response tokens plus a $2{,}048$-token prompt), temperature $1.0$, top-$p$ $0.95$, top-$k$ disabled, and thinking disabled. We evaluate checkpoints at steps $20/40/60/80$.

Given the computational budget, each configuration is trained once; we report single-run results and do not compute a cross-seed mean or standard deviation. Reference checkpoints are evaluated once under the same benchmark-specific protocol. The seven-benchmark macro-average is the unweighted mean of the seven accuracies within each run.
\subsection{Multi-Teacher Configuration}
\label{app:mopd-configuration}

\paragraph{Training data.} The multi-teacher runs draw from a 1:1:1 mixture of 82{,}043 prompts: 17{,}183 math prompts from DAPO-Math-17k~\citep{yu2025dapo}, 19{,}963 code prompts (12{,}886 PrimeIntellect and 7{,}077 TACO~\citep{li2023taco}), and 44{,}897 instruction-following prompts from Nemotron-IF-RL. All prompts are capped at 1{,}024 tokens. Each 48-example rollout batch contains 16 prompts per domain.

\paragraph{Teachers and foils.} The strong teachers are the released non-thinking Qwen3-4B-Math-RL, Qwen3-4B-Code-RL, and Qwen3-4B-IF-RL specialist checkpoints (Table~\ref{tab:model_checkpoints}). The foil for a domain is the out-of-domain specialist with the lowest reference score on that domain: IF for math (AIME24: 16.2 versus 30.8 for code) and for code (LCB: 20.7 versus 26.2 for math), and code for instruction following (IFBench: 27.0 versus 29.0 for math). This mapping is fixed before training.

\paragraph{Optimization and rollout.} All multi-teacher arms use Adam ($\beta_1{=}0.9$, $\beta_2{=}0.999$, $\epsilon{=}10^{-8}$) with a constant learning rate of $1.5\times10^{-6}$, weight decay $0.01$, and global gradient-norm clipping $1.0$. Each rollout/global batch contains 48 responses (16 per domain), and training runs for 240 optimizer updates with a checkpoint every 20 updates. The base term uses the pre-update student's Top-16 support, and the fork term uses $\lambda{=}0.3$. Rollouts use the 24K protocol (1{,}024 prompt tokens, 23{,}552 response tokens, 24{,}576 total context), temperature $1.0$, top-$p$ $0.95$, top-$k$ disabled, and thinking disabled. Each configuration is trained once with rollout seed 20260918 on one node with eight H100-80GB GPUs.

\paragraph{Multi-teacher Top-$K$ base loss.}
Let $S_{i,t}$ be the pre-update student's Top-$K$ tokens at prefix $s_{i,t}$, with $K=16$, and let $p_i^+$ be the routed domain teacher. Write $C(\rho,A)$ for the clipped per-candidate loss in \eqref{eq:impl_actual_policy_loss}. The reference probabilities are renormalized over $S_{i,t}$ and used as weights in the detached advantages:
\begin{equation}
\begin{aligned}
 \omega_{i,t}(v)
 &=\frac{p_{\bar\theta}(v\mid s_{i,t})}
          {\sum_{u\in S_{i,t}}p_{\bar\theta}(u\mid s_{i,t})},\\
 A^{(K)}_{i,t}(v)
 &=\omega_{i,t}(v)
   \left[\log p_i^+(v\mid s_{i,t})
              -\log p_{\bar\theta}(v\mid s_{i,t})\right],\\
 \ell^{\mathrm{base}}_{i,t}
 &=\sum_{v\in S_{i,t}}
      C\!\left(\rho_{i,t}(v),A^{(K)}_{i,t}(v)\right).
\end{aligned}
\label{eq:impl_topk}
\end{equation}
Here $\rho_{i,t}(v)$ uses the same clamped log-ratio as above, evaluated for candidate $v$. Student and teacher log-probabilities are normalized over the full vocabulary; the support and reference scores come from the pre-update actor and remain fixed during the update. Both fork candidates are selected over the full vocabulary and can lie outside $S_{i,t}$. The full FPD loss adds the mean fork preference term to this base loss with the reductions in \eqref{eq:fp-objective}, pooling counts across domains.

\subsection{D3 Adaptive Sampler}
\label{app:d3}

D3-MOPD~\citep{sun2026d3mopd} changes only the per-domain sampling probabilities. After each completed rollout, it reconstructs the full-vocabulary reverse KL for every domain. An exponential moving average (span 10) tracks each domain's KL trajectory. The scheduler forms a remaining-gap term, the current KL divided by the average KL over the first five rollouts (with a floor of 0.15), and a descent-velocity term, the average relative KL decrease over up to three windows of ten rollouts. Their product is normalized across domains, passed through a softmax with temperature 0.5, and mixed with a uniform floor of 0.1; proposals are refreshed every ten rollouts. Largest-remainder rounding converts the resulting probabilities into per-batch quotas, with multiplicative jitter 0.3 to preserve batch-level variation. Both D3 arms use the same loss, batch size, and 240-update budget as the fixed-mixture arms; only the domain mixture changes.

%% file: sections/appendix/algorithm.tex
\section{Training Algorithm}
\label{app:training-algorithm}

Algorithm~\ref{alg:forkpair-step} presents one optimizer update for single-teacher and multi-teacher FPD. Both instantiations generate student rollouts, identify decision forks on the visited prefixes, and optimize the unified objective $\mathcal L_{\mathrm{FPD}}$. They differ in policy routing and the support used by the base loss: single-teacher FPD uses a fixed teacher--foil pair and the sampled-token estimator, whereas multi-teacher FPD routes each prompt to its domain specialist and a preselected cross-domain foil specialist, using the reference student's Top-$K$ support for the base loss. Fork candidates are selected from the full-vocabulary predictions of the strong teacher and foil policy in both settings.

\begin{algorithm}[!htbp]
\caption{Fork Preference Distillation (FPD)}
\label{alg:forkpair-step}
\small
\begin{algorithmic}[1]
 \REQUIRE Student $p_\theta$, prompt batch $\{x_i\}$, frozen policies, fork weight $\lambda$, support size $K$, and clipping thresholds $\epsilon_{\mathrm{low}},\epsilon_{\mathrm{high}}$; domain labels $d_i$ and fixed cross-domain foil map $w$ for multi-teacher FPD.
 \STATE Freeze $\bar\theta\gets\theta$ and generate one response $y_i$ per prompt by nucleus sampling.
 \STATE Let $\mathcal R_i$ contain the valid response positions, set $s_{it}=(x_i,y_{i,<t})$, and initialize $\mathcal F\gets\varnothing$.
 \FOR{each response $i$}
   \IF{single-teacher FPD}
     \STATE Set $(p_i^+,p_i^-)\gets(p^+,p^-)$ and compute clipped sampled-token base losses $\ell^{\mathrm{base}}_{it}$ on $\mathcal R_i$.
   \ELSIF{multi-teacher FPD}
     \STATE Set $(p_i^+,p_i^-)\gets(p^+_{d_i},p^+_{w(d_i)})$.
     \STATE Compute clipped base losses $\ell^{\mathrm{base}}_{it}$ on the reference student's Top-$K$ support, weighting the advantages by its renormalized reference probabilities.
   \ENDIF
   \FOR{each valid position $t\in\mathcal R_i$}
     \STATE Select $a_{it}\gets\arg\max_v p_i^+(v\mid s_{it})$ and $b_{it}\gets\arg\max_v p_i^-(v\mid s_{it})$.
     \IF{$a_{it}\ne b_{it}$}
       \STATE Add $(i,t)$ to $\mathcal F$ and set $q_{it}\gets\sg\!\left[\frac{p_i^+(a_{it}\mid s_{it})}{p_i^+(a_{it}\mid s_{it})+p_i^+(b_{it}\mid s_{it})}\right]$.
       \STATE Set $M_{it}\gets z_{\theta,a_{it}}(s_{it})-z_{\theta,b_{it}}(s_{it})$.
       \STATE Compute $\ell^{\mathrm{fork}}_{it}\gets-q_{it}\log\sigma(M_{it})-(1-q_{it})\log\sigma(-M_{it})$.
     \ENDIF
   \ENDFOR
 \ENDFOR
 \STATE Pool $\mathcal R\gets\{(i,t):t\in\mathcal R_i\}$, with $N\gets|\mathcal R|>0$ and $F\gets|\mathcal F|$.
 \STATE Form $\mathcal L_{\mathrm{FPD}}\gets\frac{1}{N}\sum_{\mathcal R}\ell^{\mathrm{base}}_{it}
       +\frac{\lambda}{\max\{1,F\}}\sum_{\mathcal F}\ell^{\mathrm{fork}}_{it}$.
 \STATE Take one optimizer step on $\theta$ using $\nabla_\theta\mathcal L_{\mathrm{FPD}}$.
\end{algorithmic}
\end{algorithm}

The reference scores, support sets, advantages, and teacher targets remain fixed during the update. The foil policy supplies the competing token, while the strong teacher determines the target conditional odds. Thus the fork preference update remains available even when the rollout samples neither candidate.

Counts are pooled over the batch (across all domains in multi-teacher FPD), averaging the base loss over $N$ tokens and the fork loss over $F$ forks so $\lambda$ stays invariant to fork frequency; when $F=0$, the objective reduces to the base update.

%% file: sections/appendix/theory.tex
\section{Analysis of Fork Preference Distillation}
\label{app:theory}
\label{sec:method-props}

We distinguish exact-gradient dynamics, stochastic sampled-token updates, and task returns under complete responses. Unless stated otherwise, we fix a prefix, a finite vocabulary $\mathcal V$, a full-support teacher $p^+$, and distinct fork candidates $a,b$. The student is $p=\operatorname{softmax}(z)$ with freely adjustable logits. The candidates and teacher scores remain fixed while differentiating. Sections~\ref{app:convergence} and~\ref{app:sampled-dynamics} additionally assume a binary vocabulary. Section~\ref{app:task-return} analyzes complete responses, and Sections~\ref{app:full-vocabulary} and~\ref{app:parameterization} explain the scope of the fixed-prefix guarantees. All logarithms are natural. Entropy dynamics and implications for subsequent reinforcement learning are developed separately in Appendix~\ref{app:entropy}.

\subsection{Logit Gradients at a Fixed Prefix}
\label{app:gradients}

\paragraph{Sampled-token OPD.}
Let $K(p)=\KL(p\|p^+)$ and let $A(y)=\log p^+(y)-\log p(y)$ at the sampling parameters. Since $\log p(y)=z_y-\log\sum_w e^{z_w}$, its derivative with respect to $z_v$ is $\mathbf 1[y=v]-p(v)$. Holding the sampled advantage fixed gives the descent direction of $-\sg[A(y)]\log p(y)$,
\begin{equation}
 g_y(v)=A(y)\big(\mathbf 1[y=v]-p(v)\big).
 \label{eq:opd-logit}
\end{equation}
When we average over the sampled token, the indicator selects one term from the first sum. The second sum is the negative reverse KL, which gives
\begin{equation}
 \begin{aligned}
 \E_{y\sim p} g_y(v)
 &=\sum_y p(y)A(y)\mathbf 1[y=v]-p(v)\sum_y p(y)A(y)\\
 &=p(v)A(v)-p(v)\sum_y p(y)\log\frac{p^+(y)}{p(y)}\\
 &=p(v)\big(A(v)+K(p)\big).
 \end{aligned}
 \label{eq:score-fork-expectation}
\end{equation}
To recover the same direction from the exact objective, write $\ell_y=\log[p(y)/p^+(y)]$. The softmax derivative is $\partial_{z_v}p(y)=p(y)(\mathbf 1[y=v]-p(v))$. Hence
\begin{align*}
 \partial_{z_v}K
 &=\sum_y \partial_{z_v}p(y)(\ell_y+1)\\
 &=\sum_y p(y)(\mathbf 1[y=v]-p(v))\,\ell_y\\
 &=p(v)\ell_v-p(v)\sum_y p(y)\ell_y
 =-p(v)\big(A(v)+K(p)\big).
\end{align*}
The second line uses $\sum_y\partial_{z_v}p(y)=\partial_{z_v}1=0$. Negating the result proves \eqref{eq:opd-push}.

For the rare-token limit, set $p(v)=\epsilon$ and $p^+_{\min}=\min_w p^+(w)>0$. Since $\sum_y p(y)\log p(y)\le0$, we have $0\le K(p)\le-\sum_y p(y)\log p^+(y)\le\log(1/p^+_{\min})$. Substitution in \eqref{eq:score-fork-expectation} gives
\[
 -\partial_{z_v}K
 =\epsilon\log(1/\epsilon)
  +\epsilon\big[\log p^+(v)+K(p)\big]
 =\epsilon\log(1/\epsilon)+O(\epsilon).
\]
The bracket stays bounded even if the student redistributes all its remaining mass as $\epsilon\to0$.

\begin{lemma}[Conditional sampling noise on a rare token]
\label{lem:fp-variance}
Let $p(a)=\epsilon\to0$, with the teacher and vocabulary fixed, and draw $Y\sim p$ from the full softmax. For the descent coordinate $g_Y(a)$ in \eqref{eq:opd-logit},
\begin{align}
 \operatorname{Var}[g_Y(a)]
 &\sim \epsilon\log^2(1/\epsilon),\nonumber\\
 \frac{\operatorname{Var}[g_Y(a)]}{(\E[g_Y(a)])^2}
 &\sim \frac1\epsilon .
 \label{eq:fp-rare-variance}
\end{align}
These limits hold for arbitrary choices of the other student probabilities.
\end{lemma}
\begin{proof}
Write $L_\epsilon=\log(1/\epsilon)$ and $A_a=L_\epsilon+\log p^+(a)$. Separating the events $Y=a$ and $Y\ne a$ gives
\begin{align*}
 \E[g_Y(a)^2]
 &=\epsilon(1-\epsilon)^2A_a^2
   +\epsilon^2\sum_{y\ne a}p(y)A(y)^2.
\end{align*}
The second sum is uniformly bounded. Indeed, with $L_+=\log(1/p^+_{\min})$, the inequality $(x-y)^2\le2x^2+2y^2$ gives
\begin{align*}
 \sum_y p(y)A(y)^2
 &\le 2\sum_y p(y)\log^2 p(y)+2L_+^2\\
 &\le \frac{8|\mathcal V|}{e^2}+2L_+^2,
\end{align*}
where the last step uses $\sup_{0<x\le1}x\log^2x=4/e^2$. Consequently,
\[
 \E[g_Y(a)^2]
 =\epsilon L_\epsilon^2(1+o(1))+O(\epsilon^2).
\]
Equation~\eqref{eq:score-fork-expectation} gives $\E[g_Y(a)]=\epsilon L_\epsilon+O(\epsilon)$. Subtracting its square proves the variance equivalent, and division proves the relative-variance equivalent.
\end{proof}

For $B$ independent draws at the same prefix, the mean estimator has relative variance asymptotic to $1/(B\epsilon)$. The fork preference branch, conditional on this prefix and its frozen candidates, does not depend on the sampled token. Adding it leaves the base branch's absolute sampling variance unchanged but can increase the mean signal when $|q-u|$ stays bounded away from zero. Randomness in the visited prefixes remains.

\paragraph{Fork preference loss.}
Let $a$ and $b$ be the distinct fork candidates and $q=p^+(a)/(p^+(a)+p^+(b))$. Conditioning the student on the pair cancels the vocabulary normalizer, giving $u=e^{z_a}/(e^{z_a}+e^{z_b})=\sigma(M)$ for $M=z_a-z_b$. The soft-target cross-entropy is $\ell_{\mathrm{BT}}(M;q)=-q\log u-(1-q)\log(1-u)$. Differentiating first in $u$ and then in $M$ gives
\[
 \frac{\partial\ell_{\mathrm{BT}}}{\partial M}
 =\left(-\frac{q}{u}+\frac{1-q}{1-u}\right)u(1-u)
 =-q(1-u)+(1-q)u=u-q .
\]
Multiplication by $\partial_{z_v}M=\mathbf 1[v=a]-\mathbf 1[v=b]$ yields
\begin{equation}
 -\frac{\partial\ell_{\mathrm{BT}}(M;q)}{\partial z_v}
 =(q-u)\big(\mathbf 1[v=a]-\mathbf 1[v=b]\big).
 \label{eq:pair-grad}
\end{equation}
The two coordinates have opposite signs, so the pair contribution to the gradient of the logit difference is determined entirely by $u$ and $q$. In particular, for one gradient step of size $\eta$ on $\alpha\ell_{\mathrm{BT}}$, the margin increases by $2\eta\alpha(q-u)$. This increment remains at least $2\eta\alpha(q-1/2)$ when $u\le 1/2<q$, even in a larger vocabulary where $p(a)+p(b)$ is small. For the combined objective, the reverse-KL gradient supplies an additional, vocabulary-dependent contribution.

To make that dependence explicit, subtract the descent direction on $z_b$ in \eqref{eq:score-fork-expectation} from that on $z_a$. The base contribution to the margin per unit step size is
\begin{equation}
 p(a)\big(A(a)+K(p)\big)-p(b)\big(A(b)+K(p)\big).
 \label{eq:full-vocab-margin}
\end{equation}
This expression can vary with the rest of the distribution through $K(p)$ as well as through the individual probabilities of $a$ and $b$. The binary analysis below fixes this dependence by taking $p(a)+p(b)=1$.

\paragraph{Bounded margin derivatives and finite target odds.}
The fork preference loss can also be written as
\[
 \ell_{\mathrm{BT}}(M;q)=\log(1+e^M)-qM.
\]
Its first and second derivatives satisfy
\[
 |\partial_M\ell_{\mathrm{BT}}|\le1,\qquad
 0<\partial_M^2\ell_{\mathrm{BT}}
   =\sigma(M)(1-\sigma(M))\le\tfrac14.
\]
For a soft target $q\in(0,1)$, its unique minimizing margin is $M^+=\log(q/(1-q))$. The loss itself is unbounded: it is asymptotic to $(1-q)M$ as $M\to+\infty$ and to $-qM$ as $M\to-\infty$. These properties control the local margin correction; they do not imply a lower bound on the entropy of the full policy or stability of shared-parameter training.

\subsection{Stationary Distribution under the Exact Objective}
\label{app:stationary}

Consider the ideal conditional objective
\begin{equation}
 J_\alpha(p)=\KL(p\|p^+)+\alpha\,\ell_{\mathrm{BT}}(M;q),
 \qquad \alpha\ge 0 .
 \label{eq:fp-conditional}
\end{equation}
Here $\alpha$ is the coefficient relative to one unit of reverse KL at this prefix. For an optimizer step consisting of one valid token that is a fork, the two reductions in \eqref{eq:fp-objective} give $\alpha=\lambda$. With $N$ valid tokens and $F>0$ forks, factoring out the base weight $1/N$ gives the relative coefficient $\alpha=\lambda N/F$ at each fork. This correspondence describes the reduction weights. Under an ideal SGD update of independent logit blocks, a training step size $\eta_{\mathrm{train}}$ would also become $\eta_{\mathrm{train}}/N$ after this factorization. Both counts are fixed within the update but may vary across batches. The results below hold their local coefficients and step sizes fixed. The base in \eqref{eq:fp-conditional} is the exact conditional KL.

Proposition~\ref{prop:fp-main-target} specializes the following statement to the isolated-fork coefficient $\alpha=\lambda$.

\begin{proposition}[Preservation of the conditional target]
\label{prop:fp-target}
For any fixed distinct pair and any $\alpha\ge 0$, $p^+$ uniquely minimizes $J_\alpha$ on the interior of the simplex and is its only stationary distribution under free logits.
\end{proposition}
\begin{proof}
Let $H(q)=-q\log q-(1-q)\log(1-q)$. The pair loss is
\[
 \ell_{\mathrm{BT}}(M;q)
 =H(q)+\KL\!\big(\operatorname{Bern}(q)\,\big\|\,\operatorname{Bern}(u)\big).
\]
Thus $J_\alpha(p)-\alpha H(q)$ is the sum of two nonnegative terms and is zero at $p^+$. Equality requires $\KL(p\|p^+)=0$, which establishes uniqueness of the minimizer. The pair loss has value $H(q)$ at that point.

To exclude other stationary points, examine the segment $p_\tau=(1-\tau)p+\tau p^+$ from an interior point $p$ toward $p^+$. Let $m=p(a)+p(b)$ and $m^+=p^+(a)+p^+(b)$. The pair mass along the segment is $m_\tau=(1-\tau)m+\tau m^+$. Subtracting the initial conditional probability before differentiating gives
\begin{align*}
 u_\tau-u
 &=\frac{(1-\tau)mu+\tau m^+q-u[(1-\tau)m+\tau m^+]}{m_\tau}\\
 &=\frac{\tau m^+(q-u)}{m_\tau}.
\end{align*}
Since $m_0=m$, dividing by $\tau$ and taking the limit at zero gives $u'_0=(m^+/m)(q-u)$. Multiplying by the cross-entropy derivative $(u-q)/(u(1-u))$ gives the pair loss's directional derivative,
\[
 \left.\frac{d}{d\tau}\ell_{\mathrm{BT}}
      \!\left(\log\frac{u_\tau}{1-u_\tau};q\right)\right|_{\tau=0}
 = -\frac{m^+}{m}\frac{(u-q)^2}{u(1-u)}\le 0 .
\]
For the reverse KL, the derivative of $p_\tau(v)$ is $p^+(v)-p(v)$, which sums to zero. Consequently
\begin{align*}
 \left.\frac{d}{d\tau}K(p_\tau)\right|_{\tau=0}
 &=\sum_v (p^+(v)-p(v))\left(\log\frac{p(v)}{p^+(v)}+1\right)\\
 &=-\sum_v p^+(v)\log\frac{p^+(v)}{p(v)}
   -\sum_v p(v)\log\frac{p(v)}{p^+(v)}\\
 &=-\KL(p^+\|p)-\KL(p\|p^+).
\end{align*}
It is strictly negative for $p\ne p^+$. Every $p_\tau$ on the segment is positive, so $p_\tau=\operatorname{softmax}(z_\tau)$ for $z_\tau(v)=\log p_\tau(v)+c$, where a common shift $c$ makes $z_0=z$. The negative derivative along this logit curve proves that $p\ne p^+$ is nonstationary. At $p^+$ both logit gradients vanish.
\end{proof}

\paragraph{Pair odds and total mass.}
For a vocabulary with at least one token outside $P=\{a,b\}$, let $p_{\mathrm{out}}$ and $p^+_{\mathrm{out}}$ denote the distributions conditioned on its complement. Inside $P$, substitute $p(a)=mu$, $p(b)=m(1-u)$ and the corresponding teacher probabilities. The two contributions are
\begin{align*}
 \sum_{v\in P}p(v)\log\frac{p(v)}{p^+(v)}
 &=mu\log\frac{mu}{m^+q}+m(1-u)\log\frac{m(1-u)}{m^+(1-q)}\\
 &=m\log\frac{m}{m^+}+m\,\mathrm{kl}(u\|q),\\
 \sum_{v\notin P}p(v)\log\frac{p(v)}{p^+(v)}
 &=(1-m)\sum_{v\notin P}p_{\mathrm{out}}(v)
      \log\frac{(1-m)p_{\mathrm{out}}(v)}{(1-m^+)p^+_{\mathrm{out}}(v)}\\
 &=(1-m)\log\frac{1-m}{1-m^+}
       +(1-m)\KL(p_{\mathrm{out}}\|p^+_{\mathrm{out}}).
\end{align*}
Here $\mathrm{kl}$ is Bernoulli KL. The two mass-ratio terms combine to give
\begin{align}
 \KL(p\|p^+)
 &=\KL\!\big(\operatorname{Bern}(m)\,\big\|\,\operatorname{Bern}(m^+)\big)
 \nonumber\\
 &\quad + m\,\KL\!\big(\operatorname{Bern}(u)\,\big\|\,\operatorname{Bern}(q)\big)
 \nonumber\\
 &\quad + (1-m)\,\KL(p_{\mathrm{out}}\|p^+_{\mathrm{out}}).
 \label{eq:fp-kl-split}
\end{align}
Define $\widetilde p$ to agree with $p$ outside $P$ and to assign $mq$ to $a$ and $m(1-q)$ to $b$. This replacement leaves the total-mass and complement terms unchanged and sets the middle term of \eqref{eq:fp-kl-split} to zero. Thus
\begin{equation}
 \KL(p\|p^+)-\KL(\widetilde p\|p^+)
 =m\,\mathrm{kl}(u\|q).
 \label{eq:fp-conditional-gap}
\end{equation}
The pair loss fits the conditional odds with coefficient $\alpha$, independent of $m$. When $u=q$ but $m\ne m^+$, its gradient vanishes and the reverse KL still penalizes the mass discrepancy. In particular, $p(a)=mu$ can remain small after the conditional odds have been corrected. The pair-mass factor in \eqref{eq:fp-kl-split} and the rare-token softmax Jacobian are distinct sources of attenuation: the binary analysis has $m=1$ and isolates the latter.

\subsection{Time to Reverse a Binary Preference}
\label{app:convergence}

The following proposition gives an explicit FPD upper bound and the vanilla asymptotic rate, which yield the comparison in Proposition~\ref{prop:fp-main-recovery}.

\begin{proposition}[Steps to reach equal odds]
\label{prop:fp-escape}
\label{prop:fp-bound-main}
Suppose $\mathcal V=\{a,b\}$ and the teacher assigns $q=p^+(a)>1/2$. Use exact gradient descent on both logits with fixed step size $\eta>0$, starting from $p_0(a)=\epsilon\in(0,1/2)$. Let $T_\alpha(\epsilon)$ be the first step at which $p(a)\ge1/2$ under \eqref{eq:fp-conditional}. Then
\begin{equation}
 \begin{aligned}
 T_0(\epsilon)
 &=\Theta\!\left(\frac{1}{\eta\epsilon\log(1/\epsilon)}\right),\\
 T_\alpha(\epsilon)
 &\le\left\lceil\frac{\log((1-\epsilon)/\epsilon)}
 {2\eta\alpha(q-1/2)}\right\rceil,\quad \alpha>0.
 \end{aligned}
 \label{eq:fp-bound-main}
\end{equation}
The first line is asymptotic as $\epsilon\to0$ with $q,\eta$ fixed. The second holds for every $\epsilon\in(0,1/2)$ and $\alpha>0$.
\end{proposition}
\begin{proof}
In the binary case $u=\sigma(M)=p(a)$ and the teacher margin is $M^+=\log(q/(1-q))$. Differentiating the two summands in the binary reverse KL gives
\begin{align*}
 K(u)&=u\log\frac{u}{q}+(1-u)\log\frac{1-u}{1-q},\\
 \frac{dK}{du}
 &=\log\frac{u}{q}+1-\log\frac{1-u}{1-q}-1
   =M-M^+,\\
 \frac{dJ_\alpha}{dM}
 &=u(1-u)(M-M^+)+\alpha(u-q).
\end{align*}
The last line uses $du/dM=u(1-u)$ and the pair derivative from Appendix~\ref{app:gradients}. One gradient step subtracts $\eta J_\alpha'(M)$ from $z_a$ and adds the same quantity to $z_b$. Their difference changes by $-2\eta J_\alpha'(M)$, so
\begin{equation}
 \Delta M=2\eta\big[u(1-u)(M^+-M)+\alpha(q-u)\big].
 \label{eq:fp-margin-step}
\end{equation}
For $\alpha>0$, set $c=2\eta\alpha(q-1/2)>0$. Before the first crossing, $M_j<0$ implies $u_j<1/2$ and $M^+-M_j>0$. Equation~\eqref{eq:fp-margin-step} then gives $M_{j+1}-M_j\ge c$. If the margin stays negative through step $k-1$, summation yields
\[
 M_k=M_0+\sum_{j=0}^{k-1}(M_{j+1}-M_j)
 \ \ge\ \log\frac{\epsilon}{1-\epsilon}+kc.
\]
At $k=\lceil{-M_0}/c\rceil$ the right-hand side is nonnegative. Thus the margin reaches zero or crosses it by this step.

For $\alpha=0$, set $x=-M$ up to the crossing and write
\[
 \delta(x)=2\eta\,\sigma(-x)(1-\sigma(-x))(M^++x).
\]
The step moves $x$ to $x-\delta(x)$. Substituting $\sigma(-x)(1-\sigma(-x))=e^{-x}/(1+e^{-x})^2$ gives $\delta(x)=2\eta e^{-x}(M^++x)/(1+e^{-x})^2$. For large $x$, this is $\Theta(\eta x e^{-x})$. Choose $L>1$ so that $\delta(x)\le1/2$ for $x\ge L$. To count steps of unequal lengths, introduce $\Phi(x)=\int_L^x e^s/s\,ds$. For $x\ge L+1$ the integral over a single step satisfies
\[
 \delta(x)\frac{e^{x-1/2}}{x}
 \ \le\ \Phi(x)-\Phi(x-\delta(x))
 =\int_{x-\delta(x)}^x\frac{e^s}{s}\,ds
 \ \le\ \delta(x)\frac{e^x}{x-1/2}.
\]
The bounds follow from $x-1/2\le s\le x$. Substituting the estimate for $\delta(x)$ bounds each decrease in $\Phi$ between $c_1\eta$ and $c_2\eta$ for constants $c_1,c_2>0$. At the first step $\tau$ with $x_\tau<L+1$, we have $L+1/2\le x_\tau<L+1$, so $\Phi(x_\tau)=O(1)$. Summing the decreases gives
\[
 c_1\eta\tau
 \ \le\ \sum_{j=0}^{\tau-1}\big(\Phi(x_j)-\Phi(x_{j+1})\big)
 =\Phi(x_0)-\Phi(x_\tau)
 \ \le\ c_2\eta\tau .
\]
Since $(e^x/x)'=(e^x/x)(1-1/x)$, l'H\^{o}pital's rule gives $\Phi(x)/(e^x/x)\to1$. Thus $\tau=\Theta(e^{x_0}/(\eta x_0))$. On $[0,L+1]$ the continuous function $\delta$ has a positive minimum, so only a bounded number of additional steps is needed. Finally, $e^{x_0}=(1-\epsilon)/\epsilon$ and $x_0=\log((1-\epsilon)/\epsilon)\sim\log(1/\epsilon)$. Substituting both terms gives the stated rate for $T_0(\epsilon)$.
\end{proof}

\begin{corollary}[Sharp fixed-parameter escape rates]
\label{cor:fp-sharp-escape}
Under Proposition~\ref{prop:fp-escape}, fix $q\in(1/2,1)$, $\eta>0$, and $\alpha>0$. As $\epsilon\to0$,
\begin{align}
 T_0(\epsilon)
 &\sim \frac{1}{2\eta\epsilon\log(1/\epsilon)},\nonumber\\
 T_\alpha(\epsilon)
 &\sim \frac{\log(1/\epsilon)}{2\eta\alpha q}.
 \label{eq:fp-sharp-times}
\end{align}
\end{corollary}
\begin{proof}
For vanilla OPD, use $x=-M$ and $\delta(x)$ from the preceding proof. Choose a sufficiently large fixed cutoff $L$, and define
\[
 \Psi(x)=\int_L^x\frac{e^s}{s+M^+}\,ds.
\]
Since $\delta(x)\to0$, the mean value theorem gives
\[
 \Psi(x)-\Psi(x-\delta(x))
 \sim \Psi'(x)\delta(x)
 =\frac{2\eta}{(1+e^{-x})^2}
 \longrightarrow2\eta.
\]
More explicitly, for any $\nu\in(0,1)$, a larger fixed cutoff makes each potential decrement lie between $2\eta(1-\nu)$ and $2\eta(1+\nu)$. The entry point below that cutoff has bounded potential, and the remaining crossing time is bounded independently of $\epsilon$. Summing and then taking $\nu\to0$ yields $T_0\sim\Psi(x_0)/(2\eta)$. Since $\Psi(x)\sim e^x/(x+M^+)$ and $x_0=\log((1-\epsilon)/\epsilon)$, this proves the first equivalent.

For FPD, the increment in \eqref{eq:fp-margin-step} converges to $c_\alpha=2\eta\alpha q>0$ as $M\to-\infty$. For any $\nu\in(0,1)$, all increments below a sufficiently negative fixed cutoff lie between $c_\alpha(1-\nu)$ and $c_\alpha(1+\nu)$. The increment is bounded on $(-\infty,0]$, so the crossing of this cutoff has bounded overshoot. The remaining time to reach zero is bounded by the positive lower increment in Proposition~\ref{prop:fp-escape}. Hence $T_\alpha\sim(-M_0)/c_\alpha$, which gives the second equivalent.
\end{proof}

For fixed coefficients, the speedup ratio therefore satisfies
\[
 \frac{T_0(\epsilon)}{T_\alpha(\epsilon)}
 \sim\frac{\alpha q}{\epsilon[\log(1/\epsilon)]^2}
 \longrightarrow\infty.
\]
These equivalents are not uniform when the teacher odds, step size, or pair coefficient vary with $\epsilon$.

\subsection{Escape under Stochastic Sampled-Token Updates}
\label{app:sampled-dynamics}

The preceding results iterate the exact reverse-KL gradient. We now analyze freshly sampled gradients at the same binary prefix. At update $k$, draw $Y_k\sim p_k$ from the full softmax and recompute $A_k(Y_k)=\log[p^+(Y_k)/p_k(Y_k)]$. With $p_z=\operatorname{softmax}(z)$, take one SGD step on both logits using the local surrogate
\[
 \begin{aligned}
 \widehat J_k(z)&=-\sg[A_k(Y_k)]\log p_z(Y_k)\\
 &\quad+\alpha\ell_{\mathrm{BT}}(z_a-z_b;q),
 \end{aligned}
\]
whose gradient is evaluated at $z=z_k$. The advantage is detached only during differentiation. Define the random first crossing
\[
 \tau_\alpha=\inf\{k\ge0:p_k(a)\ge1/2\}.
\]

\begin{proposition}[Stochastic escape at a binary fork]
\label{prop:fp-sampled-escape}
Fix $q\in(1/2,1)$ and $\eta>0$, and initialize $p_0(a)=\epsilon\in(0,1/2)$. For vanilla sampled-token OPD,
\begin{equation}
 \E[\tau_0]=\Theta(1/\epsilon)
 \qquad(\epsilon\to0).
 \label{eq:fp-stochastic-vanilla}
\end{equation}
For any fixed $\alpha>0$, every sample path of the augmented update satisfies
\begin{equation}
 \tau_\alpha\le
 \left\lceil
 \frac{\log((1-\epsilon)/\epsilon)}
 {2\eta\alpha(q-1/2)}
 \right\rceil.
 \label{eq:fp-sampled-bound}
\end{equation}
Constants in \eqref{eq:fp-stochastic-vanilla} may depend on $q$ and $\eta$.
\end{proposition}
\begin{proof}
When the sampled token is $a$, the base update changes the margin by
\[
 D_a(u)=2\eta(1-u)\log(q/u).
\]
When it is $b$, the increment is
\[
 D_b(u)=2\eta u\log((1-u)/(1-q)).
\]
Thus the stochastic augmented recursion is
\begin{equation}
 M_{k+1}-M_k
 =D_{Y_k}(u_k)+2\eta\alpha(q-u_k).
 \label{eq:fp-sampled-step}
\end{equation}
Before the first crossing, $u_k<1/2<q$, so both base increments are positive. The pair increment is at least $2\eta\alpha(q-1/2)$ regardless of $Y_k$. Summing this lower bound proves \eqref{eq:fp-sampled-bound}.

For vanilla, first establish a pathwise upper bound. Define
\[
 c_q=\min\left\{\log(2q),\log\frac1{2(1-q)}\right\}>0.
\]
For $u<1/2$, both $D_a(u)$ and $D_b(u)$ are at least $2\eta c_qu$. Set $w=e^{-M}=(1-u)/u>1$ and $\kappa=\eta c_q$. Since $u=1/(1+w)\ge1/(2w)$, every pre-crossing increment is at least $\kappa/w$. Therefore
\begin{align*}
 w-w_{\mathrm{next}}
 &\ge w(1-e^{-\kappa/w})\\
 &\ge \frac{\kappa}{1+\kappa/w}
 \ge \frac{\kappa}{1+\kappa}
 =:d>0,
\end{align*}
using $1-e^{-x}\ge x/(1+x)$ for $x\ge0$. It follows that
\[
 \tau_0\le 1+\left\lceil\frac{w_0-1}{d}\right\rceil
 =O(1/\epsilon)
\]
on every path.

For the expected lower bound, let $b_q=-\log(1-q)>0$ and choose
\[
 C=\min\left\{\frac18,\frac{\log2}{8\eta b_q}\right\},
 \qquad n=\lfloor C/\epsilon\rfloor.
\]
Consider the deterministic trajectory that draws $b$ at every step. Whenever $u\le2\epsilon$,
\[
 D_b(u)\le4\eta b_q\epsilon.
\]
For sufficiently small $\epsilon$, a first-exit argument keeps this trajectory below $2\epsilon$ for all $j\le n$. Indeed, at any hypothetical first exit,
\[
 M_j-M_0\le4\eta b_q\epsilon n
 \le\tfrac12\log2,
 \qquad
 \frac{u_j}{\epsilon}\le e^{M_j-M_0}\le\sqrt2<2,
\]
a contradiction. The probability that the stochastic trajectory follows these $n$ draws is consequently at least
\[
 \prod_{j=0}^{n-1}(1-u_j^{(b)})
 \ge1-\sum_{j=0}^{n-1}u_j^{(b)}
 \ge1-2\epsilon n
 \ge1-2C\ge\tfrac34.
\]
On this event no crossing has occurred by step $n$. Hence
\[
 \E[\tau_0]\ge n\,\mathbb P(\tau_0>n)
 \ge\tfrac34\lfloor C/\epsilon\rfloor
 =\Omega(1/\epsilon).
\]
Together with the pathwise upper bound, this proves \eqref{eq:fp-stochastic-vanilla}.
\end{proof}

The expectation of a random hitting time differs from the hitting time of the expected-gradient recursion. Proposition~\ref{prop:fp-sampled-escape} and Proposition~\ref{prop:fp-escape} therefore describe different update processes. The stochastic pair bound controls the first crossing; without additional step-size conditions it does not assert that later iterates remain above $1/2$. The matched-budget return ordering below concerns the exact-gradient recursion.

\subsection{When Fork Corrections Improve Task Return}
\label{app:task-return}

\paragraph{One update at a binary fork.}
Start both students from margin $M$ and take a gradient step of size $\eta>0$ on both logits. Denote the common base increment by $\Delta M_{\rm base}$. The vanilla margin becomes $M_V=M+\Delta M_{\rm base}$. Adding the pair term gives $M_F=M_V+2\eta\alpha(q-u)$ by \eqref{eq:pair-grad}. For fixed conditional expected rewards $r_a$ and $r_b$, the expected reward at margin $x$ is $J^{\rm task}(x)=r_a\sigma(x)+r_b(1-\sigma(x))$. Collecting the coefficients of each action gives
\begin{equation}
\begin{aligned}
 J^{\rm task}(M_F)-J^{\rm task}(M_V)
 &=r_a[\sigma(M_F)-\sigma(M_V)]+r_b[\sigma(M_V)-\sigma(M_F)]\\
 &=(r_a-r_b)[\sigma(M_F)-\sigma(M_V)] .
\end{aligned}
\label{eq:fp-local-gain}
\end{equation}
Because $\sigma$ is strictly increasing, the bracket has the sign of $q-u$ for $\alpha>0$. This proves the sign criterion in \eqref{eq:fp-local-gain}, including equality if $q=u$ or $r_a=r_b$. For example, when $q>u$ but $r_a=0$ and $r_b=1$, the pair term shifts mass toward the action with lower reward and strictly decreases return.

\begin{corollary}[Strict return gain at matched training budgets]
\label{cor:fp-matched-return}
In the binary model with free logits, let $q\in(1/2,1)$, $M^+=\log(q/(1-q))$, $\alpha>0$, and $r_a>r_b$. Start both students at the same $p_0(a)=\epsilon\in(0,q)$ and run gradient descent on $J_\alpha$ and $J_0$ with $0<\eta<2/(1+\max\{\alpha,M^+\})$. Write $J^{\rm task}_{j,k}=r_b+(r_a-r_b)p_{j,k}(a)$ for the task return of student $j\in\{F,V\}$ after $k$ steps. Then $J^{\rm task}_{F,k}>J^{\rm task}_{V,k}$ for every finite $k\ge1$. Moreover, for $\epsilon<1/2$, let $K_\epsilon$ be the ceiling in \eqref{eq:fp-bound-main}. With all other quantities fixed,
\[
 \liminf_{\epsilon\to0}\big(J^{\rm task}_{F,K_\epsilon}
        -J^{\rm task}_{V,K_\epsilon}\big)\ \ge\ (r_a-r_b)/2 .
\]
\end{corollary}
\begin{proof}
For either coefficient $\gamma\in\{0,\alpha\}$, let $f_\gamma(M)$ be the updated margin in \eqref{eq:fp-margin-step}. For $M<M^+$, the mean value theorem gives $q-u=\sigma'(\xi)(M^+-M)$ for some $\xi$ between $M$ and $M^+$. Thus
\[
 f_\gamma(M)=M+\beta_\gamma(M^+-M),
 \qquad
 \beta_\gamma=2\eta[u(1-u)+\gamma\sigma'(\xi)].
\]
The step-size assumption implies
\[
 0<\beta_\gamma\le\frac{\eta(1+\gamma)}2
 \le\frac{\eta(1+\alpha)}2<1.
\]
Both margins therefore increase and remain below $M^+$.

Once the two students have different logits, comparing the next steps requires the derivative of the vanilla update,
\[
 f'_0(M)=1+2\eta u(1-u)\big[(1-2u)(M^+-M)-1\big].
\]
For $M\le0$ the product $(1-2u)(M^+-M)$ is nonnegative. For $0<M<M^+$ it is bounded below by $-M^+$. Since $u(1-u)\le1/4$, we have $f'_0(M)\ge 1-\eta(1+M^+)/2>0$ throughout. Also $f_\alpha(M)-f_0(M)=2\eta\alpha(q-u)>0$ there. Now suppose that $M_{F,k}\ge M_{V,k}$, as holds with equality at initialization. Then
\[
 M_{F,k+1}=f_\alpha(M_{F,k})
 > f_0(M_{F,k})\ \ge\ f_0(M_{V,k})=M_{V,k+1}.
\]
The strict inequality compares the two updates at the same margin. The weak inequality uses monotonicity of the vanilla update at two different margins. Since both students stay below $M^+$, this argument applies at every finite step. Multiplying the resulting positive probability difference by $r_a-r_b$ proves the first claim.

For the quantitative bound, Proposition~\ref{prop:fp-escape} and monotonicity give $p_{F,K_\epsilon}(a)\ge1/2$. To bound the vanilla probability at the same step, write $M_0=\log(\epsilon/(1-\epsilon))$. For sufficiently small $\epsilon$ and any $M\in[M_0,M_0+1]$, we have $u\le e^M\le e\epsilon/(1-\epsilon)$ and $0<M^+-M\le M^+-M_0$. Hence the vanilla increment obeys
\[
 0<f_0(M)-M
 \le 2\eta\,\frac{e\epsilon}{1-\epsilon}(M^+-M_0)
 = O\big(\eta\epsilon\log(1/\epsilon)\big).
\]
With $q,\eta,\alpha$ fixed, $K_\epsilon=O(\log(1/\epsilon))$. Even if the last step were the first to leave the interval, summing this bound through that step would give a total increase of at most $O(\epsilon\log^2(1/\epsilon))=o(1)$, too small to cross its upper endpoint. Thus $h_\epsilon=M_{V,K_\epsilon}-M_0=o(1)$. Converting back to probabilities gives
\[
 \frac{p_{V,K_\epsilon}(a)}{\epsilon}
 =\frac{\sigma(M_0+h_\epsilon)}{\sigma(M_0)}
 =\frac{e^{h_\epsilon}}{1-\epsilon+\epsilon e^{h_\epsilon}}\ \longrightarrow\ 1.
\]
The return gap at $K_\epsilon$ is therefore at least $(r_a-r_b)[1/2-\epsilon(1+o(1))]$. Taking its lower limit proves the second claim.
\end{proof}

\paragraph{From a fork to a sequence.}
Consider two trained generation policies $\pi_F$ (FPD) and $\pi_V$ (vanilla), evaluated on the same prompt distribution and terminal score $R_d\in[0,1]$ in each domain $d$. Let $\rho_d\ge0$ be the evaluation weights, $\sum_d\rho_d=1$, and write $J_\rho(\pi)=\sum_d\rho_d\E_{\pi,d}R_d$. Assume a finite response horizon $H$, carrying each completed prefix and its terminal score forward to step $H$. For a prefix $s$, let $Q_d^V(s,v)$ be the expected terminal score after taking $v$ and following $\pi_V$ thereafter. Thus $Q_d^V(s,a)-Q_d^V(s,b)$ measures the task advantage of the strong teacher's candidate under a common continuation policy.

At a fork, let $m_j=\pi_j(a\mid s)+\pi_j(b\mid s)$ for $j\in\{F,V\}$, and let $u_j=\pi_j(a\mid s)/m_j$ when $m_j>0$. Define the signed probability shift toward $a$ within the mass shared by both pairs as
\[
 \delta_s=\min(m_F,m_V)(u_F-u_V).
\]
Set $\delta_s=0$ when either mass is zero or the prefix has no fork. Decompose the change in the next-token distribution as
\begin{equation}
 \pi_F(\cdot\mid s)-\pi_V(\cdot\mid s)
 =\delta_s(e_a-e_b)+r_s .
 \label{eq:fp-policy-shift}
\end{equation}
At a nonfork the first term is zero. The residual $r_s$ includes changes in total pair mass and in probabilities of other tokens.

\begin{lemma}[Explicit residual at a fork]
\label{lem:fp-residual}
At a fork with pair $P=\{a,b\}$,
\begin{equation}
 \begin{aligned}
 \|r_s\|_1
 &=|m_F-m_V|\\
 &\quad+\sum_{v\notin P}|\pi_F(v\mid s)-\pi_V(v\mid s)|.
 \end{aligned}
 \label{eq:fp-residual-mass}
\end{equation}
This identity also holds if either pair mass is zero. At a nonfork, $\|r_s\|_1=\|\pi_F(\cdot\mid s)-\pi_V(\cdot\mid s)\|_1$.
\end{lemma}
\begin{proof}
Suppress $s$. If $m_F\ge m_V>0$, subtracting $\delta=m_V(u_F-u_V)$ from the difference on $a$, and adding it on $b$, gives
\[
 r(a)=(m_F-m_V)u_F,\qquad
 r(b)=(m_F-m_V)(1-u_F).
\]
Both entries have the same sign, and their absolute sum is $|m_F-m_V|$. If $m_V\ge m_F>0$, the same calculation uses $u_V$ instead of $u_F$ and gives the same absolute sum. If either pair mass is zero, $\delta=0$, and the identity follows directly without defining the missing conditional probability. Outside $P$, the residual equals the original probability difference. At a nonfork the definition sets $\delta=0$, proving the final claim.
\end{proof}
 Write $\mathcal E_d[f]=\sum_{t=1}^H\E_{\pi_F,d}[f(s_t)]$ for the expected sum over prefixes visited by $\pi_F$. The summands below are zero after termination.

\begin{proposition}[Sufficient condition for a task-return gain]
\label{prop:fp-task-return}
Let $Q_s(v)=Q_d^V(s,v)$ and $\operatorname{span}(Q_s)=\max_v Q_s(v)-\min_v Q_s(v)$. Define
\begin{align*}
 G_d&=\mathcal E_d\!\left[\delta_s\big(Q_s(a)-Q_s(b)\big)\right],\\
 E_d&=\tfrac12\,\mathcal E_d\!\left[\|r_s\|_1\operatorname{span}(Q_s)\right].
\end{align*}
Then
\begin{equation}
 J_\rho(\pi_F)-J_\rho(\pi_V)\ \ge\ \sum_d\rho_d(G_d-E_d).
 \label{eq:fp-return-bound}
\end{equation}
In particular, FPD has strictly higher expected return whenever the evaluation-weighted gain from within-pair probability shifts exceeds the evaluation-weighted residual bound.
\end{proposition}
\begin{proof}
Fix a domain and write $V(s)=\sum_v\pi_V(v\mid s)Q_s(v)$ for the vanilla continuation value. Let $s_1$ be the prompt, $s_t$ the prefix before action $t$, and $s_{H+1}$ the completed response with its terminal score. Because both policies receive the same prompt distribution, $\E_{\pi_F,d}V(s_1)=\E_{\pi_V,d}R_d$. Taking the terminal value to be the score gives $\E_{\pi_F,d}V(s_{H+1})=\E_{\pi_F,d}R_d$. Subtracting and expanding the difference along the trajectory gives
\begin{align*}
 \E_{\pi_F,d}R_d-\E_{\pi_V,d}R_d
 &=\E_{\pi_F,d}\!\left[V(s_{H+1})-V(s_1)\right]\\
 &=\sum_{t=1}^H\E_{\pi_F,d}\!\left[V(s_{t+1})-V(s_t)\right].
\end{align*}
To evaluate a summand, condition on $s_t=s$. For any next token $v$, the expected vanilla value of the next prefix equals $Q_s(v)$. Averaging over the action drawn by FPD and expanding the current vanilla value gives
\begin{align*}
 \E_{\pi_F,d}[V(s_{t+1})-V(s_t)\mid s_t=s]
 &=\sum_v\pi_F(v\mid s)Q_s(v)-\sum_v\pi_V(v\mid s)Q_s(v)\\
 &=\big\langle\pi_F(\cdot\mid s)-\pi_V(\cdot\mid s),Q_s\big\rangle .
\end{align*}
Summing these conditional expectations over the prefixes visited by FPD gives the finite-horizon performance-difference identity~\citep{kakade2002approximately,schulman2015trust}. Substituting \eqref{eq:fp-policy-shift} into the inner product separates the two contributions at $s$,
\[
 \big\langle\pi_F-\pi_V,Q_s\big\rangle
 =\delta_s\big(Q_s(a)-Q_s(b)\big)+\sum_v r_s(v)Q_s(v).
\]
The entries of $r_s$ sum to zero because both policies have unit mass and $e_a-e_b$ has zero mass. Write $c_s=(\max_v Q_s(v)+\min_v Q_s(v))/2$. Subtracting this constant from all values preserves the residual and yields
\begin{align*}
 \left|\sum_v r_s(v)Q_s(v)\right|
 &=\left|\sum_v r_s(v)(Q_s(v)-c_s)\right|\\
 &\le\sum_v |r_s(v)|\,|Q_s(v)-c_s|
 \le \tfrac12\|r_s\|_1\operatorname{span}(Q_s).
\end{align*}
Taking the expected sum over prefixes gives $\E_{\pi_F,d}R_d-\E_{\pi_V,d}R_d\ge G_d-E_d$. Multiplying by the nonnegative evaluation weights $\rho_d$ and summing over domains gives \eqref{eq:fp-return-bound}.
\end{proof}

For a single domain, suppose $\delta_s\ge0$ at all visited forks and $Q_s(a)-Q_s(b)\ge g>0$ wherever $\delta_s>0$. Then $G_d\ge g\,\mathcal E_d[\delta_s]$, which gives a strict improvement if $g\,\mathcal E_d[\delta_s]>E_d$. For a mixture, \eqref{eq:fp-return-bound} allows gains in one domain to offset losses in another according to the evaluation weights. These weights describe the task being measured. The fractions $F_d/F$ in Section~\ref{sec:method-mopd} describe how the training step weights fork losses.

\subsection{Conditional Recovery in a Full Vocabulary}
\label{app:full-vocabulary}

For a vocabulary larger than two tokens, let
\[
 B(p)=p(a)(A(a)+K)-p(b)(A(b)+K)
\]
be the exact base contribution from \eqref{eq:full-vocab-margin}. The free-logit recursion becomes
\begin{equation}
 M_{k+1}-M_k
 =\eta\big[B(p_k)+2\alpha(q-u_k)\big].
 \label{eq:fp-full-margin-step}
\end{equation}
Unlike the binary contribution, $B(p)$ need not be positive when $u<q$, because the base objective also adjusts pair mass and the rest of the vocabulary.

\begin{proposition}[Conditional recovery under controlled base interference]
\label{prop:fp-full-vocab-bound}
Fix a pair with $q>1/2$, and start at $u_0<1/2$. Suppose $B(p_k)\ge-\beta$ before the first crossing of $u_k=1/2$, for some $\beta\ge0$. If $c=2\alpha(q-1/2)-\beta>0$, then exact-gradient descent on the free logits satisfies
\[
 T_{\mathrm{pair}}\le
 \left\lceil
 \frac{\log((1-u_0)/u_0)}{\eta c}
 \right\rceil.
\]
\end{proposition}
\begin{proof}
Before crossing, $u_k<1/2$, and \eqref{eq:fp-full-margin-step} gives $M_{k+1}-M_k\ge\eta c$. Summing until the proposed bound forces the margin to become nonnegative.
\end{proof}

This condition controls the contribution absent from the binary reduction. A uniform, although conservative, bound follows directly from the finite-vocabulary setup. With $m=p(a)+p(b)$ and $L_+=\log(1/p^+_{\min})$,
\begin{align*}
 |B(p)|
 &\le\sum_{v\in P}p(v)|A(v)|+mK\\
 &\le mL_+-\sum_{v\in P}p(v)\log p(v)+mL_+\\
 &\le m\big[\log(1/m)+\log2+2L_+\big].
\end{align*}
Here the pair entropy equals $m\log(1/m)+mH(u)$, and $H(u)\le\log2$. Thus the magnitude of the base margin contribution vanishes uniformly as $m\to0$, whereas a fixed conditional gap $q-u$ gives a nonvanishing fork contribution.

\paragraph{Conditional calibration and total mass.}
Even after $u$ reaches $q$, the absolute probability $p(a)=mu$ may remain small. Under a pair-only free-logit SGD step, let $c=\eta\alpha(q-u)$. Expanding the softmax gives
\begin{align*}
 u_{\mathrm{next}}-u
 &=2c\,u(1-u)+O(c^2),\\
 m_{\mathrm{next}}-m
 &=c\,m(1-m)(2u-1)+O(c^2).
\end{align*}
For $u<1/2$ and $q>u$, the conditional share increases while the pair mass can decrease. The pair term corrects conditional odds; the base term remains responsible for distributional mass allocation.

\paragraph{Per-step KL progress need not be ordered.}
Let $g=\nabla_z K$ and $h=\nabla_z\ell_{\mathrm{BT}}$. Then
\[
 \langle g,h\rangle=(q-u)B(p).
\]
For the constructed distributions
\[
 p=(0.004,0.006,0.99),\qquad
 p^+=(0.64,0.16,0.20),
\]
with $P=\{a,b\}$ corresponding to the first two entries, $u=0.4$, $q=0.8$, and $B(p)\approx-0.0024866$. A foil whose first choice is $b$ forms a valid fork. Hence $\langle g,h\rangle<0$. For the two steps $z_V=z-\eta g$ and $z_F=z-\eta(g+\alpha h)$,
\[
 K(z_F)-K(z_V)
 =-\eta\alpha\langle g,h\rangle+O(\eta^2)>0
\]
for fixed $\alpha>0$ and sufficiently small $\eta$. This illustrates why target preservation does not imply uniformly better KL progress at every full-vocabulary update.

\subsection{Shared Parameters and Implemented Training}
\label{app:parameterization}

\paragraph{From logit corrections to parameter updates.}
For a fixed batch with $F>0$, write $h_s=\nabla_\theta M_s$ and let $\mathcal L_{\mathrm{base}}$ include the token-mean base reduction. The full parameter gradient is
\[
 G=\nabla_\theta\mathcal L_{\mathrm{base}}
   +\frac{\lambda}{F}\sum_{r\in\mathcal F}(u_r-q_r)h_r.
\]
Under ordinary SGD, $\theta_{\mathrm{next}}=\theta-\eta G$. Taylor expansion gives
\begin{align}
 M_s(\theta_{\mathrm{next}})-M_s(\theta)
 &=-\eta\langle h_s,\nabla_\theta\mathcal L_{\mathrm{base}}\rangle
 \nonumber\\
 &\quad+\frac{\eta\lambda}{F}
   \sum_{r\in\mathcal F}(q_r-u_r)\langle h_s,h_r\rangle+R_s .
 \label{eq:fp-shared-margin}
\end{align}
If the margin Jacobian is $L_s$-Lipschitz along this update segment, the remainder satisfies
\[
 |R_s|\le\tfrac12L_s\eta^2\|G\|^2.
\]
An isolated fork has a self-correction proportional to $(q_s-u_s)\|h_s\|^2$. Cross-prefix inner products can have either sign, so monotonic improvement of every margin requires additional control of interference and the base update. The Gram matrix is positive semidefinite, but that property does not make every individual cross-prefix contribution positive. Adam's preconditioning and momentum further distinguish implemented updates from the SGD recursions above.

\paragraph{Full-softmax and truncated sampling.}
At a fixed prefix, nucleus sampling uses a behavior distribution $\mu(v)=p(v)\mathbf1[v\in S]/Z_S$, where $Z_S=\sum_{v\in S}p(v)$. If advantages use the untruncated student log-probabilities, its expected sampled descent direction is
\[
 g_S=\frac1{Z_S}\sum_{v\in S}p(v)A(v)\nabla_z\log p(v).
\]
The full-softmax direction instead sums over the entire vocabulary without the factor $1/Z_S$. The renormalized reference-student Top-$K$ base has the same support-restricted form at the reference parameters when clipping is inactive. Thus the full-softmax gradient identity does not apply to these truncated updates.

A fixed-prefix example also separates their stationary sets. Let
\[
 \begin{aligned}
 p^+&=(0.60,0.37,0.02,0.01),\\
 p&=(0.60,0.37,0.015,0.015).
 \end{aligned}
\]
Nucleus sampling at threshold $0.95$ selects the first two tokens. Their advantages are zero, and a fork using these two tokens has $u=q$. Both expected base and fork gradients therefore vanish, although $p\ne p^+$. This constructed example concerns the specified truncated fixed-prefix surrogate and does not contradict Proposition~\ref{prop:fp-target}, which assumes exact KL.

\paragraph{What the guarantees establish.}
The exact objective retains the teacher target. The deterministic and stochastic binary results quantify recovery of an initially rare teacher preference under their respective updates. Proposition~\ref{prop:fp-full-vocab-bound} gives an extension under controlled base interference, and Proposition~\ref{prop:fp-task-return} states sufficient conditions for sequence-level gains. These results do not establish universal reward improvement, entropy preservation, or convergence of nucleus-sampled, clipped, shared-parameter Adam training. The empirical results assess that practical regime. In multi-teacher FPD, $\lambda F_d/F$ weights each domain's mean fork loss during training, whereas $\rho_d$ specifies the task-return mixture being evaluated.

%% file: sections/appendix/entropy.tex
\section{Entropy Dynamics and Subsequent Reinforcement Learning}
\label{app:entropy}

This appendix retains the notation and fixed-prefix assumptions of Appendix~\ref{app:theory}. We first derive when fork preference corrections recover entropy and how calibration can produce a finite entropy level. We then quantify implications for discovering useful continuations and accessing them under a KL constraint in subsequent reinforcement learning.

\subsection{Entropy Recovery through Conditional Calibration}
\label{app:entropy-dynamics}

We analyze the Shannon entropy of the full student distribution,
\[
 \mathcal H(p)=-\sum_{v\in\mathcal V}p(v)\log p(v),
\]
and retain $H(u)=-u\log u-(1-u)\log(1-u)$ for binary entropy. These are distinct from the entropy of a nucleus-truncated sampling policy and from semantic diversity of complete responses. The analysis first fixes a prefix and uses free-logit gradient flow. A finite SGD step has the corresponding first-order expansion. Later paragraphs address shared parameters and changing rollout prefixes.

\paragraph{An entropy identity for the fork direction.}
The generic softmax entropy identity expresses entropy change as a covariance between log-probabilities and logit updates~\citep{cui2025entropymechanism}. We specialize that identity to the fork preference gradient.

\begin{lemma}[Entropy production of a fork correction]
\label{lem:fp-entropy-direction}
For a positive student distribution, define
\begin{equation}
 \begin{aligned}
 \Gamma_{ab}(p)
 &=p(b)\big(\log p(b)+\mathcal H(p)\big)\\
 &\quad-p(a)\big(\log p(a)+\mathcal H(p)\big).
 \end{aligned}
 \label{eq:fp-entropy-alignment}
\end{equation}
Along the pair-only logit flow $\dot z=\alpha(q-u)(e_a-e_b)$,
\begin{equation}
 \dot{\mathcal H}_{\mathrm{pair}}
 =\alpha(q-u)\Gamma_{ab}(p).
 \label{eq:fp-entropy-production}
\end{equation}
If $p(a)<p(b)$ and $p(b)\ge e^{-\mathcal H(p)}$, then $\Gamma_{ab}(p)>0$. In particular, if the teacher strictly prefers $a$ to $b$ and $b$ is a most probable student token, the fork correction strictly increases entropy while $p(a)<p(b)$.
\end{lemma}
\begin{proof}
Using the softmax derivative and $\sum_v p(v)\log p(v)=-\mathcal H(p)$,
\begin{align*}
 \frac{\partial\mathcal H}{\partial z_j}
 &=-\sum_v p(v)(\mathbf1[v=j]-p(j))(\log p(v)+1)\\
 &=-p(j)\big(\log p(j)+\mathcal H(p)\big).
\end{align*}
Consequently, for any logit velocity $\dot z$,
\[
 \dot{\mathcal H}
 =-\operatorname{Cov}_{v\sim p}(\log p(v),\dot z_v).
\]
Substituting the pair velocity proves \eqref{eq:fp-entropy-production}.

For the sign condition, set $g=e^{-\mathcal H(p)}$ and $f(x)=x\log(x/g)$. This function is nonpositive on $(0,g]$ and strictly increasing on $[g,1]$, because $f'(x)=\log(x/g)+1>0$ there. If $p(a)<g\le p(b)$, then $f(p(a))<f(p(b))$. If $g\le p(a)<p(b)$, strict increase gives the same conclusion. Thus $\Gamma_{ab}=f(p(b))-f(p(a))>0$. A most probable token satisfies $\log p(b)\ge\E_p\log p=-\mathcal H(p)$, so it satisfies the threshold. Finally, $q>1/2$ and $p(a)<p(b)$ imply $q-u>0$.
\end{proof}

The threshold distinguishes corrections of an overconfident dominant choice from comparisons between two negligible tail tokens. The sign need not be positive for every fork. More generally, comparing a vanilla and an augmented free-logit SGD step from the same distribution gives
\[
 \mathcal H(p_F)-\mathcal H(p_V)
 =\eta\alpha(q-u)\Gamma_{ab}(p)+O(\eta^2),
\]
for fixed smooth update directions as $\eta\to0$. Thus a positive pair contribution can offset entropy loss from the base update, but positivity of the complete entropy change also depends on that base contribution.

\paragraph{Binary entropy recovery and its rate.}
For $\mathcal V=\{a,b\}$, let $u=\sigma(M)$, $\chi(u)=u(1-u)$, and $M^+=\log(q/(1-q))$. The entropy gradient and exact augmented gradient flow are
\begin{align}
 \frac{dH(u)}{dM}&=-M\chi(u),\nonumber\\
 \dot M&=2\big[\chi(u)(M^+-M)+\alpha(q-u)\big].
 \label{eq:fp-binary-entropy-flow}
\end{align}
Their product gives
\begin{equation}
 \dot H
 =-2M\chi(u)\big[\chi(u)(M^+-M)+\alpha(q-u)\big].
 \label{eq:fp-binary-entropy-rate}
\end{equation}
In particular, if $u<1/2<q$, both exact OPD and FPD increase entropy, but their rates differ substantially near a collapsed wrong preference. At $u=\epsilon\to0$, with $q$ and $\alpha>0$ fixed,
\begin{align}
 \dot H_0
 &\sim 2\epsilon^2[\log(1/\epsilon)]^2,\nonumber\\
 \dot H_{\mathrm{pair}}
 &\sim 2\alpha q\,\epsilon\log(1/\epsilon),\nonumber\\
 \frac{\dot H_\alpha}{\dot H_0}
 &\sim\frac{\alpha q}{\epsilon\log(1/\epsilon)}
 \longrightarrow\infty.
 \label{eq:fp-entropy-rate-separation}
\end{align}
These equivalents follow by substituting $M\sim-\log(1/\epsilon)$ and $\chi(u)\sim\epsilon$ into \eqref{eq:fp-binary-entropy-rate}. They compare instantaneous flows at the same distribution. The absolute entropy-production rate still vanishes as $\epsilon\to0$; the nonvanishing pair signal is a logit signal.

For the pair branch alone,
\[
 \dot H_{\mathrm{pair}}=-2\alpha M\chi(u)(q-u).
\]
With $q>1/2$, this is positive both when $u<1/2$ and when $u>q$, and negative when $1/2<u<q$. The first regime corrects a dominant competing choice. The second softens a student preference that is sharper than the teacher's, while retaining $a$ as the greedy choice. Between $1/2$ and $q$, calibration can reduce entropy. FPD therefore restores entropy in specific overconfident regimes rather than maximizing entropy at every update.

\begin{corollary}[Simultaneous entropy and return gains at a finite budget]
\label{cor:fp-entropy-return}
Under the assumptions of Corollary~\ref{cor:fp-matched-return}, let $K_\epsilon$ be the bound in \eqref{eq:fp-bound-main}, with $q,\eta,\alpha>0$ fixed as $\epsilon\to0$. Then
\begin{align}
 H(p_{F,K_\epsilon}(a))&\ge H(q)>0,\nonumber\\
 H(p_{V,K_\epsilon}(a))
 &\sim\epsilon\log(1/\epsilon)\longrightarrow0.
 \label{eq:fp-matched-entropy}
\end{align}
If $r_a>r_b$, the same budget also satisfies
\begin{align}
 \liminf_{\epsilon\to0}
 \big(H(p_{F,K_\epsilon}(a))-H(p_{V,K_\epsilon}(a))\big)
 &\ge H(q),\nonumber\\
 \liminf_{\epsilon\to0}
 \big(J^{\mathrm{task}}_{F,K_\epsilon}
       -J^{\mathrm{task}}_{V,K_\epsilon}\big)
 &\ge (r_a-r_b)/2 .
 \label{eq:fp-joint-gain}
\end{align}
\end{corollary}
\begin{proof}
The step-size condition in Corollary~\ref{cor:fp-matched-return} keeps both margins below $M^+$ and increasing from the common initialization. Proposition~\ref{prop:fp-escape} therefore gives $p_{F,K_\epsilon}(a)\in[1/2,q)$. Binary entropy is decreasing on $[1/2,1)$, so its entropy is at least $H(q)$. The same corollary proves $p_{V,K_\epsilon}(a)=\epsilon(1+o(1))$, and substitution into binary entropy yields the second line of \eqref{eq:fp-matched-entropy}. Taking the lower limit proves the entropy gap. The return gap is the bound already established in Corollary~\ref{cor:fp-matched-return}.
\end{proof}

This result exhibits a regime in which greater entropy accompanies better task performance: the student first assigns excessive confidence to the inferior action, and FPD recovers probability on the better one. It does not require a general positive relation between entropy and reward.

\begin{corollary}[Entropy and return separation under sampled updates]
\label{cor:fp-stochastic-entropy-return}
Use the fresh binary sampled updates of Proposition~\ref{prop:fp-sampled-escape}, with fixed $q\in(1/2,1)$, $\alpha>0$, common initialization $\epsilon\in(0,1/2)$, and common step size
\[
 0<\eta<\frac{2}{4+\alpha}.
\]
At the budget $K_\epsilon$ in \eqref{eq:fp-bound-main}, every FPD sample path satisfies $u_{F,K_\epsilon}\in[1/2,q)$. For vanilla, as $\epsilon\to0$,
\begin{align*}
 \E[u_{V,K_\epsilon}]
 &=O\big(\epsilon\log(1/\epsilon)\big),\\
 \E[H(u_{V,K_\epsilon})]
 &=O\big(\epsilon\log(1/\epsilon)\big).
\end{align*}
Consequently, averaging over the randomness of training,
\begin{align*}
 \liminf_{\epsilon\to0}
 \big(\E[H(u_{F,K_\epsilon})]-\E[H(u_{V,K_\epsilon})]\big)
 &\ge H(q).
\end{align*}
If $r_a>r_b$, the lower limit of the corresponding expected task-return gap is at least $(r_a-r_b)/2$.
\end{corollary}
\begin{proof}
First show that sampled updates cannot overshoot the teacher margin under this step-size condition. For $M<M^+$, the mean value theorem gives
\begin{align*}
 \log(q/u)&=(1-\sigma(\xi_a))(M^+-M),\\
 \log((1-u)/(1-q))&=\sigma(\xi_b)(M^+-M),\\
 q-u&=\sigma'(\zeta)(M^+-M),
\end{align*}
for intermediate points between $M$ and $M^+$. Each sampled augmented step can therefore be written as
\[
 M_{\mathrm{next}}=M+\gamma(M^+-M),
 \qquad
 0<\gamma\le2\eta+\eta\alpha/2<1.
\]
The same bound applies to vanilla by dropping the pair coefficient. Both trajectories increase and stay below $M^+$, regardless of their draws. The pathwise crossing bound in Proposition~\ref{prop:fp-sampled-escape} then places every FPD trajectory in $[1/2,q)$ by $K_\epsilon$.

For vanilla, $K_\epsilon=O(\log(1/\epsilon))$. Along the trajectory consisting only of $b$ draws, the first-exit argument used in Proposition~\ref{prop:fp-sampled-escape} gives $u_j\le2\epsilon$ for $j\le K_\epsilon$, since the accumulated margin increment is at most
\[
 4\eta[-\log(1-q)]\epsilon K_\epsilon=o(1).
\]
The probability of a first $a$ draw within these updates is at most $2\epsilon K_\epsilon$. Outside that event $u_{V,K_\epsilon}\le2\epsilon$; on the event it is at most $q$. Hence
\[
 \E[u_{V,K_\epsilon}]
 \le2\epsilon+2q\epsilon K_\epsilon
 =O(\epsilon\log(1/\epsilon)).
\]
Similarly, for sufficiently small $\epsilon$,
\[
 \E[H(u_{V,K_\epsilon})]
 \le H(2\epsilon)+2\epsilon K_\epsilon\log2
 =O(\epsilon\log(1/\epsilon)).
\]
Every FPD trajectory has entropy at least $H(q)$ and task return at least $r_b+(r_a-r_b)/2$. Combining these bounds with the vanishing vanilla terms proves the claims.
\end{proof}

The same finite-budget separation of full-softmax entropies also holds under a fixed nucleus threshold $p_{\mathrm{nuc}}\in(0,1)$, provided the binary updates use the current untruncated log-probabilities in their advantages. For sufficiently small $\epsilon$, the all-$b$ vanilla trajectory remains at $u_j=\epsilon(1+o(1))<1-p_{\mathrm{nuc}}$ throughout $K_\epsilon=O(\log(1/\epsilon))$ updates, so its nucleus contains only $b$. The FPD crossing and no-overshoot arguments hold for every draw sequence and hence also under nucleus sampling. This statement concerns the entropy of the underlying softmax, not the entropy of its truncated behavior policy.

\paragraph{Why calibration produces a finite entropy level.}
For exact binary gradient flow, the mean value theorem rewrites \eqref{eq:fp-binary-entropy-flow} as
\[
 \dot M=-2\big[\chi(u)+\alpha\sigma'(\xi)\big](M-M^+),
\]
where $\xi$ lies between $M$ and $M^+$. The coefficient is positive. Thus the margin stays between its finite initialization and $M^+$ and moves monotonically toward $M^+$. A limit different from $M^+$ would have a nonzero velocity, so
\[
 u(t)\longrightarrow q,\qquad
 H(u(t))\longrightarrow H(q)>0.
\]
For $e=M-M^+$ near zero,
\[
 \dot e=-2(1+\alpha)q(1-q)e+O(e^2).
\]
Hence the local approach to the target is exponential. If the student starts with $u_0>q>1/2$, its entropy increases monotonically toward $H(q)$, while its greedy prediction remains $a$. If it starts below $1/2$, entropy first rises toward its maximum and can later decrease toward $H(q)$. This distinction is important when interpreting an average curve over many prefixes.

The pair-only flow with a hard label $q=1$ instead has $\dot M=2\alpha(1-u)>0$ and no finite target margin; its entropy tends to zero. The comparison concerns the pair branch: adding other objectives can change its limiting behavior. Soft-label calibration supplies a finite target and can oppose further sharpening when $u>q$.

\paragraph{A finite entropy band under persistent margin drift.}
The following conditional result allows an additional contribution that need not vanish at the teacher odds. It captures one way a calibrated correction can balance persistent sharpening.

\begin{proposition}[Calibration under bounded margin drift]
\label{prop:fp-entropy-drift}
Consider a fixed pair with continuous-time margin dynamics
\[
 \dot M=B(t,M)+2\alpha(q-\sigma(M)),
 \qquad |B(t,M)|\le b,
\]
where $q\in(0,1)$, $\alpha>0$, and a solution exists from a finite initial margin. Suppose
\[
 \ell=q-\frac{b}{2\alpha}>0,\qquad
 r=q+\frac{b}{2\alpha}<1.
\]
Then
\[
 \liminf_{t\to\infty}u(t)\ge\ell,\qquad
 \limsup_{t\to\infty}u(t)\le r,
\]
and
\begin{equation}
 \liminf_{t\to\infty}H(u(t))
 \ge\min\{H(\ell),H(r)\}>0.
 \label{eq:fp-entropy-band}
\end{equation}
For a full-vocabulary pair, if its mass eventually satisfies $m(t)\ge m_{\min}>0$, the full entropy has lower limit at least $m_{\min}\min\{H(\ell),H(r)\}$.
\end{proposition}
\begin{proof}
Choose any $\zeta>0$ with $\ell-\zeta>0$ and $r+\zeta<1$. Whenever $u\ge r+\zeta$,
\[
 \dot M\le b+2\alpha(q-r-\zeta)=-2\alpha\zeta.
\]
Whenever $u\le\ell-\zeta$, the analogous lower bound is $\dot M\ge2\alpha\zeta$. These inward-pointing velocities bring the trajectory into the interval $[\ell-\zeta,r+\zeta]$ in finite time if it starts outside, and prevent subsequent exit. Sending $\zeta\downarrow0$ proves the probability bounds. Concavity and continuity of binary entropy give \eqref{eq:fp-entropy-band}. Finally, the entropy chain rule for the pair and its complement is
\[
 \mathcal H(p)=H(m)+mH(u)+(1-m)\mathcal H(p_{\mathrm{out}}).
\]
All terms are nonnegative, so $\mathcal H(p)\ge mH(u)$, proving the full-vocabulary statement.
\end{proof}

The bound on $B$ is an assumption, not a consequence of a bounded BT margin derivative. It can represent an effective base contribution or other controlled logit interference. The full-vocabulary conclusion also requires pair mass to remain nonnegligible.

An illustrative case takes $B(t,M)=b_0>0$ constant and $b_0<2\alpha(1-q)$. Without the pair term, $M(t)=M_0+b_0t$ and entropy collapses to zero. With the pair term, the unique globally attracting equilibrium is
\[
 u_*=q+\frac{b_0}{2\alpha}<1,\qquad
 M_*=\log\frac{u_*}{1-u_*}.
\]
The velocity has the opposite sign to $M-M_*$, and its derivative at equilibrium is $-2\alpha u_*(1-u_*)<0$. When $u_0>u_*>1/2$, entropy rises monotonically and saturates at $H(u_*)$. This is a model of effective sharpening drift, not the exact binary reverse-KL recursion, whose base contribution already vanishes at $q$.

\paragraph{From fixed prefixes to the logged entropy curve.}
For a fixed diagnostic prefix distribution $\nu$ and a fixed training batch with $F>0$, define $\overline{\mathcal H}_\nu(\theta)=\E_{s\sim\nu}\mathcal H(p_\theta(\cdot\mid s))$ and $h_s=\nabla_\theta M_s$. Under the shared-parameter gradient flow,
\begin{align}
 \dot{\overline{\mathcal H}}_\nu
 &=-\langle\nabla_\theta\overline{\mathcal H}_\nu,
               \nabla_\theta\mathcal L_{\mathrm{base}}\rangle
 \nonumber\\
 &\quad+\frac{\lambda}{F}\sum_{s\in\mathcal F}
 (q_s-u_s)\langle\nabla_\theta\overline{\mathcal H}_\nu,h_s\rangle .
 \label{eq:fp-shared-entropy}
\end{align}
Thus the pair component can offset a negative base contribution if its aggregate alignment is sufficiently positive. Cross-prefix effects and Adam updates require the qualifications in Section~\ref{app:parameterization}.

The logged token-mean entropy additionally changes with the prefixes visited by the rollout policy. Let $d_k$ denote its normalized valid-token prefix distribution and $\mathcal H_k(s)=\mathcal H(p_{\theta_k}(\cdot\mid s))$. Then the population diagnostic obeys the exact decomposition
\begin{align}
 \overline{\mathcal H}_{k+1}-\overline{\mathcal H}_k
 &=\E_{d_k}[\mathcal H_{k+1}-\mathcal H_k]\nonumber\\
 &\quad+\big(\E_{d_{k+1}}-\E_{d_k}\big)\mathcal H_{k+1}.
 \label{eq:fp-entropy-occupancy}
\end{align}
The first term is the same-prefix distributional change; the second is the change in visited contexts, including effects of response lengths. A plateau can reflect reduced calibration error, a balance of entropy-producing and entropy-reducing updates, changing prefix coverage, or sampling fluctuations. A flat entropy curve alone does not establish matched teacher odds, and these results do not predict a particular training step at which a plateau must occur.

\subsection{Implications for Subsequent Reinforcement Learning}
\label{app:rl-initialization}

The relevant benefit for later RL is probability on useful alternative continuations. Higher vocabulary entropy alone does not imply that such continuations receive more mass. We quantify the benefit in a reward-aligned binary setting and in a KL-constrained local policy-improvement problem. The sampling statements use independent full-softmax draws, or an actual behavior policy with the stated candidate probabilities.

\begin{proposition}[Discovery and informative reward groups]
\label{prop:fp-rl-discovery}
At the matched budget of Corollary~\ref{cor:fp-entropy-return}, write $u_F=p_{F,K_\epsilon}(a)$ and $u_V=p_{V,K_\epsilon}(a)$. Then $u_F\in[1/2,q)$ and $u_V=\epsilon(1+o(1))$. If $a$ is a useful action, the number of independent draws needed to observe it with probability at least $1-\delta$, for fixed $\delta\in(0,1)$, satisfies
\[
 n_F\le\left\lceil\frac{\log(1/\delta)}{\log2}\right\rceil,
 \qquad
 n_V\sim\frac{\log(1/\delta)}{\epsilon}.
\]
If the two actions yield deterministic distinct binary rewards, a group of $G\ge2$ independent samples has nonzero reward variance with probability
\[
 P_{\mathrm{mix}}(u)=1-u^G-(1-u)^G.
\]
For fixed $G$,
\begin{align}
 P_{\mathrm{mix}}(u_F)
 &\ge1-q^G-(1-q)^G>0,\nonumber\\
 P_{\mathrm{mix}}(u_V)&\sim G\epsilon.
 \label{eq:fp-informative-groups}
\end{align}
\end{proposition}
\begin{proof}
The probability of observing $a$ in $n$ draws is $1-(1-u)^n$. Since $u_F\ge1/2$, its failure probability is at most $2^{-n}$. For vanilla, the minimal draw count is the ceiling of $\log\delta/\log(1-u_V)$; the equivalent follows from $-\log(1-u_V)\sim\epsilon$.

With deterministic distinct binary rewards, zero group variance occurs exactly when every draw selects the same action. These disjoint events have probabilities $u^G$ and $(1-u)^G$. The derivative $G[(1-u)^{G-1}-u^{G-1}]$ shows that $P_{\mathrm{mix}}$ is decreasing for $u\ge1/2$, giving the FPD lower bound. Expanding at $u_V=\epsilon(1+o(1))$ gives $P_{\mathrm{mix}}(u_V)\sim G\epsilon$.
\end{proof}

Nonzero group reward variance is necessary for a nonzero reward-normalized advantage in GRPO-style training without an additional reward-independent signal~\citep{yu2025dapo}. This proposition addresses a particular source of uninformative groups; it does not assert that every higher-accuracy policy produces more mixed groups on every prompt. If future return is stochastic conditional on the action, the corresponding group formula uses the overall success probability, which need not equal $u$.

\paragraph{Local gradient sensitivity.}
In the same binary model, the Fisher information for the margin parameter is
\[
 \mathcal I_M(u)=\E\left[\left(\partial_M\log p(Y)\right)^2\right]
 =u(1-u).
\]
At the matched budget,
\[
 \mathcal I_M(u_F)\ge q(1-q)>0,\qquad
 \mathcal I_M(u_V)\sim\epsilon.
\]
For a subsequent local task with $r_a>r_b$, the margin derivative of expected return is $(r_a-r_b)u(1-u)$. Thus recovering a nondegenerate useful branch also avoids a vanishing local probability-to-logit sensitivity. These scalar statements do not give a lower bound on the full neural Fisher matrix.

\begin{proposition}[KL cost of accessing a useful continuation]
\label{prop:fp-rl-kl}
Let $\pi_0$ be a full-support reference distribution and $\mathcal A$ a nonempty proper event of useful actions, with $\pi_0(\mathcal A)=u_0\in(0,1)$. For a desired probability $r\in(0,1)$,
\[
 \inf_{\pi:\,\pi(\mathcal A)=r}\KL(\pi\|\pi_0)
 =\mathrm{kl}(r\|u_0).
\]
For fixed $r$, a reference mass $u_0=\epsilon(1+o(1))$ incurs
\[
 \mathrm{kl}(r\|u_0)=r\log(1/\epsilon)+O(1).
\]
If instead $u_0$ lies in a fixed compact subinterval of $(0,1)$, this cost is bounded independently of $\epsilon$.
\end{proposition}
\begin{proof}
Apply the KL chain rule to the partition $\{\mathcal A,\mathcal A^c\}$:
\begin{align*}
 \KL(\pi\|\pi_0)
 &=\mathrm{kl}(r\|u_0)\\
 &\quad+r\,\KL(\pi(\cdot\mid\mathcal A)\|
                         \pi_0(\cdot\mid\mathcal A))\\
 &\quad+(1-r)\,\KL(\pi(\cdot\mid\mathcal A^c)\|
                         \pi_0(\cdot\mid\mathcal A^c)).
\end{align*}
The two conditional terms are nonnegative, and both vanish when their reference conditionals are preserved. This proves the infimum and its attainability. Expanding Bernoulli KL at $u_0=\epsilon(1+o(1))$ gives the displayed asymptotic; continuity gives the compact-interval bound.
\end{proof}

At the matched FPD budget, taking $\mathcal A=\{a\}$ gives $u_F\in[1/2,q]$, a fixed compact interval, whereas the vanilla reference mass vanishes. A fixed KL budget can therefore restrict access to a useful branch that an initialization severely underweights. The conclusion concerns the useful branch's mass, rather than entropy in isolation.

For completeness, consider the local KL-regularized objective with fixed continuation values $Q(v)$ and coefficient $\beta_{\mathrm{RL}}>0$,
\[
 \max_\pi\left\{\sum_v\pi(v)Q(v)
                  -\beta_{\mathrm{RL}}\KL(\pi\|\pi_0)\right\}.
\]
The normalization-constrained first-order condition gives
\[
 \pi_*(v)=
 \frac{\pi_0(v)e^{Q(v)/\beta_{\mathrm{RL}}}}
      {\sum_w\pi_0(w)e^{Q(w)/\beta_{\mathrm{RL}}}}.
\]
For a binary pair with value gap $\Delta Q=Q(a)-Q(b)>0$,
\[
 \log\frac{\pi_*(a)}{\pi_*(b)}
 =\log\frac{u_0}{1-u_0}+\frac{\Delta Q}{\beta_{\mathrm{RL}}}.
\]
With fixed value gap and regularization, a vanilla mass $u_0\sim\epsilon$ still yields $\pi_*(a)=O(\epsilon)$. A reference mass bounded away from zero removes this asymptotic barrier. This result applies to the stated KL-regularized local problem; methods without a reference-KL penalty are instead covered by the discovery argument under its sampling assumptions.

\paragraph{What a higher-entropy initialization supports.}
The results identify a useful combination: stronger reward-aligned choices together with nonnegligible probability on alternatives. Soft-target odds matching can recover this combination without an explicit entropy bonus, and calibration can limit further sharpening. Token entropy alone does not establish semantic diversity, useful-action coverage, or better final RL performance. In particular, nucleus sampling can exclude alternatives despite positive full-softmax entropy. The observed entropy trajectory supports an exploration-oriented initialization hypothesis; a subsequent RL comparison is needed to establish its realized downstream advantage.

%% file: sections/appendix/additional_experiments.tex
\section{Additional Experimental Results}
\label{app:additional-experiments}

This appendix supplements Sections~\ref{sec:experiments} and~\ref{sec:ablations} with per-checkpoint trajectories for the 8B single-teacher runs and seed-repeat run, the full per-benchmark fork-weight sweeps, per-domain diagnostics for multi-teacher distillation, and the target sampling trajectories of the D3 adaptive sampler.

\paragraph{Rollout diagnostics.} Repeat coverage is the fraction of overlapping 8-character n-grams in the last 2{,}000 characters of a response that occur more than once, averaged over the rollouts at that step. It is a character-level diagnostic that catches repeated symbolic or formula spans rather than a semantic repetition classifier. In single-teacher runs (Figures~\ref{fig:4b-curves}, \ref{fig:8b-curves}, and~\ref{fig:save5-8b}), entropy denotes mean student token entropy per update; in multi-teacher runs (Figure~\ref{fig:mopd-4x4}), entropy is estimated as the median sampled-token negative log-probability per update. Truncation marks responses that reach the response-token budget.

\subsection{Per-Checkpoint 8B Single-Teacher Curves}
\label{app:8b-curves}

Figure~\ref{fig:8b-curves} plots accuracy across all evaluated checkpoints (steps~20, 40, 60, and 80) alongside per-update training-rollout diagnostics for the four 8B single-teacher methods summarized in Table~\ref{tab:main_8b_24k}.
FPD ($\lambda{=}0.3$) makes a decisive early jump by step~20 (reaching nearly 50 on the seven-benchmark average versus roughly 42 for vanilla OPD, ExOPD, and AOPD) and remains ahead of every baseline at all four checkpoints.
While AOPD ($K{=}32$, $\tau{=}0$) briefly narrows the gap at step~40 (peaking at 51.1) before receding at steps~60--80, FPD continues climbing to 55.0 at step~60 and holds near 54.5 at step~80, sweeping the highest scores across all seven benchmarks.

The rollout diagnostics in panels~(i)--(l) show why vanilla OPD and ExOPD stall under the 8B teacher.
Both collapse in student entropy toward 0.10--0.15 early in training and climb to 50--65\% 8-gram repeat coverage, whereas FPD sustains entropy near 0.45--0.65 and lowers repeat coverage to roughly 28--35\% (even below AOPD's 35--42\%).
At the same time, FPD adopts the 8B teacher's longer reasoning chains (13K--20K tokens), pushing into the 24K truncation boundary (panel~j), which accounts for the sensitivity to larger $\lambda$ documented in Section~\ref{sec:ablation-foil} and Table~\ref{tab:timing}.
\input{figures/8b_curves}

\subsection{Per-Checkpoint 8B Seed-Repeat Curves}
\label{app:8b-seed-repeat}

Figure~\ref{fig:save5-8b} reports an independent 8B seed-repeat run saving a checkpoint every 5 updates (16 checkpoints from step~5 to step~80), together with its per-update rollout diagnostics.
FPD outperforms vanilla OPD on the seven-benchmark average at every evaluated checkpoint, opening a clear margin by steps~15--20 and peaking at 54.9 at step~65 versus 47.4 for vanilla OPD---reproducing the primary 8B run's 55.0 peak (Table~\ref{tab:main_8b_24k}) and confirming that the gap is not an artifact of checkpoint selection.
The largest gains appear on competition benchmarks (AIME24/25, AMC23/24, and OlympiadBench) and MinervaMath, where vanilla OPD remains close to the step-0 student while FPD climbs steadily.

Panels~(i)--(l) show the accompanying rollout dynamics.
Although both arms lengthen against the 24K response limit as they adopt the 8B teacher's longer reasoning chains, vanilla OPD's student entropy collapses from 0.26 to 0.11 while 8-gram repeat coverage climbs from 51\% to 58\%.
FPD instead maintains entropy within 0.13--0.68 (ending at 0.46) and lowers repeat coverage from 48\% down to 30\%, preventing the repetitive degeneration that stalls vanilla OPD.
\input{figures/8b_acc_metrics}

\subsection{Fork-Weight Sweep: Full Results}
\label{app:lambda-all-full}

Table~\ref{tab:lambda_all} reports the seven-benchmark results behind the fork-weight curves in Figure~\ref{fig:lambda-all} for both the 4B and 8B teachers, following the unified 24K evaluation protocol of Section~\ref{sec:exp-setup}.
Across the 4B sweeps, all 17 FPD configurations---nine with Qwen3-4B-Base (B, $\lambda\in[0.15,2.00]$) and eight with Qwen3-4B instruct (I, $\lambda\in[0.15,3.00]$)---outperform 4B vanilla OPD (59.7) within a narrow 1.2-point band (60.8--62.0), improving on AIME25 (35.5--40.0 versus 34.1), AMC24 (66.1--68.4 versus 65.3), and AIME24 (47.7--52.2 versus 47.6) in every single run.
The base foil (B) attains the overall peak of 62.0 ($\lambda{=}1.50$), stays within 0.5 points of the peak across $\lambda\in[0.60,2.00]$ (61.5--62.0), and leads on AMC23 (up to 85.9), AMC24 (up to 68.4 at $\lambda{=}2.00$), and OlympiadBench (up to 63.9); the instruct foil (I) remains flat through $\lambda{=}3.00$ (60.9--61.5) and posts the top scores on AIME24 (52.2 at $\lambda{=}0.15$), AIME25 (40.0 at $\lambda{=}2.00$), and MATH500 (91.6 at $\lambda{=}1.00$).

In the 8B sweep, all six tested FPD settings ($\lambda\in[0.30,2.00]$) remain well above 8B vanilla OPD (46.2), spanning 49.1--55.0.
Unlike the flat 4B curves (where mean lengths stay within 9.2--11.4K and truncation remains 6.0--14.7\% through $\lambda{=}3.00$; Table~\ref{tab:timing}), 8B accuracy peaks at moderate fork weights ($\lambda{=}0.30$ at 55.0 and $\lambda{=}0.50$ at 54.6) and declines monotonically at larger $\lambda$ (52.8 at $\lambda{=}0.70$, down to 49.1 at $\lambda{=}2.00$), with best checkpoints shifting earlier to step~20 at $\lambda\in\{1.50,2.00\}$.
As measured in Table~\ref{tab:timing}, stronger fork weighting pulls the student more rapidly toward the 8B teacher's long-chain style, raising mean rollout length from 13.2K ($\lambda{=}0.30$) to 16.5K ($\lambda{=}2.00$) and training truncation from 31.9\% to 52.4\%; under the fixed 24K ceiling, this truncates more trajectories before a final answer is reached.
\input{tables/7_lambda_all}

\subsection{MOPD Per-Domain Diagnostics}
\label{app:mopd-4x4}

Figure~\ref{fig:mopd-4x4} shows all 12 evaluated checkpoints (steps~19--239) and the per-domain rollout diagnostics for the four multi-teacher arms in Table~\ref{tab:mopd_240}.
In the top row, both FPD arms (fixed 1:1:1 and FPD+D3) separate rapidly from their vanilla counterparts by step~39 and maintain a three-benchmark average near 30--33 through step~239.
Both vanilla arms instead plateau around 27--29 through step~159 before falling to roughly 23--24 after step~179 as LiveCodeBench collapses toward zero (0.0--1.5).
On AIME24, both FPD arms hold steady around 49--53 versus 44--46 for the vanilla arms; on IFBench, all four arms improve over the initial student (18.7), with both FPD arms reaching a peak of 28.0.

Rows~2--4 trace the collapse directly in the training rollouts.
In the Code domain (row~3), both vanilla arms drive token entropy close to zero by step~40 (near 0.01 nats/token, save for a brief transient spike in D3 vanilla around step~180), accompanied by truncation climbing toward 100\%, response lengths saturating at the 23.5K ceiling, and 8-gram repeat coverage rising above 80--95\%.
Both FPD arms preserve code entropy around 0.12--0.18 nats/token, hold code truncation near 40--55\% with mean lengths around 15K--17K tokens, and keep repeat coverage near 65--72\%.
In Math (row~2), FPD sustains higher entropy (roughly 0.28--0.32 versus 0.18--0.20) and lower repetition (36--39\% versus 40--43\%) at comparable response lengths (9K--11K) and low truncation.
In IF (row~4), FPD arms maintain markedly lower repeat coverage than vanilla arms (especially during mid-training) alongside higher rollout entropy, avoiding the degenerate repetition loops seen under vanilla MOPD.
\input{figures/mopd_4x4}

\subsection{D3 Sampling Composition}
\label{app:d3-composition}

Figure~\ref{fig:d3-composition} tracks the target domain sampling probabilities produced by the D3 scheduler (Appendix~\ref{app:d3}) for the D3-MOPD baseline and FPD+D3 over 240 updates.
Both arms begin from a uniform 1:1:1 mixture and initially shift budget toward Math and Code over the first 60 steps, where early reverse-KL descent is steepest.
Thereafter the two trajectories diverge.
In the D3-MOPD baseline (left), the scheduler repeatedly oscillates across Math, Code, and IF as code distillation stalls and degenerates, finishing at 15\% Math, 37\% Code, and 48\% IF.
In FPD+D3 (right), Math and Code stabilize quickly under fork preference supervision; the scheduler concentrates heavily on Code through mid-training (steps~60--180, with brief Math and IF reallocations) and then commits most of the late-stage budget to IF from step~180 onward, finishing at 17\% Math, 17\% Code, and 65\% IF.
This late shift toward IF aligns with Table~\ref{tab:mopd_240}: FPD+D3 achieves the highest IFBench score (28.0 at its best-average checkpoint) while conceding modest margins to the fixed 1:1:1 FPD arm on AIME24 (50.6 versus 53.4) and LCB (17.3 versus 20.5).
\input{figures/d3_composition}

%% file: figures/8b_curves.tex
\begin{figure*}[!t]
  \centering
  \includegraphics[width=\linewidth]{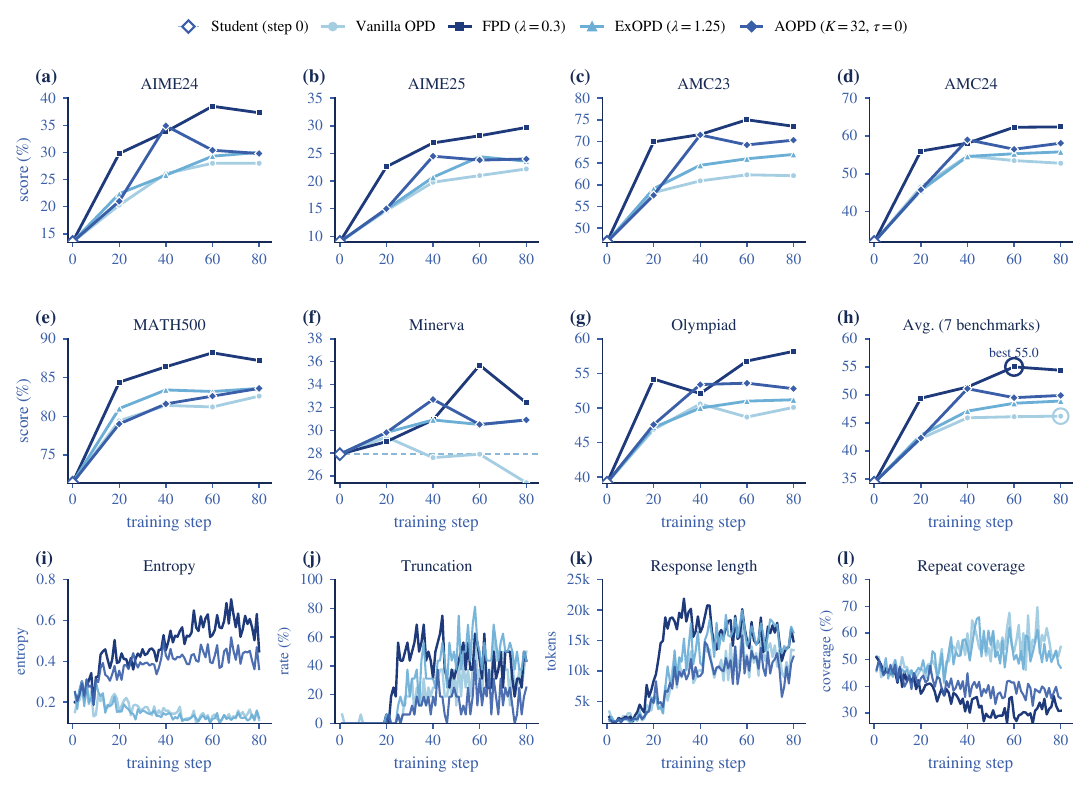}
  \caption{Per-checkpoint accuracy and training-rollout diagnostics for the 8B-teacher runs (checkpoints every 20 updates). Panels~(a)--(g) show individual benchmarks and (h) their unweighted average, with the diamond marking the initial student (step~0). Panels~(i)--(l) track per-update student entropy, rollout truncation rate, mean response length, and 8-gram repeat coverage.}
  \label{fig:8b-curves}
\end{figure*}

%% file: figures/8b_acc_metrics.tex
\begin{figure*}[!t]
  \centering
  \includegraphics[width=\linewidth]{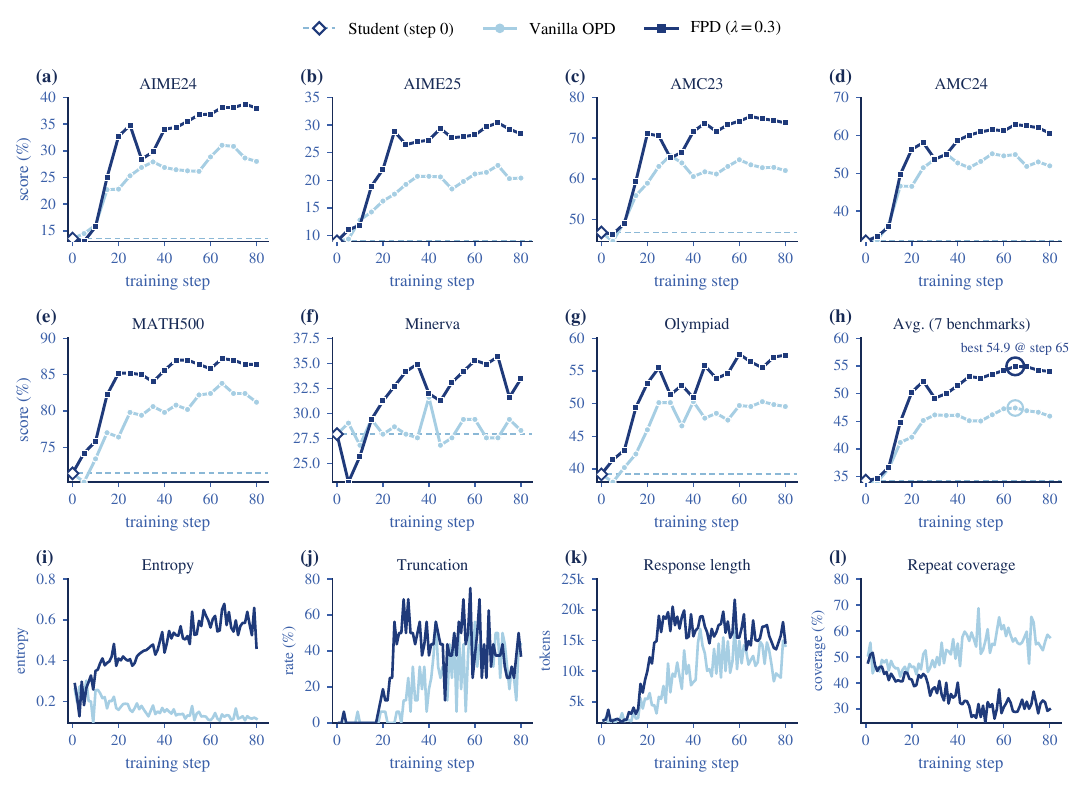}
  \caption{Per-checkpoint accuracy and training-rollout diagnostics for the 8B-teacher seed-repeat run (checkpoints every 5 updates). Panels~(a)--(g) show individual benchmarks and (h) their unweighted average, with the diamond marking the initial student (step~0). Panels~(i)--(l) track per-update mean student entropy, rollout truncation rate, mean response length, and 8-gram repeat coverage.}
  \label{fig:save5-8b}
\end{figure*}

%% file: tables/7_lambda_all.tex
\begin{table*}[!t]
\centering
\small
\setlength{\tabcolsep}{3pt}
\renewcommand{\arraystretch}{1.03}
\caption{Per-benchmark fork-weight ($\lambda$) sweep for single-teacher distillation with the 4B (foils B and I) and 8B teachers under the 24K protocol (best checkpoint per setting; rankings marked separately within the 4B and 8B groups).}
\label{tab:lambda_all}
\begin{adjustbox}{max width=\linewidth}
\begin{tabular}{l|c|c|ccccccc|c}
\toprule
Method & $\lambda$ & Best step & AIME24 & AIME25 & AMC23 & AMC24 & MATH500 & Minerva & Olympiad & Avg. \\
\midrule
Vanilla OPD (4B) & -- & 60 & 47.6 & 34.1 & 83.4 & 65.3 & 89.6 & 36.0 & 62.0 & 59.7 \\
\midrule
\multirow{9}{*}{4B, Qwen3-4B-Base (B)} & 0.15 & 60 & \underline{51.4} & 36.0 & 84.6 & 67.7 & 90.0 & 36.8 & 62.3 & 61.3 \\
 & 0.30 & 80 & 49.5 & 36.1 & 84.6 & 66.1 & 90.4 & 35.7 & 63.1 & 60.8 \\
 & 0.50 & 60 & 50.6 & 37.6 & 84.1 & 66.5 & 90.4 & 37.5 & 61.9 & 61.2 \\
 & 0.60 & 80 & 50.9 & 37.1 & 83.8 & \underline{68.3} & 90.0 & \textbf{39.0} & 62.6 & \underline{61.7} \\
 & 0.70 & 60 & 47.8 & 38.6 & 83.7 & 67.4 & 90.6 & \textbf{39.0} & 63.5 & 61.5 \\
 & 0.80 & 80 & 50.1 & 37.7 & \underline{85.4} & 67.1 & \underline{91.4} & 36.4 & \textbf{63.9} & \underline{61.7} \\
 & 1.00 & 60 & 50.6 & 36.2 & 83.5 & 67.1 & 90.8 & \textbf{39.0} & 63.5 & 61.5 \\
\rowcolor{tblbase!25} & 1.50 & 60 & 49.2 & 38.3 & \textbf{85.9} & 68.1 & 90.2 & \textbf{39.0} & 63.1 & \textbf{62.0} \\
 & 2.00 & 80 & 49.0 & 38.4 & 84.8 & \textbf{68.4} & 89.4 & 37.5 & 62.9 & 61.5 \\
\midrule
\multirow{8}{*}{4B, Qwen3-4B (I)} & 0.15 & 60 & \textbf{52.2} & 37.0 & 82.9 & 66.2 & 90.4 & 36.0 & 61.7 & 60.9 \\
 & 0.30 & 60 & 51.2 & 35.5 & 84.0 & 66.3 & 90.2 & 37.1 & 63.4 & 61.1 \\
 & 0.50 & 60 & 51.1 & 37.4 & 84.9 & 66.5 & 91.0 & 36.4 & 62.0 & 61.3 \\
 & 0.60 & 80 & 47.7 & 38.2 & 84.0 & 67.5 & 90.4 & 36.4 & 62.5 & 61.0 \\
 & 1.00 & 80 & 48.5 & 37.5 & 83.4 & 66.6 & \textbf{91.6} & \textbf{39.0} & 62.9 & 61.4 \\
 & 1.50 & 60 & 48.8 & \underline{39.2} & 82.2 & 66.1 & 90.4 & 36.8 & \underline{63.6} & 61.0 \\
 & 2.00 & 60 & 48.8 & \textbf{40.0} & 83.1 & 66.4 & 91.0 & \underline{38.6} & 62.5 & 61.5 \\
 & 3.00 & 40 & 48.3 & 37.9 & 85.3 & 66.7 & 88.8 & 37.5 & 62.8 & 61.0 \\
\midrule
\midrule
Vanilla OPD (8B) & -- & 80 & 28.0 & 22.2 & 62.1 & 52.8 & 82.6 & 25.4 & 50.1 & 46.2 \\
\midrule
\rowcolor{tblbase!25} & 0.30 & 60 & \textbf{38.5} & \underline{28.2} & \underline{75.0} & \textbf{62.3} & \textbf{88.2} & \underline{35.7} & \textbf{56.8} & \textbf{55.0} \\
 & 0.50 & 80 & \underline{37.3} & \textbf{29.0} & \textbf{75.2} & \underline{61.5} & \underline{87.0} & \textbf{36.0} & \underline{55.9} & \underline{54.6} \\
 & 0.70 & 59 & 36.0 & 27.7 & 69.8 & 61.4 & 86.2 & 34.6 & 54.2 & 52.8 \\
 & 1.00 & 60 & 34.5 & 26.9 & 67.5 & 58.3 & 85.4 & 34.6 & 51.5 & 51.2 \\
 & 1.50 & 20 & 31.0 & 24.3 & 69.6 & 56.2 & 83.8 & 28.7 & 53.1 & 49.5 \\
\multirow{-6}{*}{8B, Qwen3-8B} & 2.00 & 20 & 29.5 & 21.7 & 69.8 & 55.5 & 82.8 & 28.7 & \underline{55.9} & 49.1 \\
\bottomrule
\end{tabular}
\end{adjustbox}
\end{table*}

%% file: figures/mopd_4x4.tex
\begin{figure*}[!t]
  \centering
  \includegraphics[width=\linewidth]{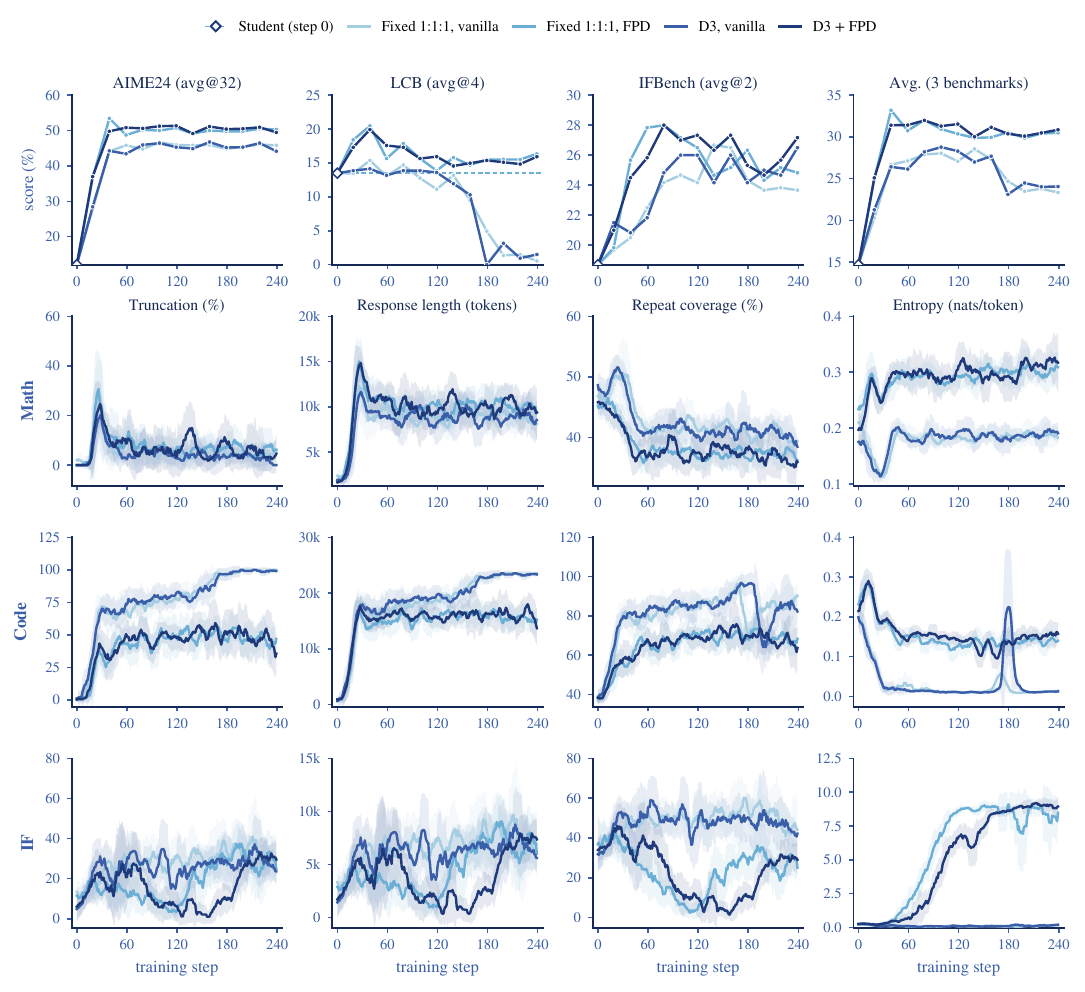}
  \caption{Per-checkpoint accuracy and per-domain rollout diagnostics for the four multi-teacher (MOPD) arms. \emph{Top row:} AIME24, LCB, IFBench, and their unweighted mean (diamond: initial student at step~0). \emph{Rows~2--4:} Math, Code, and IF rollout truncation rate, mean response length, 8-gram repeat coverage, and median sampled-token negative log-probability (entropy) per update.}
  \label{fig:mopd-4x4}
\end{figure*}

%% file: figures/d3_composition.tex
\begin{figure*}[!t]
  \centering
  \includegraphics[width=\linewidth]{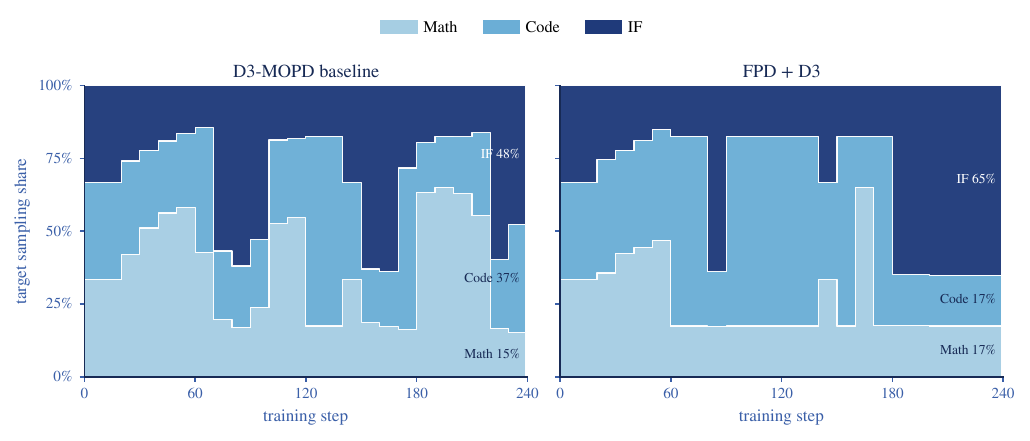}
  \caption{Target per-domain sampling probabilities of the two D3 arms over training. Stacked bands show the scheduler's target shares for Math, Code, and IF; actual batch counts apply multiplicative jitter around these targets.}
  \label{fig:d3-composition}
\end{figure*}

%% file: sections/appendix/complexity.tex
\section{Computational Complexity and Training Overhead}
\label{app:complexity}

We compare one FPD update (Algorithm~\ref{alg:forkpair-step}) with its corresponding vanilla OPD update on the same batch of completed responses. This holds sequence lengths and the base objective fixed. Single-teacher OPD uses the sampled-token base, and multi-teacher OPD uses the pre-update student's Top-$K$ base with $K=16$. In both settings the student generates each response once and the actor takes one optimizer step. The additional work arises when scoring those responses with the foil policy and evaluating the fork preference loss.

\subsection{Teacher and foil scoring and routing}
\label{app:complexity-scoring}

For response $i$, let $L_i$ be the prompt-plus-response length and $R_i$ the number of generated positions passed to the scorer. Let $S=\sum_i R_i$, $N$ be the number of valid loss positions, and $F$ the number of valid forks, so $F\le N\le S$. A fork can be identified only after both the strong teacher and the foil policy have supplied their top token at a position. Thus even a batch with $F=0$ incurs the cost of foil scoring.

On a cache miss, frozen policy $j$ runs one causal forward pass over all $L_i$ tokens, then applies its vocabulary head at the $R_i$ positions that predict response tokens. For a dense decoder with $n_j$ layers, hidden width $h_j$, feed-forward width proportional to $h_j$, vocabulary size $V$, and full causal attention, the arithmetic cost per miss is
\[
 C_j(L,R)=O\!\left(
 \underbrace{n_j Lh_j^2}_{\text{projections and feed-forward layers}}
 +\underbrace{n_j L^2h_j}_{\text{causal attention}}
 +\underbrace{R h_j V}_{\text{vocabulary projection}}\right).
\]
The server obtains the top ID and the log-normalizer from the same output logits. Their $O(RV)$ reductions require a scan over the vocabulary and reuse the projection above. If $\mathcal U_-$ indexes foil requests that miss the cache, the extra forward work is $\sum_{i\in\mathcal U_-} C_{w_i}(L_i,R_i)$, where $w_i$ is the foil policy assigned to response $i$. A cache hit reuses the saved scores and still incurs lookup and data-access costs. With cold caches, a batch of $B$ distinct responses requires $B$ strong-teacher and $B$ foil prefills, compared with $B$ strong-teacher prefills for vanilla OPD. Both frozen policies score completed sequences in a single pass without decoding additional responses.

In multi-teacher FPD, the domain of each prompt indexes a fixed routing table with one cross-domain foil specialist per domain (Equation~\eqref{eq:mopd-weak}). With $D$ domains the table has $D$ entries. Each response is sent to its domain specialist and one foil specialist, so the number of scoring requests per response remains two as $D$ grows. Whereas single-teacher FPD serves an additional frozen foil model, multi-teacher FPD reuses an already resident specialist, which serves strong-teacher requests in its own domain and foil requests in others.

\subsection{Candidate scores and the actor update}
\label{app:complexity-actor}

Once the top IDs are cached, comparing them across the batch requires $O(S)$ work. At each processed position the trainer recovers the strong teacher's log-probability for each candidate $v$ from its cached hidden row $h_t$ and output-head row $W_v$. Subtracting the candidate log-probabilities cancels the normalizer in the soft target,
\[
 \log p^+(v\mid s_t)=W_v^\top h_t-\log Z_t,\qquad
 q_t=\sigma\!\left((W_a-W_b)^\top h_t\right).
\]
The implementation gathers the two strong-head rows before masking nonforks. The extra dot products cost $O(\sum_i R_i h_{+,i})$, or $O(S h_+)$ when all strong teachers have width $h_+$. The single-domain sampled score and the multi-domain $K$ base scores are also needed by their respective vanilla objectives. The cached-ID path avoids another teacher vocabulary projection on the actor.

For fixed candidate scores, computing the sigmoid target, student margin and binary cross-entropy takes constant work per position. Only $F$ positions contribute to the masked sum. The loss kernels operate on blocks of response rows before applying the mask, so the elementwise fork arithmetic scales as $O(S)$. The cost of obtaining the two student scores depends on how each actor queries its vocabulary logits.

\begin{itemize}
 \item \textbf{Single-teacher FPD.} The actor requests the sampled token and both fork candidates in one call. All three share the full-vocabulary log-normalizer already required by the sampled-token base. The two additional selections add $O(S)$ score entries. The actor differentiates the base and fork preference terms in the same backward pass.
 \item \textbf{Multi-teacher FPD.} The actor requests the $K$ base scores in one call, then requests each of the two fork candidate scores separately. These two calls each recompute the row log-normalizer. For already computed student logits, they add two vocabulary scans, together $\Theta(SV)$ arithmetic across vocabulary shards, as well as reductions when the vocabulary is sharded. The scans do not repeat the student's transformer or its output-head projection.
\end{itemize}

By \eqref{eq:pair-grad}, at most two logit coordinates receive a direct fork preference gradient at a fork. Backpropagation passes through shared student parameters, with the same trainable parameters and optimizer state as vanilla OPD. For normalization, the implementation reduces fork counts across context-parallel workers. In multi-teacher FPD this includes a vector of $D$ per-domain counts, whose sum gives the step-wide denominator in \eqref{eq:fp-objective}. Diagnostics can add work beyond the loss itself; for example, logging whether the single-domain student's top choice is a third token requires a vocabulary argmax.

\subsection{Storage, latency and inference}
\label{app:complexity-storage}

For each uncached response, the foil server writes $R_i$ hidden rows, $R_i$ log-normalizers and $R_i$ top IDs. With bfloat16 hidden states, float32 normalizers and int64 IDs, the uncompressed tensor payload is $R_i(2h_{w_i}+4+8)$ bytes, excluding file headers and metadata. This cache is indexed by sequence and policy. It scales with the number of scored positions even if the two policies agree everywhere. On the actor, FPD adds $O(S)$ candidate IDs, scores and masks on top of the base references, which include $O(SK)$ support entries for multi-teacher training.

Each frozen policy's dense output head uses $O(h_j V)$ weights. Its server computes scores in chunks of at most 32 response rows, so the temporary head-logit matrix has at most $32V$ entries. Multi-teacher FPD also accesses float32 copies of the routed teacher heads during reference preparation. Casting a head to float32 from a lower precision can move $O(h_jV)$ values even on the cached-ID path. These working buffers, model activations and actor autograd buffers contribute separately to peak memory.

Within a response, the trainer waits for its strong-teacher score before requesting its foil score. Across responses, frozen-policy scoring can overlap with student generation. Thus the additional prefill count measures compute, while the increase in elapsed training time also depends on sequence lengths, cache misses, scorer queues, and overlap with actor work. At inference the trained student generates directly: fork identification, teacher/foil service calls, and fork preference computation are strictly training-only operations.

\subsection{Measured training cost}
\label{app:timing}

Table~\ref{tab:timing} reports wall-clock measurements for FPD alongside AOPD and ExOPD under the same runtime, teacher-cache path, and 24K protocol. On 4B, soft-label FPD increases per-step time by 9--21\% over vanilla OPD (84.6--93.8\,s versus 77.5\,s), matching AOPD (85.9\,s, +11\%) and staying at or below ExOPD (93.9\,s, +21\%); hard-label FPD ($\lambda{=}1.0$) is slower (98.1\,s, +27\%) due to longer completions and higher truncation (17.3\%). On 8B, FPD increases per-step time by 12--18\% at its optimal weights ($\lambda\in\{0.3,0.5\}$, 107.0--112.4\,s versus 95.4\,s for vanilla and 121.6\,s for ExOPD) and up to 37\% at $\lambda{=}2.0$ (130.4\,s). Most of the increase comes from rollout time rather than the actor update: soft-label FPD generates 20--49\% more response tokens per step on 4B and 51--88\% more on 8B, keeping the decoding batch occupied longer.

Normalized per generated token, soft-label FPD attains 10--24\% higher end-to-end throughput on 4B and 35--40\% higher on 8B. After accounting for truncation, 4B soft-label FPD maintains a 0--9\% effective-throughput advantage (1{,}355--1{,}467 versus 1{,}351\,tok/s), while 8B FPD retains a 23--32\% advantage for $\lambda\in\{0.3,0.5,0.7\}$ (799--856 versus 648\,tok/s) before declining to 505\,tok/s ($-22\%$) at $\lambda{=}2.0$ as truncation climbs from 31.9\% to 52.4\% (versus 20.7\% for vanilla). The total 80-step budget on six GPUs rises from 10.3 to 11.3--12.5 GPU-hours for 4B soft-label FPD (13.1 for hard label) and from 12.7 to 14.3--15.0 GPU-hours for 8B at $\lambda\in\{0.3,0.5\}$ (up to 17.4 at $\lambda{=}2.0$). These measurements include the additional foil-prefill and fork-scoring work; FPD adds no trainable parameters or optimizer state.

Two caveats apply. The effective-throughput column treats every truncated sample as consuming exactly 23{,}552 response tokens, so it is an approximation. Second, because the vanilla and FPD actor-update timers account for active tokens differently, we compare cost using wall-clock step time and generated-token throughput rather than the train-time column alone.
\input{tables/8_timing}

%% file: tables/8_timing.tex
\begin{table*}[!t]
\centering
\footnotesize
\setlength{\tabcolsep}{3pt}
\renewcommand{\arraystretch}{0.98}
\caption{Measured single-teacher training cost averaged over 80 optimizer steps on six NVIDIA H100-80GB GPUs (four actor GPUs, one strong-teacher server, and one foil server). Eff.\ tok/s estimates non-truncated response throughput; GPU-h covers the full 80-step run. B and I denote the Qwen3-4B-Base and Qwen3-4B-Instruct foils; the hard-label row uses target $q{=}1$.}
\label{tab:timing}
\begin{tabular}{l|ccccccccc}
\toprule
Method & Step (s) & Rollout (s) & Train (s) & Resp. k/step & Avg. len & Trunc. (\%) & E2E tok/s & Eff. tok/s & GPU-h \\
\midrule
\multicolumn{10}{l}{\textit{4B teacher (Qwen3-4B-Math-RL)}} \\
4B vanilla & 77.5 & 58.9 & 17.1 & 122.1 & 7{,}631 & 4.6 & 1{,}575 & 1{,}351 & 10.34 \\
4B AOPD ($K{=}32$, $\tau{=}0$) & 85.9 & 64.5 & 19.7 & 127.3 & 7{,}955 & 4.4 & 1{,}482 & 1{,}290 & 11.45 \\
4B ExOPD ($\lambda{=}1.25$) & 93.9 & 66.4 & 25.9 & 132.7 & 8{,}291 & 4.9 & 1{,}412 & 1{,}215 & 12.52 \\
4B FPD $\lambda{=}1.0$ (I) & 85.6 & 69.3 & 14.7 & 159.4 & 9{,}964 & 9.1 & 1{,}862 & 1{,}463 & 11.42 \\
4B FPD $\lambda{=}1.5$ (I) & 90.9 & 73.6 & 15.8 & 173.3 & 10{,}832 & 12.6 & 1{,}906 & 1{,}385 & 12.12 \\
4B FPD $\lambda{=}2.0$ (I) & 91.4 & 74.0 & 15.8 & 174.1 & 10{,}878 & 12.3 & 1{,}905 & 1{,}399 & 12.18 \\
4B FPD $\lambda{=}3.0$ (I) & 93.8 & 75.8 & 16.4 & 182.4 & 11{,}399 & 14.7 & 1{,}945 & 1{,}355 & 12.50 \\
4B FPD $\lambda{=}1.0$ (B) & 84.6 & 67.3 & 15.6 & 147.0 & 9{,}188 & 6.1 & 1{,}739 & 1{,}467 & 11.27 \\
4B FPD $\lambda{=}1.5$ (B) & 85.2 & 68.1 & 15.4 & 147.4 & 9{,}213 & 6.0 & 1{,}731 & 1{,}465 & 11.35 \\
4B FPD $\lambda{=}1.0$ (I, hard label) & 98.1 & 78.0 & 18.5 & 191.7 & 11{,}978 & 17.3 & 1{,}953 & 1{,}287 & 13.09 \\
4B FPD $\lambda{=}1.0$ (self-foil) & 91.6 & 72.5 & 17.4 & 173.0 & 10{,}812 & 12.4 & 1{,}888 & 1{,}377 & 12.22 \\
\midrule
\multicolumn{10}{l}{\textit{8B teacher (Logics-STEM-8B-RL)}} \\
8B vanilla & 95.4 & 71.6 & 22.3 & 139.8 & 8{,}736 & 20.7 & 1{,}465 & 648 & 12.72 \\
8B AOPD ($K{=}32$, $\tau{=}0$) & 95.0 & 69.6 & 23.7 & 130.8 & 8{,}177 & 11.2 & 1{,}377 & 934 & 12.67 \\
8B ExOPD ($\lambda{=}1.25$) & 121.6 & 82.4 & 37.5 & 178.4 & 11{,}150 & 30.5 & 1{,}467 & 521 & 16.21 \\
8B FPD $\lambda{=}0.3$ & 107.0 & 86.0 & 19.5 & 211.7 & 13{,}233 & 31.9 & 1{,}978 & 856 & 14.27 \\
8B FPD $\lambda{=}0.5$ & 112.4 & 90.6 & 20.2 & 226.7 & 14{,}167 & 34.8 & 2{,}017 & 851 & 14.99 \\
8B FPD $\lambda{=}0.7$ & 117.1 & 94.2 & 21.3 & 238.1 & 14{,}879 & 38.4 & 2{,}033 & 799 & 15.61 \\
8B FPD $\lambda{=}1.0$ & 121.1 & 95.6 & 23.8 & 247.8 & 15{,}485 & 42.9 & 2{,}047 & 711 & 16.14 \\
8B FPD $\lambda{=}1.5$ & 126.8 & 100.5 & 24.6 & 257.0 & 16{,}065 & 48.8 & 2{,}027 & 576 & 16.91 \\
8B FPD $\lambda{=}2.0$ & 130.4 & 103.1 & 25.6 & 263.3 & 16{,}457 & 52.4 & 2{,}020 & 505 & 17.38 \\
\bottomrule
\end{tabular}
\end{table*}

%% file: sections/appendix/case_study.tex
\section{A Fork in a Mathematical Reasoning Rollout}
\label{app:case-study}

At the prefix examined here, the student continues with \texttt{So}, while the strong teacher assigns most of its probability to \texttt{Wait}. We use a vanilla-OPD rollout to trace this disagreement from the student's reasoning to the supervision supplied by each objective. Applying the losses in Appendix~\ref{app:gradients} to the three models' next-token probabilities makes the fork preference signal of FPD explicit.

\subsection{A disagreement at a student-generated prefix}
\label{app:case-record}

The student is Qwen3-1.7B, supervised by a Qwen3-4B math teacher trained with RL. A frozen Qwen3-4B model serves as the foil policy. All three models are scored on the same student-generated prefix of the following problem.

\begin{tcolorbox}[title={Problem},breakable,colback=white,colframe=linegray,colbacktitle=boxgray,coltitle=black,fonttitle=\bfseries,coltext=black]
The number $125$ can be expressed as a sum of pairwise coprime integers, each greater than $1$. Determine the maximum number of terms possible in such a sum.
\end{tcolorbox}

The student starts with small primes and repeatedly adjusts the candidate set. Near the selected prefix, its running totals become inconsistent and it reintroduces terms, then recalls that each number can appear only once. The excerpts preserve the generated statements and arithmetic, with mathematical notation typeset for readability and omissions marked by $[\ldots]$.

\begin{tcolorbox}[title={Student prefix before the fork (excerpt)},breakable,colback=white,colframe=linegray,colbacktitle=boxgray,coltitle=black,fonttitle=\bfseries,coltext=black]
\small
So maybe try a smaller set. Let's go back.

Try:
\begin{itemize}[nosep,leftmargin=*]
 \item $2,3,5,7,11,13$ $\rightarrow$ sum $=41$
 \item Add $17$ $\rightarrow$ sum $=58$
 \item Add $19$ $\rightarrow$ sum $=77$
 \item Add $23$ $\rightarrow$ sum $=100$
 \item Remove $29$ $\rightarrow$ sum $=77$
 \item Add $29$ $\rightarrow$ sum $=106$
 \item Add $17$ $\rightarrow$ sum $=123$
 \item Add $2$ $\rightarrow$ sum $=125$
\end{itemize}
But we need to be \textbf{pairwise coprime}, and we can only use \textbf{one of each number}.
\end{tcolorbox}

The student next samples \texttt{So}, which is also the foil policy's first choice. The strong teacher instead prefers \texttt{Wait}, giving the fork candidates $a=\texttt{Wait}$ and $b=\texttt{So}$. Table~\ref{tab:case-token-scores} shows the model probabilities and the token counts from a diagnostic set of 64 draws at this prefix. Although \texttt{Wait} receives 78.3272\% probability from the teacher, it ranks sixth under the student, with probability 0.9757\%, and appears in none of the 64 draws.

\begin{table}[htbp]
\centering
\small
\caption{Next-token full-vocabulary probabilities (\%) and counts across 64 diagnostic draws at this prefix; dashes mark tokens outside the strong teacher's top-12 list.}
\label{tab:case-token-scores}
\begin{tabular}{lrrrrr}
\toprule
Token & Token ID & Student (\%) & Strong teacher (\%) & Foil (\%) & Count \\
\midrule
\texttt{So} & 4416 & 53.2721 & 1.8421 & 41.0069 & 33 \\
\texttt{Let} & 10061 & 25.1640 & 0.6777 & 31.9362 & 19 \\
\texttt{\detokenize{---\n\n}} & 44364 & 11.8866 & --- & 4.8976 & 11 \\
\texttt{Try} & 21453 & 3.8590 & --- & 4.8976 & 1 \\
\texttt{Wait} & 14190 & 0.9757 & 78.3272 & 7.1259 & 0 \\
\bottomrule
\end{tabular}
\end{table}

After \texttt{So}, the student returns to the same set of nine primes and again observes that adding the next prime exceeds $125$.

\begin{tcolorbox}[title={Actual student continuation (excerpt)},breakable,colback=white,colframe=linegray,colbacktitle=boxgray,coltitle=black,fonttitle=\bfseries,coltext=black]
\small
\textbf{So} try this set:
\begin{itemize}[nosep,leftmargin=*]
 \item $2,3,5,7,11,13,17,19,23$
 \item These are all prime numbers.
 \item Sum:
 \[
 2+3+5+7+11+13+17+19+23=100
 \]
\end{itemize}
Still too small. Add $29$: sum $=129$ $\rightarrow$ too big.

So instead, use \textbf{some of the primes and some composites}. $[\ldots]$
\end{tcolorbox}

The student's rollout concludes with a final answer of $9$. Section~\ref{app:case-answer} verifies that the maximum is $8$.

\subsection{Supervision supplied by the two objectives}
\label{app:case-supervision}

Fix this prefix $s$ and write $p=p_\theta(\cdot\mid s)$ for the student distribution. We evaluate the single-position losses with freely adjustable logits, then account for their weights in the batch objective. All numerical values are computed in full precision before rounding for display.

\paragraph{Sampled-token feedback.}
For the sampled token $b=\texttt{So}$, the advantage in \eqref{eq:opd-adv} is
\begin{equation}
 A(b)=\log\frac{p^+(b)}{p(b)}
 =\log\frac{0.0184207987}{0.5327209234}
 \approx -3.36452 .
 \label{eq:case-advantage}
\end{equation}
The resulting logit descent direction from \eqref{eq:opd-logit} is
\begin{equation}
 \begin{aligned}
 g_b(a)&=-A(b)\,p(a)
        \approx 0.032828,\\
 g_b(b)&=A(b)\bigl(1-p(b)\bigr)
        \approx -1.572169 .
 \end{aligned}
 \label{eq:case-base-gradient}
\end{equation}
The update suppresses \texttt{So} and indirectly raises the \texttt{Wait} logit through softmax normalization. The latter increase depends on the student's probability of \texttt{Wait} and the teacher's score for \texttt{So}. Every other unchosen token receives a push scaled by its own student probability. Thus this draw discourages the sampled continuation while providing little information about the teacher's relative preference among the alternatives.

\paragraph{Fork preference feedback (FPD).}
The foil policy identifies \texttt{So} as the competing candidate, and the strong teacher scores both tokens. Conditioning each model on the pair gives
\begin{equation}
 \begin{aligned}
 q
 &=\frac{p^+(a)}{p^+(a)+p^+(b)}
   =\frac{0.7832722664}{0.7832722664+0.0184207987}
   \approx 0.977023,\\
 u
 &=\frac{p(a)}{p(a)+p(b)}
   =\frac{0.0097571239}{0.0097571239+0.5327209234}
   \approx 0.017986 .
 \end{aligned}
 \label{eq:case-conditional-probabilities}
\end{equation}
The student and teacher therefore favor opposite candidates. Their logit margins and the fork preference descent direction are
\begin{equation}
 \begin{aligned}
 M&=\log\frac{p(a)}{p(b)}\approx -4.0000,\\
 M^+&=\log\frac{p^+(a)}{p^+(b)}\approx 3.7500,\\
 -\nabla_z\ell^{\mathrm{fork}}
 &=(q-u)(e_a-e_b)
 \approx 0.959036\,(e_a-e_b).
 \end{aligned}
 \label{eq:case-pair-gradient}
\end{equation}
The fork preference loss directly raises the \texttt{Wait} logit and lowers the \texttt{So} logit according to this conditional preference gap. The comparison is available even if the rollout selects a third token, and its gradient vanishes once the student's conditional odds match the teacher's.

In the batch objective \eqref{eq:fp-objective}, the base direction at this position is weighted by $1/N$ and the fork preference direction by $\lambda/F$, where $N$ counts valid tokens and $F>0$ counts forks. For an isolated fork with $N=F=1$, a step of size $\eta$ directly on the logits adds
\[
 2\eta\lambda(q-u)\approx1.918073\,\eta\lambda
\]
to the margin through the fork preference term. In training, these gradients propagate through shared model parameters and the optimizer.

The pair's total student probability is $p(a)+p(b)\approx0.542478$. This case illustrates a rare teacher-preferred token competing with a high-probability rival. The low-total-probability regime analyzed in Section~\ref{sec:method-slow} provides an additional motivation for supervising conditional preferences.

\subsection{Reasoning after the teacher-preferred token}
\label{app:case-continuations}

To inspect how reasoning unfolds after the pivot, we condition both the strong teacher and the student on \texttt{Wait}. The teacher explicitly calls for a more systematic argument, whereas the student recalls the uniqueness constraint but makes an arithmetic error by summing the first ten primes to $125$. In the boxes below, the bracketed \texttt{Wait} indicates the conditioning token preceding each continuation.

\begin{tcolorbox}[title={Strong teacher continuation conditioned on \texttt{Wait} (excerpt)},breakable,colback=white,colframe=linegray,colbacktitle=boxgray,coltitle=black,fonttitle=\bfseries,coltext=black]
\small
[\texttt{Wait}], this approach is getting messy. Let's think more systematically.

$[\ldots]$

But composite numbers must not share any prime factors with the others.

For example, $4$ is not coprime with $2$, so if we use $4$, we can't use $2$. $[\ldots]$
\end{tcolorbox}

\begin{tcolorbox}[title={Student continuation conditioned on \texttt{Wait} (excerpt)},breakable,colback=white,colframe=linegray,colbacktitle=boxgray,coltitle=black,fonttitle=\bfseries,coltext=black]
\small
[\texttt{Wait}] --- we can only use each number \textbf{once} in the sum.

$[\ldots]$

Let's try the following set of \textbf{pairwise coprime integers}:
\[
2,3,5,7,11,13,17,19,23,29 .
\]
These are all \textbf{prime numbers}, and they are \textbf{pairwise coprime}.

Let's compute their sum:
\[
2+3+5+7+11+13+17+19+23+29=125
\]
\end{tcolorbox}

The displayed primes sum to $129$. Although these short continuations do not reach a final answer, they illustrate that accurate reasoning after the selected token also matters for task success. Appendix~\ref{app:task-return} formalizes this via continuation values when relating token-preference changes to terminal reward.

\subsection{Verification of the maximum number of terms}
\label{app:case-answer}

Let $n_1,\ldots,n_k>1$ be pairwise coprime integers summing to $125$. Choose a prime divisor $r_i$ of each $n_i$. Pairwise coprimality makes the $r_i$ distinct, and $n_i\ge r_i$.

If $k\ge10$, the sum is at least the sum of the first ten primes,
\[
 \sum_{i=1}^k n_i
 \ge 2+3+5+7+11+13+17+19+23+29
 =129>125.
\]
For $k=9$, at most one summand can be even. If exactly one were even, the other eight would be odd and the total would be even. Hence all nine summands must be odd to sum to $125$, and their distinct prime divisors must also be odd. This yields
\[
 \sum_{i=1}^{9}n_i
 \ge 3+5+7+11+13+17+19+23+29
 =127>125.
\]
Eight summands are attained by the distinct primes
\[
 125=2+3+5+7+11+13+17+67.
\]
Therefore the maximum number of terms is $8$.